\documentclass{article}

\usepackage{microtype}

\usepackage{booktabs} 

\usepackage{amsmath,amsfonts}
\usepackage{algorithmic}
\usepackage{textcomp}
\usepackage{amssymb,bm,amsthm}
\usepackage{subfig}
\usepackage{floatrow}
\usepackage{multirow}
\usepackage{array}
\usepackage{color}
\usepackage{colortbl}
\usepackage{wrapfig}

\usepackage{graphicx}
\usepackage{bbding}
\usepackage[utf8]{inputenc} 
\usepackage[T1]{fontenc}    
\usepackage{hyperref}       
\usepackage{url}            
\usepackage{nicefrac}       
\usepackage{xcolor}         
\usepackage{subcaption}
\usepackage{graphicx}

\newcommand\etal{\textit{et al.}}
\newcommand\ie{\textit{i.e.,}}
\newcommand\eg{\textit{e.g.,}}

\newcommand\wrt{\textit{w.r.t.}}
\newcommand\etc{\textit{etc.}}

\newcommand\niid{\textit{non-i.i.d.}}

\newcommand{\norm}[1]{\left\lVert#1\right\rVert}

\newcommand{\beq}{\begin{equation}}
\newcommand{\eeq}{\end{equation}}
\newcommand{\beqnn}{\begin{equation*}}
\newcommand{\eeqnn}{\end{equation*}}
\newcommand{\beqy}{\begin{eqnarray}}
\newcommand{\eeqy}{\end{eqnarray}}
\newcommand{\beqynn}{\begin{eqnarray*}}
\newcommand{\eeqynn}{\end{eqnarray*}}
\newcommand{\bit}{\begin{itemize}}
\newcommand{\eit}{\end{itemize}}
\newcommand{\ben}{\begin{enumerate}}
\newcommand{\een}{\end{enumerate}}
\newcommand{\bex}{\begin{example}}
\newcommand{\eex}{\end{example}}

\newcommand{\balg}[1]{\begin{algorithm} \caption{#1}}
\newcommand{\ealg}{\end{algorithm}}

\newcommand{\balgc}{\begin{algorithmic}[1]}
\newcommand{\ealgc}{\end{algorithmic}}

\newcommand{\bary}{\begin{array}}
\newcommand{\eary}{\end{array}}
\newcommand{\bmx}{\begin{bmatrix}}
\newcommand{\emx}{\end{bmatrix}}
\newcommand{\bsmx}{\left[\begin{smallmatrix}}
\newcommand{\esmx}{\end{smallmatrix}\right]}
\newcommand{\bmxc}[1]{\left[\begin{array}{@{}#1@{}}}
\newcommand{\emxc}{\end{array}\right]}
\newcommand{\bcn}{\begin{center}}
\newcommand{\ecn}{\end{center}}

\newcommand{\diag}{\mathrm{diag}}

\newcommand{\Rbb}{{\mathbb{R}}}

\providecommand{\norm}[1]{\lVert#1\rVert}

\providecommand{\abs}[1]{\left| #1 \right|}

\DeclareMathOperator*{\argmax}{arg\,max}

\usepackage{tikz}
\usetikzlibrary{trees, positioning}
\usetikzlibrary{shadows}
\usepackage{mathrsfs}
\usepackage{forest}
\tikzset{
    my node/.style={
        draw=gray,
        inner color=gray!5,
        outer color=gray!10,
        thick,
        minimum width=3cm,
        text height=1.5ex,
        text depth=0ex,
        font=\sffamily,
    }
}

\usepackage{caption}

\usepackage{nccmath}

\usepackage{mathtools}

\usepackage[accepted]{icml2024}

\usepackage[capitalize,noabbrev]{cleveref}

\theoremstyle{plain}
\theoremstyle{definition}
\theoremstyle{remark}

\usepackage[textsize=tiny]{todonotes}

\icmltitlerunning{The Heterophilic Graph Learning Handbook}

\begin{document}

\twocolumn[
\icmltitle{The Heterophilic Graph Learning Handbook: \\Benchmarks, Models, Theoretical Analysis, Applications and Challenges
}

\icmlsetsymbol{equal}{*}

\vspace{\baselineskip}
\begin{icmlauthorlist}
\icmlauthor{Sitao Luan}{mila,mcgill}
\icmlauthor{Chenqing Hua}{mila,mcgill}
\icmlauthor{Qincheng Lu}{mcgill}
\icmlauthor{Liheng Ma}{mila,mcgill}
\icmlauthor{Lirong Wu}{westlake}
\icmlauthor{Xinyu Wang}{mcgill}
\icmlauthor{Minkai Xu}{stanford}
\icmlauthor{Xiao-Wen Chang}{mcgill}
\icmlauthor{Doina Precup}{mila,deepmind,mcgill}
\icmlauthor{Rex Ying}{yale}
\icmlauthor{Stan Z. Li}{westlake}
\icmlauthor{Jian Tang}{mila,hec,cifar}
\icmlauthor{Guy Wolf}{mila,udem,cifar}
\icmlauthor{Stefanie Jegelka}{mit,tum}
\end{icmlauthorlist}

\icmlaffiliation{mcgill}{McGill University}
\icmlaffiliation{mila}{Mila - Quebec Artificial Intelligence Institute}
\icmlaffiliation{westlake}{Westlake University}
\icmlaffiliation{stanford}{Stanford University}
\icmlaffiliation{yale}{Yale University}
\icmlaffiliation{deepmind}{Google DeepMind}
\icmlaffiliation{hec}{HEC Montreal}
\icmlaffiliation{udem}{Université de Montréal}
\icmlaffiliation{mit}{Massachusetts Institute of Technology}
\icmlaffiliation{cifar}{CIFAR AI Chair}
\icmlaffiliation{tum}{TU Munich}

\icmlcorrespondingauthor{Sitao Luan}{sitao.luan@mail.mcgill.ca}

\vspace{\baselineskip}
\vspace{\baselineskip}
\begin{abstract}    
\vspace{\baselineskip}
\vspace{\baselineskip}
Homophily principle, \ie{} nodes with the same labels or similar attributes are more likely to be connected, has been commonly believed to be the main reason for the superiority of Graph Neural Networks (GNNs) over traditional Neural Networks (NNs) on graph-structured data, especially on node-level tasks. However, recent work has identified a non-trivial set of datasets where GNN's performance compared to the NN's is not satisfactory. Heterophily, \ie{} low homophily, has been considered the main cause of this empirical observation.  People have begun to revisit and re-evaluate most existing graph models, including graph transformer and its variants, in the heterophily scenario across various kinds of graphs, \eg{} heterogeneous graphs, temporal graphs and hypergraphs. Moreover, numerous graph-related applications are found to be closely related to the heterophily problem. In the past few years, considerable effort has been devoted to studying and addressing the heterophily issue.

In this survey, we provide a comprehensive review of the latest progress on heterophilic graph learning, including an extensive summary of benchmark datasets and evaluation of homophily metrics on synthetic graphs, meticulous classification of the most updated supervised and unsupervised learning methods, thorough digestion of the theoretical analysis on homophily/heterophily, and broad exploration of the heterophily-related applications. Notably, through detailed experiments, we are the first to categorize benchmark heterophilic datasets into three sub-categories: malignant, benign and ambiguous heterophily. Malignant and ambiguous datasets are identified as the real challenging datasets to test the effectiveness of new models on the heterophily challenge. Finally, we propose several challenges and future directions for heterophilic graph representation learning. We hope this paper can become a friendly tutorial and an encyclopedia for your research on heterophilic graph representation learning. (Updated until June 30th, 2024.)
\vspace{\baselineskip}
\vspace{\baselineskip}
\end{abstract}


\vskip 0.3in
]



\printAffiliationsAndNotice{}

\setcitestyle{numbers,square}

\clearpage
\tableofcontents
\newpage

\section{Introduction}
\label{sec:introduction}
As a generic data structure, graphs are capable of modeling complex relations among objects in many real-world problems~\cite{sperduti1993encoding, goller1996learning, sperduti1997supervised, frasconi1998general}. In the last decade, Graph Neural Networks (GNNs) have gained popularity as a powerful tool for graph-based machine learning tasks. By combining graph signal processing and convolutional neural networks, various GNN architectures have been proposed~\cite{scarselli2008graph, bruna2014spectral, defferrard2016convolutional, kipf2016classification, hamilton2017inductive, gilmer2017neural, velivckovic2018graph, wang2020multi, xu2018powerful, liao2019lanczos, luan2019break, hua2022high, zhang2024deep, hua2024effective}, and have been shown to outperform traditional neural networks (NNs) in the modeling of relational data and graph-structured data~\cite{monti2017geometric, zhang2018link, pfaff2020learning,  klissarov2020reward, wu2021self, gaudelet2021utilizing, hua2024mudiff}.

\begin{figure*}[htbp!]
    \centering
     {
     \subfloat[Homophilic Graph]{
     \captionsetup{justification = centering}
     \includegraphics[width=0.46\textwidth]{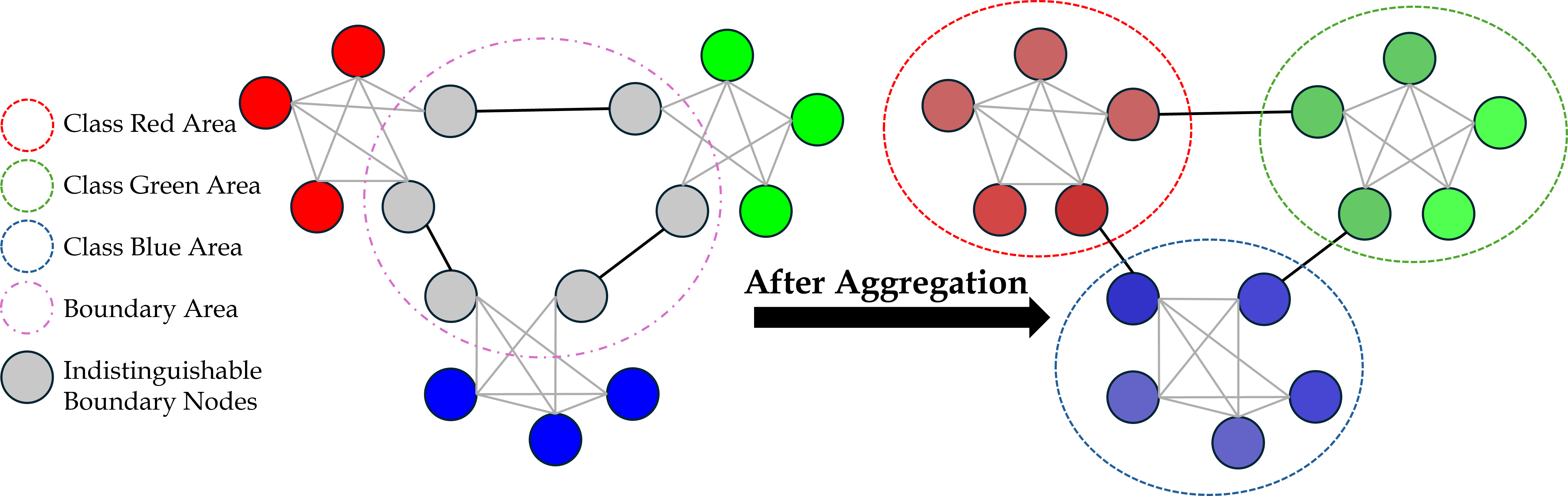}
     }
     \subfloat[Heterophilic Graph]{
     \captionsetup{justification = centering}
     \includegraphics[width=0.46\textwidth]{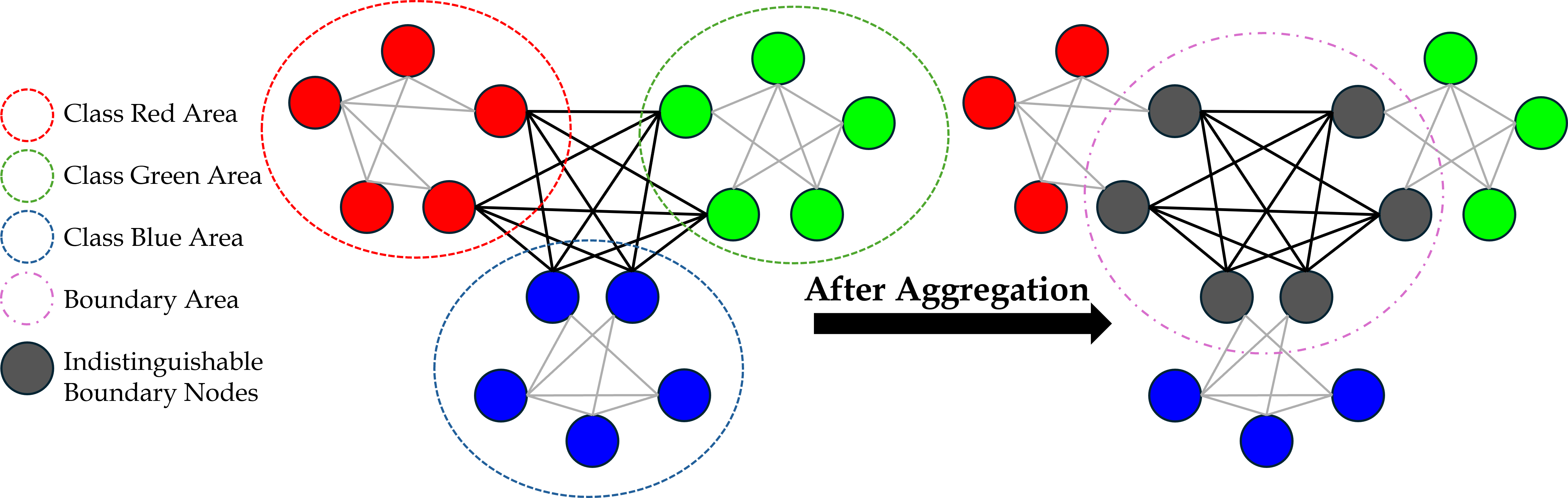}
     }
     }
     \caption{Example of feature aggregation in homophilic and heterophilic graphs. There are $3$ classes of nodes, and each color indicates different node features.}
     \label{fig:example_heterophily_homophily}
\end{figure*}

The success of GNNs, especially on node-level tasks, is commonly believed to be rooted in the homophily principle~\cite{mcpherson2001birds}. Homophily is a concept in sociology~\cite{mcpherson2001birds, ferguson2017false} and evolutionary biology~\cite{jiang2013assortative}, stating the tendency that individuals with similar characteristics are easier to communicate and bond with each other. In graph representation learning, it describes the networks where connected nodes are more likely to have similar labels~\cite{pei2020geom} or attributes~\cite{hamilton2020graph}. As shown in Figure~\ref{fig:example_heterophily_homophily}(a), for the indiscernible boundary nodes, homophilic structure provides extra useful information to their aggregated features over the original node features. Such relational inductive bias is thought to be a major contributor to the superiority of GNNs over traditional NNs on various tasks~\cite{battaglia2018relational}.

On the other hand, the lack of homophily, \ie{} heterophily~\cite{lozares2014homophily}, is considered as the main cause of the inferiority of GNNs on heterophilic graphs. As shown in Figure~\ref{fig:example_heterophily_homophily}(b), the boundary nodes have more heterophilic neighbors than homophilic ones. Since heterophilic edges connect nodes of different classes, they can lead to mixed and indistinguishable node embeddings, making the classification task more difficult for GNNs~\cite{zhu2020beyond, luan2022complete}. Numerous analyses, benchmarks, and models have been proposed to study and address the heterophily challenge lately~\cite{pei2020geom, zhu2020beyond, luan2022complete, bo2021beyond, lim2021new, chien2021adaptive, yan2022two, he2021bernnet, luan2021heterophily, li2022finding, chen2023dirichlet, wang2022acmp, luan2022revisiting}. And a large number of applications are found to be related to heterophily problems, \eg{} fraud/anomaly detection~\cite{gao2023addressing}, graph clustering~\cite{pmlr-v202-pan23b}, recommender systems~\cite{jiang2024challenging}, generative models~\cite{lin2022hl}, link prediction~\cite{zhou2022link}, graph classification~\cite{ye2022incorporating} and coloring~\cite{wang2024graph}, \etc{} 


In this article, we provide a comprehensive review of recent research related to heterophily, including benchmarks, models, theoretical studies and analysis, applications, challenges and future directions. Compared to the existing surveys on heterophily~\cite{zheng2022graph, zhang2023survey, gong2024towards}, our contributions are: 

\begin{itemize}
    \item{Detailed and novel categorization of homophily metrics (Section~\ref{sec:homophily_metrics_beyond}) and heterophilic benchmarks (Section~\ref{sec:benchmark_evaluation}):}
    \begin{itemize}
    \item{We are the first to summarize the homophily metrics comprehensively and compare them meticulously on synthetic graphs with various generation methods}
    \item{We are the first to propose and identify benign, malignant and ambiguous heterophily datasets}
    \item{We summarize and categorize the homogeneous and heterogeneous graph benchmarks extensively}
    \end{itemize}
    \item Thorough review of models for heterophily (Section~\ref{sec:graph_models_for_heterophily},~\ref{sec:heterophily_on_heterogeneous_graphs}), including
    \begin{itemize}
        \item Latest updated models for heterophily on homogeneous graphs (Section~\ref{sec:graph_models_for_heterophily})
        \item The first review of models for heterophily problem on heterogeneous graphs (Section~\ref{sec:heterophily_on_heterogeneous_graphs})
        \end{itemize}
    \item Extensive summary of unsupervised learning methods on heterophilic graphs (Section~\ref{sec:unsupervised_learning})
    \item First to provide in-depth review of theoretical studies on homophily and heterophily (Section~\ref{sec:theoretical_understanding})
    \item Full outline of heterophily related applications \eg fraud/anomaly detection, privacy, federated learning, knowledge distillation and point cloud segmentation, \etc{}(Section~\ref{sec:related_applications})
    \item The most broad and insightful overview of the challenges and future directions \eg large language model (LLM), graph foundation model and scaling law, \etc{}(Section~\ref{sec:challeges_future_directions})
\end{itemize}


\section{Preliminaries}
\label{sec:preliminaries}
\subsection{Notation and Basic Concepts}

We define a graph $\mathcal{G}=(\mathcal{V},\mathcal{E})$, where $\mathcal{V}=\{1,2,\ldots,N\}$ is the set of nodes and $\mathcal{E}=\{e_{ij}\}$ is the set of edges without self-loops. The adjacency matrix of $\mathcal{G}$ is denoted by $A=(A_{i,j})\in \Rbb^{N\times N}$ with $A_{i,j}=1$ if there is an edge between nodes $i$ and $j$, otherwise $A_{i,j}=0$.  The diagonal degree matrix of $\mathcal{G}$ is denoted by $D$ with $D_{i,i} = d_i = \sum_j A_{i,j}$. The neighborhood set $\mathcal{N}_i$ of node $i$ is defined as $\mathcal{N}_i=\{j: e_{ij} \in \mathcal{E}\}$ with size $\abs{\mathcal{N}_i}$. A graph signal is a vector in $\mathbb{R}^N$, whose $i$-th entry is a feature of node $i$.  Additionally, we use ${X} \in \mathbb{R}^{N\times F_h}$ to denote the feature matrix, whose columns are graph signals and $i$-th row ${X_{i,:}} = \bm{x}_i^\top$ is the feature vector of node $i$  (we use \textbf{bold} font for vectors).
The label encoding matrix is $Y \in \mathbb{R}^{N\times C}$, where $C$ is the number of classes, and its $i$-th row $Y_{i,:}$ is the one-hot encoding of the label of node $i$. For simplicity, we denote $y_i = \argmax_{j} Y_{i,j} \in \{1,2,\dots C\}$. 
The indicator function $\bm{1}_B$ equals 1 when event $B$ happens and 0 otherwise. 

For nodes $i,j \in \mathcal{V}$, if $y_i=y_j$, then they are considered as \textit{intra-class nodes}; if $y_i \neq y_j$, then they are considered to be \textit{inter-class nodes}. Similarly, an edge $e_{i,j} \in \mathcal{E}$ is considered to be an \textit{intra-class edge} if $y_i = y_j$, and an \textit{inter-class edge} if $y_i \neq y_j$.

\paragraph{Graph Laplacian, Affinity Matrices and Variants}
The (combinatorial) graph Laplacian is defined as $L = D - A$, which is Symmetric Positive Semi-Definite (SPSD) ~\cite{chung1997spectral}. Its eigendecomposition is $L=U\Lambda U^T$, where the columns $\bm{u}_i$ of $U\in \Rbb^{N\times N}$ are orthonormal eigenvectors, namely the \textit{graph Fourier basis}, $\Lambda = \diag(\lambda_1, \ldots, \lambda_N)$ with $\lambda_1 \leq \cdots \leq \lambda_N$. These eigenvalues are also called \textit{frequencies}. Large eigenvalues are called high frequencies and small eigenvalues are called low frequencies.

In additional to $L$, some variants are also commonly used, \eg{} the symmetric normalized Laplacian $L_{\text{sym}} = D^{-1/2} L D^{-1/2} = I-D^{-1/2} A D^{-1/2}$ and the random walk normalized Laplacian $L_{\text{rw}} = D^{-1} L = I - D^{-1} A$.
Note that $L_{\text{rw}}$ and $L_{\text{sym}}$ share the same eigenvalues, which are inside $[0,2]$, and their corresponding eigenvectors satisfy $\bm{u}_{\text{rw}}^i = D^{-1/2} \bm{u}_{\text{sym}}^i$. 

The affinity matrices can be derived from the Laplacians, \eg{} $A_\text{rw} = I - L_\text{rw} = D^{-1} A$, $A_\text{sym} = I-L_\text{sym} = D^{-1/2} A D^{-1/2}$. Their eigenvalues satisfy $\lambda_i(A_\text{rw}) = \lambda_i(A_\text{sym}) = 1- \lambda_i(L_\text{sym}) = 1- \lambda_i(L_\text{rw}) \in [-1,1]$. The affinity matrices are used as aggregation operators in GNNs.

Applying the renormalization trick~\cite{kipf2016classification} to affinity and Laplacian matrices respectively leads to $\hat{A}_\text{sym} = \tilde{D}^{-1/2} \tilde{A} \tilde{D}^{-1/2}, \hat{A}_{\text{rw}} = \tilde{D}^{-1} \tilde{A}$ and $\hat{L}_{\text{sym}} = I - \hat{A}_\text{sym}, \hat{L}_{\text{rw}} = I - \hat{A}_\text{rw}$, 
where $\tilde{A} \equiv A+I$ and $\tilde{D} \equiv D+I$. The renormalized affinity matrix essentially adds a self-loop to each node in the graph

\paragraph{Low-Pass and High-Pass Filters on Graph}

Following the convention of discrete signal processing (DSP), people call a filter low-pass if
it does not significantly affect the content of low-frequency signals but attenuates the magnitude of high-frequency components. Analogously, high-pass filter passes high-frequency signals while reducing low-frequency parts~\cite{sandryhaila2014discrete}\footnote{For an ideal low-pass filter, there exists a cut-off frequency where all contents with larger frequencies will be rejected and those with smaller frequencies will be retained. The opposite holds for an ideal high-pass filter~\cite{shenoi2005introduction, sandryhaila2014discrete, ekambaram2014graph}. However, for graph neural networks, we do not have such ideal filters. Therefore, the definitions of low-pass and high-pass filters do not have such cut-off frequency, which are a little bit different from the conventional definitions.}.

Suppose we have a graph without non-trivial bipartite graph as connected component, since the maximum eigenvalue of the normalized Laplacian is $2$ if and only if the graph contains a non-trivial bipartite subgraph~\cite{chung1997spectral} and $\lambda_i(A_\text{rw}) = 1- \lambda_i(L_\text{rw}), \; \lambda_i(A_\text{sym}) = 1 - \lambda_i(L_\text{sym})$, the aggregation operation, \eg{} mean pooling, will attenuate the magnitude of high-frequency components more than the low-frequency parts~\cite{maehara2019revisiting}. Therefore, multiplying the normalized affinity matrix acts as a low-pass filter. In addition, the low-pass filtering effect will be provably boosted in renormalized affinity matrices by adding self-loops~\cite{maehara2019revisiting, hamilton2020graph, luan2022complete}.

Generally, the Laplacian matrices ($L_\text{sym}$, $L_\text{rw}$, $\hat{L}_\text{sym}$, $\hat{L}_\text{rw}$) can be regarded as HP filters~\cite{ekambaram2013critically, ekambaram2014graph} and affinity matrices ($A_\text{sym}$, $A_\text{rw}$, $\hat{A}_\text{sym}$, $\hat{A}_\text{rw}$) can be treated as LP filters~\cite{maehara2019revisiting, hamilton2020graph}. In this paper, the LP and original (non-filtered) feature matrices are represented as $H$ and $X$. 

\paragraph{Smoothness, Dirichlet Energy and Frequency} A graph signal $\bm{x}$ is smooth if the nodes connected by the edges (with large edge weights) in the graph tend to have similar signal values~\cite{dong2016learning}. Dirichlet energy is often used to measure the smoothness of graph signals and it is defined as follows\footnote{To be consistent the previous definition, we do not consider edge weights here},
\begin{align*}
E_D^\mathcal{G}(\bm{x}) = \bm{x}^T L \bm{x} = \frac{1}{2}\sum\limits_{e_{ij} \in \mathcal{E}} ({x}_{i} - {x}_{j})^2
\end{align*}
It measures the total pairwise distance between connected nodes. $E_D^\mathcal{G}(\bm{x})$ can also be written as 
$$\bm{x}^T L \bm{x} = \sum\limits_i \lambda_i (\bm{u_i}^T \bm{x})^T \bm{u_i}^T \bm{x} = \sum\limits_i \lambda_i \norm{\bm{u_i}^T \bm{x}}_2^2 $$ where $\bm{u}_i^T \bm{x}$ is the component of $\bm{x}$ in the direction of $\bm{u}_i$. The frequency $\lambda_i$ before $\norm{\bm{u}_i^T \bm{x}}_2^2$ can be considered as a scalar weight. 

From the above equation we can see that, a smooth signal
(i.e., $E_D^\mathcal{G}(\bm{x})$ is small) indicates that most of the contents contained in $\bm{x}$ are low-frequency components\footnote{To be more specific, the contents with large $\norm{\bm{u}_i^T \bm{x}}_2^2$ will correspond to small $\lambda_i$, and the contents with small $\norm{\bm{u}_i^T \bm{x}}_2^2$ will correspond to large $\lambda_i$. Otherwise, $E_D^\mathcal{G}(\bm{x})$ will be large.}, and will be mostly preserved after low-pass filtering (aggregation), and vice versa. Therefore, it is commonly stated that low-pass filter mainly captures the smooth components of the signals (neighborhood similarity), while high-pass filter mainly extracts the non-smooth ones (neighborhood dissimilarity).

\subsection{Neural Message Passing Framework}
\label{sec:neural_message_passing}
Most graph models are built on neural message passing (MP) framework~\cite{gilmer2017neural}, which is an iterative learning process, where the embedding of each node is updated according to the aggregated local neighborhood information. The mechanism can be expressed as follows, 
\begin{equation}
\label{eq:message_passing}
\begin{aligned}
&\mathbf{h}_u^{(k+1)} =\operatorname{UPDATE}^{(k)}\left(\mathbf{h}_u^{(k)}, \mathbf{m}_{\mathcal{N}_u}^{(k)}\right), \\
&\mathbf{m}_{\mathcal{N}_u}^{(k)} = \operatorname{AGGREGATE}^{(k)} \left(\mathbf{h}_u^{(k)}, \mathbf{h}_v^{(k)}, e_{uv} \Large| v \in \mathcal{N}_u\right)
\end{aligned}
\end{equation}

where $\mathbf{h}_u^{(k)}$ is the hidden representation of $u \in \mathcal{V}$ at layer $k$; UPDATE (\eg{} mean, sum, concatenation, skip connection, \etc{}) and AGGREGATE (\eg{} mean, sum, max pooling, LSTM, \etc{}) are arbitrary differentiable functions~\cite{hamilton2020graph}; $\mathbf{m}_{\mathcal{N}(u)}^{(k)}$ is the aggregated message from the neighborhood of $u$ at layer $k$. 

\subsection{Graph-aware Models and Graph-agnostic Models} 
\label{sec:graph_aware_and_agnostic_models}
A neural network that includes the feature aggregation step according to graph structure is called a graph-aware model, \eg{} GCN~\cite{kipf2016classification}, SGC~\cite{wu2019simplifying}; 
and a network that does not use graph structure information is called a graph-agnostic model, such as MLP-2 (Multi-Layer Perceptron with 2 layers) and MLP-1. The graph-aware model is always coupled with a graph-agnostic model because, when we remove the aggregation step in graph-aware model, it becomes exactly the same as its coupled graph-agnostic model,
\eg{} GCN is coupled with MLP-2 and SGC-1 is coupled with MLP-1 as shown below:
\begin{equation}
\begin{aligned}
    \label{eq:gcn_original}
   &\textbf{GCN: } \text{Softmax} (\hat{A}_\text{sym} \; \text{ReLU} (\hat{A}_\text{sym} {X} W_0 ) \; W_1 ),\\
   &\textbf{MLP-2: } \text{Softmax} (  \text{ReLU} (  {X} W_0 ) \; W_1 )\\
   &\textbf{SGC-1: } \text{Softmax} (\hat{A}_\text{sym}  {X} W_0 ), 
   \\
   &\textbf{MLP-1: } \text{Softmax} ( {X} W_0 )
   \end{aligned}
\end{equation}
where $W_0 \in \Rbb^{F_0 \times F_1}$ and $W_1 \in \Rbb^{F_1\times O}$ are learnable parameter matrices. 

To determine the likelihood of the graph-aware model outperforming its coupled graph-agnostic counterpart before training them (\ie{} assessing whether the aggregation step based on graph structure aids in node classification), numerous homophily metrics have been proposed and the most commonly used ones will be introduced in Section~\ref{sec:homophily_metrics_beyond}.

\subsection{Heterogeneous Graph (HetG) and Meta-path}
\paragraph{Heterogeneous Graphs} A Heterogeneous Graph (HetG) is a graph with multiple types of nodes or edges. Specifically, a HetG is defined as $\mathcal{G}=\{\mathcal{V}, \mathcal{E}, \mathcal{A}, \mathcal{R}, \phi, \psi\}$, where $\mathcal{V}$ and $\mathcal{E}$ represent the node and edge sets, respectively; $\mathcal{A}$ denotes the set of node types and $\mathcal{R}$ denotes the set of edge types; $\phi: \mathcal{V} \rightarrow \mathcal{A}$ is the node type mapping function, $\psi: \mathcal{E} \rightarrow \mathcal{R}$ is the edge type mapping function and they satisfy 
$|\mathcal{A}|+|\mathcal{R}|>2$. 
When $|\mathcal{A}|=|\mathcal{R}|=1,\ \mathcal{G}$ refers to a homogeneous graph, which only has one type of nodes and edges.
\paragraph{Meta-path} A meta-path of length $n$ is denoted as $\mathcal{P} \triangleq A_1 \xrightarrow{R_1}$ $A_2 \xrightarrow{R_2} \cdots \xrightarrow{R_n} A_{n+1}$ (or $A_1 A_2 \cdots A_{n+1}$), where $A_i \in \mathcal{A}$ and $R_i \in \mathcal{R}$, describing a composite relation $R=R_1 R_2 \cdots R_n$ between node types $A_1$ and $A_{n+1}$. 
For example, "author $\xrightarrow{\text{write}}$ paper $\xrightarrow{\text{written}}$ author"
is a meta-path with length two in an academic heterogeneous graph which indicates the co-author relationship, where $\mathcal{A} = \{ \text{author, paper} \},
\mathcal{R}=\{\text{write, written}\}$. The set of meta-paths is denoted as $\mathcal{M}$

\paragraph{Meta-Path Induced Subgraph} Given a meta-path, two nodes of $\cal{G}$ 
are considered adjacent in the induced subgraph if they are connected in the original graph and their node types are connected by the meta-path~\cite{hu2021ogb}.
To be more specific, given a meta-path $\mathcal{P}=A_1 A_2 \cdots A_{n+1}$, we can construct the corresponding meta-path induced subgraph $\mathcal{G_P}$, 
which satisfies that the edge $e_{uv}^{\mathcal{P}} \in \mathcal{E_P}$ connects nodes $u,v$ (1) whose node types are $A_1$ and $A_{n+1}$ separately; (2) there exists at least one length-$n$ path between $u$ and $v$ following the meta-path $\mathcal{P}$ in the original graph $\mathcal{G}$. 
In the above example, two authors are considered to be connected if they are the co-authors of at least one paper, \ie{} there exists a co-author induced subgraph.
If $A_1=A_{n+1}$, we have an induced homogeneous graph $\mathcal{G_P}$ built on a meta-path with the end nodes of the same type.

\paragraph{Meta-Path Based Heterogeneous Graph Neural Networks (HetGNNs) and Heterophily}
Metapath-based HetGNNs propagate and aggregate neighbor features using hand-crafted or automatically selected meta-paths~\cite{dong2017metapath2vec, yun2019graph, wang2019heterogeneous, fu2020magnn}. Current research on HetGNNs typically focuses on some HetGs with homophilic meta-path induced subgraphs. However, in many situations, HetGs with heterophily exist~\cite{li2023hetero,guo2023homophily}. For example, in a network of movie associations, films featuring the same actor typically belong to distinct categories. Guo \etal~\cite{guo2023homophily} empirically found that the performance of HGNN is related to the homophily levels of the meta-path induced subgraphs. More specifically, HetGNNs tend to perform worse on HetGs with heterophilic meta-path subgraphs than on HetGs with homophilic meta-path subgraphs.
We will summarize several homophily metrics defined on HetGs in Section~\ref{sec:metrics_heterogeneous_graphs} and discuss the heterophily problem for HetGNNs in Section~\ref{sec:benchmark_heterogeneous_graphs}.

\section{Homophily Metrics and Beyond}
\label{sec:homophily_metrics_beyond}
To measure whether graph-aware models can outperform their coupled graph-agnostic counterparts without training, many homophily metrics have been proposed. In this section, we introduce the most commonly used ones for homogeneous and HetGs.

\subsection{Homophily Metrics on Homogeneous Graphs}
\label{sec:metrics_on_homogeneous_graphs}
There are mainly four ways to define the metrics that describe the relations among node labels, features and graph structures. 

\paragraph{Graph-Label Consistency} Four commonly used homophily metrics based on the consistency between node labels and graph structures are edge homophily~\cite{abu2019mixhop,zhu2020beyond}, node homophily~\cite{pei2020geom}, class homophily~\cite{lim2021new} and adjusted homophily~\cite{platonov2023characterizing}
defined as follows:
\begin{equation}
\begin{aligned}
\label{eq:homo_metrics_graph_label_consistency}
& \textit{H}_\text{edge}(\mathcal{G}) = \frac{\big|\{e_{uv} \mid e_{uv}\in \mathcal{E}, y_{u} = y_{v}\}\big|}{|\mathcal{E}|}; \\
& \resizebox{1\hsize}{!}{$\textit{H}_\text{node}(\mathcal{G}) = \frac{1}{|\mathcal{V}|} \sum_{v \in \mathcal{V}}  \textit{H}_\text{node}^v= \frac{1}{|\mathcal{V}|} \sum_{v \in \mathcal{V}} \frac{\big|\{u \mid u \in \mathcal{N}_v, y_{u} = y_{v}\} \big|}{d_v};$} \\
& \textit{H}_\text{class}(\mathcal{G}) \!=\! \frac{1}{C\!-\!1} \sum_{k=1}^{C}\bigg[h_{k}
    \!-\! \frac{\big|\{v \!\mid\! Y_{v,k} \!=\! 1 \}\big|}{N}\bigg]_{+},\\
& \text{with } h_{k}\! =\! \frac{\sum_{v \in \mathcal{V}, Y_{v,k}\! =\! 1} \big|\{u \!\mid\!  u \in \mathcal{N}_v, y_{u}\!=\! y_{v}\}\big| }{\sum_{v \in \{v|Y_{v,k}=1\}} d_{v}};  \\
& \textit{H}_\text{adj}(\mathcal{G}) = \frac{\textit{H}_\text{edge} - \sum_{c=1}^C \bar{p}_c^2}{1-\sum_{c=1}^C \bar{p}_c^2}, \text{with } \bar{p}_c = \frac{\sum_{v: y_v=c} d_v}{2|\mathcal{E}|} \\
\end{aligned}
\end{equation}
where $\textit{H}_\text{node}^v$ is the local homophily value for node $v$; $[a]_{+}=\max (a, 0)$, $h_{k}$ is the class-wise homophily metric~\cite{lim2021new}.

Note that $H_\text{edge}(\mathcal{G})$ measures the proportion of edges that connect two nodes in the same class; $H_\text{node}(\mathcal{G})$ evaluates the average proportion of edge-label consistency of all nodes; $H_\text{class}(\mathcal{G})$ is designed to reduce sensitivity to class imbalance, which can make $H_\text{edge}(\mathcal{G})$ misleadingly large; $\textit{H}_\text{adj}(\mathcal{G})$ is constructed to satisfy maximal agreement and constant baseline properties. The above definitions are all based on the \textbf{linear feature-independent graph-label consistency}. The inconsistency relation indicated by a small metric value implies that the graph structure has a negative effect on the performance of GNNs. 

\paragraph{Similarity Based Metrics} Generalized edge homophily~\cite{jin2022raw} and aggregation homophily~\cite{luan2022revisiting} leverage similarity functions to define the metrics,
\begin{equation}
\begin{aligned}
\label{eq:homo_metrics_graph_similarity_based}
& \textit{H}_\text{GE} (\mathcal{G}) = \frac{\sum\limits_{(i,j) \in \mathcal{E}} \text{cos}(\bm{x}_{i}, \bm{x}_{j})}{|\mathcal{E}|}; \\
& \resizebox{1\hsize}{!}{$\textit{H}_{\text{agg}}(\mathcal{G}) =  \frac{1}{\left| \mathcal{V} \right|} \times \left| \left\{v \,\big| \, \mathrm{Mean}_u  \big( \{S(\hat{{A}},Y)_{v,u}^{y_{u} =y_{v}} \}\big) \geq  \mathrm{Mean}_u\big(\{S(\hat{{A}},Y)_{v,u}^{y_{u} \neq y_{v}}  \} \big) \right\} \right|;$} \\
\end{aligned}
\end{equation}
where $\mathrm{Mean}_u\left(\{\cdot\}\right)$ takes the average over $u$ of a given multiset of values or variables and $S(\hat{{A}},Y)=\hat{{A}}Y(\hat{{A}}Y)^\top$ is the post-aggregation node similarity matrix. These two metrics are feature-dependent.

$\textit{H}_\text{GE} (\mathcal{G})$ generalizes $\textit{H}_\text{edge}(\mathcal{G})$ to the the cosine similarities between node features; $\textit{H}_{\text{agg}}(\mathcal{G})$ measures the proportion of nodes $v\in\mathcal{V}$ whose average $S(\hat{{A}},Y)$ weights on the set of nodes in the same class (including $v$) is larger than that in other classes. They are both feature-dependent metrics.

\paragraph{Neighborhood Identifiability/Informativeness}  Label informativeness ~\cite{platonov2023characterizing} and neighborhood identifiability~\cite{chen2023exploiting} leverage the neighborhood distribution instead of pairwise comparison to define the metrics,

\begin{equation}
\begin{aligned}
\label{eq:homo_metrics_graph_identifiable_informative}
& \resizebox{1\hsize}{!}{$\text{LI} = -\frac{\sum_{c_1, c_2} p_{c_1, c_2 } \log \frac{p_{c_1, c_2}}{\bar{p}_{c_1} \bar{p}_{c_2}}}{\sum_c \bar{p}_c \log \bar{p}_c} = 2-\frac{\sum_{c_1, c_2} p_{c_1, c_2} \log p_{c_1, c_2}}{\sum_c \bar{p}_c \log \bar{p}_c};$}\\
&\textit{H}_{\text {neighbor }}(\mathcal{G}) = \sum_{k=1}^C \frac{n_k}{N} \textit{H}_{\text {neighbor }}^k,\\
&\text{with } \textit{H}_{\text {neighbor }}^k = \frac{-\sum_{i=1}^C \tilde{\sigma}_i^k \log \left( \tilde{\sigma}_i^k \right)}{\log (C)}.
\end{aligned}
\end{equation}
where $p_{c_1, c_2} = \sum_{(u, v) \in \mathcal{E}} \frac{\bm{1}_{\left\{y_u = c_1, y_v = c_2 \right\} } }{2|\mathcal{E}|}, c_1, c_2 \in \{1,\dots,C\}$, $\bar{p}_c$ is the same as in Equation \eqref{eq:homo_metrics_graph_label_consistency};
$A_N^k \in \mathbb{R}^{n_k \times C}$ is a class-level neighborhood label distribution matrix for each class $k$, where $k=1, \ldots, C$ for different classes and $n_k$ indicates the number of nodes with the label $k$, $(A_N^k)_{i,c}$ is the proportion of the neighbors of node $i$ belonging to class $c$, $\sigma_1^k, \sigma_2^k, \ldots, \sigma_C^k$ denote singular values of $A_{\mathcal{N}}^k$, they are normalized to $\sum_{c=1}^C \tilde{\sigma}_c^k=1$, $c=1, \ldots, C$ is the index of singular values.

$\text{LI}$ is to characterize different connectivity patterns by measuring the informativeness of the label of a neighbor for the label of a node; $\textit{H}_{\text {neighbor}}(\mathcal{G})$ is a weighted sum of $\textit{H}_{\text {neighbor}}^k(\mathcal{G})$, quantifying neighborhood identifiability through the entropy of the singular value distribution of $A_N^k$, which is a generalization of the von Neumann entropy in quantum statistical mechanics~\cite{bengtsson2017geometry} that measures the pureness/information of a quantum-mechanical system. This metric effectively measures the complexity/randomness of neighborhood distributions by indicating the number of vectors (or neighbor patterns) necessary to sufficiently describe the neighborhood label distribution matrix.
\paragraph{Hypothesis Testing Based Performance Metrics} Luan \etal~\cite{luan2024graph} proposed classifier-based performance metric (CPM)\footnote{Luan \etal~\cite{luan2023we} also conducted hypothesis testing to find out when to use GNNs for node classification, but what they tested was the differences between connected nodes and unconnected nodes instead of intra- and inter-class nodes and they did not propose a metric based on hypothesis testing.}, which uses the p-value of hypothesis testing as the metric to measure the node distinguishability of the aggregated features compared with the original features.

They first randomly sample 500 labeled nodes from $\mathcal{V}$ and splits them into 60\%/40\% as "training" and "test" data. The original features $X$ and aggregated features $H$ of the sampled training and test nodes can be calculated and are then fed into a given classifier. The prediction accuracy on the test nodes will be computed directly with the feedforward method. This process will be repeated 100 times to get 100 samples of prediction accuracy for $X$ and $H = \hat{A}X$. Then, for the given classifier, they compute the p-value of the following hypothesis testing, 
\begin{equation}
\begin{aligned}
\label{eq:definition_homophily_metrics}
&\text{H}_0: \text{Acc}(\text{Classifier}(H)) \geq \text{Acc}(\text{Classifier}(X)); \\ &\text{H}_1: \text{Acc}(\text{Classifier}(H)) < \text{Acc}(\text{Classifier}(X))
\end{aligned}
\end{equation}
The p-value can provide a statistical threshold value to indicate whether $H$ is significantly better than $X$ for node classification. To capture the feature-based linear or non-linear information efficiently, Luan \etal choose Gaussian Naïve Bayes (GNB)~\cite{hastie2009elements} and Kernel Regression (KR) with Neural Network Gaussian Process (NNGP)~\cite{lee2018deep,arora2019exact, garriga2018deep,matthews2018gaussian} as the classifiers, which do not require iterative training.

Overall, $\textit{H}_\text{adj}$ can assume negative values, while other metrics all fall within the range of $[0,1]$. Except for $\textit{H}_{\text {neighbor}}(\mathcal{G})$, where a smaller value indicates more identifiable, the other metrics with a value closer to $1$ indicate strong homophily and suggest that the connected nodes tend to share the same label, implying that graph-aware models are more likely to outperform their coupled graph-agnostic model, and vice versa. 

$\textit{H}_\text{edge}, \textit{H}_\text{node}, \textit{H}_\text{class}, \textit{H}_\text{adj}$ and $\text{LI}$ are linear feature-independent metrics. LI and ${H}_{\text{neighbor}}(\mathcal{G})$ are nonlinear feature-independent metrics. $\textit{H}_\text{GE}$ and $\textit{H}_{\text{agg}}$ are feature-dependent and measure the linear similarity between nodes. CPM is the first metric that can capture nonlinear feature-dependent information and provide accurate threshold values to indicate the superiority of graph-aware models. In Section~\ref{sec:comparison_metrics_synthetic_graphs}, we will introduce the approach for the comparison of the above metrics by synthetic graphs with different generation methods.

\subsection{Homophily Metrics on Heterogeneous Graphs}
\label{sec:metrics_heterogeneous_graphs}

\paragraph{Heterogeneous Graph Homophily Ratio (HHR)} Guo \etal~\cite{guo2023homophily} propose a meta-path induced metric to measure the homophily degree of a HetG. The Meta-Path Subgraph Homophily Ratio (MSHR) is defined as follows,
$$
\textit{H}\left(\mathcal{G}_{\mathcal{P}}\right) = \frac{\sum_{\left(v_i, v_j\right) \in \mathcal{E}_{\mathcal{P}}} \bm{1}_{\left(y_i=y_j\right)}}{\left|\mathcal{E}_{\mathcal{P}}\right|},
$$
where $\bm{1}_{(\cdot)}$ is the indicator function; 
$y_i$ is the label of node $v_i$ and $|\mathcal{E_P}|$ is the size of the edge set.

Given a HetG and meta-path set $\mathcal{M}$, 
the heterogeneous graph homophily ratio (HHR) of HetG is defined by:
$$
\operatorname{MH}(\mathcal{G}) = \max \left(\left\{\textit{H} \left(\mathcal{G}_{\mathcal{P}}\right) \mid \mathcal{P} \in \mathcal{M} \right\}\right)
$$
It calculates the maximum homophily ratio over subgraphs of meta-path, which is the maximum potential homophily ratio of a given graph.

\paragraph{Meta-Path Based Label Homophily (MLH) and Meta-Path Based Dirichlet Energy (MDE)}
Li \etal~\cite{li2023hetero} introduce meta-path-based label homophily (MLH) and Dirichlet energy (MDE) to measure the homophily of labels and features. For a given node type $ A \in \mathcal{A}$:
$$
\begin{aligned}
& \operatorname{MLH}_A (\mathcal{G}) = \frac{1}{|\mathcal{P}|} \sum_{\mathcal{P} \in \mathscr{P}_A} \textit{H}(\mathcal{G_P}), \\
& \operatorname{MDE}_A(\mathcal{G})=\frac{1}{|\mathcal{P}|} \sum_{\mathcal{P} \in \mathscr{P}_A} E_D \left(\mathcal{G}_{\mathcal{P}}\right),
\end{aligned}
$$
where $\mathscr{P}_A = \left\{A A_i  A \mid A_i \in \mathcal{A}\right\}$ 
represents the set of all possible length-2 meta-paths starting from node type $A$ and ending at node type $A$.
Both H and $E_D$ can be either edge-level or node-level metrics. Typically, a larger MLH or a smaller MDE suggests a stronger homophily of a heterogeneous graph.


\section{Benchmark and Evaluation}
\label{sec:benchmark_evaluation}
\begin{table*}[htbp]
  \centering
  \caption{Categorization of Homogeneous Benchmark Datasets for Heterophily}
  \resizebox{1\hsize}{!}{
    \begin{tabular}{c|c|c|ccccccp{5.145em}|cccc|c}
    \toprule
    \toprule
    \multicolumn{2}{p{11.21em}|}{Categories} & {Datasets} & {\#Nodes} & {\#Edges} & {\#Feature Dim} & {\#Class} & {$\textit{H}_\text{edge}$} & {$\textit{H}_\text{node}$} & Eval Metric  &{GCN} &{MLP-2} & {SGC-1} & {MLP-1} & {Literature} \\
    \midrule
          &       & {Cornell} & 183   & 295   & 1,703 & 5     & 0.2983 & 0.2001 & Accuracy & 82.46 $\pm$ 3.11 & \cellcolor[rgb]{ .439,  .678,  .278}91.30 $\pm$ 0.70 & 70.98 $\pm$ 8.39 & \cellcolor[rgb]{ .439,  .678,  .278}93.77 $\pm$ 3.34 & \cite{pei2020geom} \\
          &       & {Wisconsin} & 251   & 499   & 1,703 & 5     & 0.1703 & 0.0991 & Accuracy & 75.5 $\pm$ 2.92 & \cellcolor[rgb]{ .439,  .678,  .278}93.87 $\pm$ 3.33 & 70.38 $\pm$ 2.85 & \cellcolor[rgb]{ .439,  .678,  .278}93.87 $\pm$ 3.33 & \cite{pei2020geom} \\
          &       & {Texas} & 183   & 309   & 1,703 & 5     & 0.0615 & 0.0555 & Accuracy & 83.11 $\pm$ 3.2 & \cellcolor[rgb]{ .439,  .678,  .278}92.26 $\pm$ 0.71 & 83.28 $\pm$ 5.43 & \cellcolor[rgb]{ .439,  .678,  .278}93.77 $\pm$ 3.34 & \cite{pei2020geom} \\
          &       & {Film} & 7,600 & 33,544 & 931   & 5     & 0.2163 & 0.2023 & Accuracy & 35.51 $\pm$ 0.99 & \cellcolor[rgb]{ .439,  .678,  .278}38.58 $\pm$ 0.25 & 25.26 $\pm$ 1.18 & \cellcolor[rgb]{ .439,  .678,  .278}34.53 $\pm$ 1.48 & \cite{pei2020geom} \\
          & {Malignant} & {Deezer-Europe} & 28,281 & 92,752 & 31,241 & 2     & 0.5251 & 0.5299 & Accuracy & 62.23 $\pm$ 0.53 & \cellcolor[rgb]{ .439,  .678,  .278}66.55 $\pm$ 0.72 & 61.63 $\pm$ 0.25 & \cellcolor[rgb]{ .439,  .678,  .278}63.14 $\pm$ 0.41 & \cite{lim2021new} \\
          &       &  genius & 421,961 & 984,979 & 12    & 2     & 0.6176 & 0.0985 & Accuracy & 83.26 $\pm$ 0.14 & \cellcolor[rgb]{ .439,  .678,  .278}86.62 $\pm$ 0.08 & 82.31 $\pm$ 0.45 & \cellcolor[rgb]{ .439,  .678,  .278}86.48 $\pm$ 0.11 & \cite{lim2021large} \\
          &       &  roman-empire  & 22,662 & 32,927 & 300   & 18    & 0.0469 & 0.046 & Accuracy & 48.92 $\pm$ 0.46 & \cellcolor[rgb]{ .439,  .678,  .278}66.04 $\pm$ 0.71 & 44.60 $\pm$ 0.52 & \cellcolor[rgb]{ .439,  .678,  .278}64.12 $\pm$ 0.61 & \cite{platonov2022critical} \\
          &       & BlogCatalog & 5,196 & 171,743 & 8,189 & 6     & 0.4011 & 0.3914 & \multicolumn{1}{c|}{Accuracy} & 79.67 $\pm$ 1.06 & \cellcolor[rgb]{ .439,  .678,  .278}92.97 $\pm$ 0.89 & 71.07 $\pm$ 1.15 & \cellcolor[rgb]{ .439,  .678,  .278}91.86 $\pm$ 0.93 & \cite{zhou2024opengsl} \\
          &       & Flickr & 7,575 & 239,738 & 12,047 & 9     & 0.2386 & 0.2434 & \multicolumn{1}{c|}{Accuracy} & 71.38 $\pm$ 1.00 & \cellcolor[rgb]{ .439,  .678,  .278}90.24 $\pm$ 0.96 & 60.10 $\pm$ 1.21 & \cellcolor[rgb]{ .439,  .678,  .278}89.91 $\pm$ 0.97 & \cite{zhou2024opengsl} \\
          &       & BGP   & 63,977 & 174,803 & 287   & 8     & 0.2545 & 0.083 & \multicolumn{1}{c|}{Accuracy} & 62.56 $\pm$ 0.94 & \cellcolor[rgb]{ .439,  .678,  .278}65.56 $\pm$ 0.55 & 61.74 $\pm$ 0.73 & \cellcolor[rgb]{ .439,  .678,  .278}64.67 $\pm$ 0.81 & \cite{sun2022beyond} \\
\cmidrule{2-15}    {Heterophilic } &       & {Chameleon} & 2,277 & 36,101 & 2,325 & 5     & 0.2339 & 0.2467 & Accuracy & \cellcolor[rgb]{ .439,  .678,  .278}64.18 $\pm$ 2.62 & 46.72 $\pm$ 0.46 & \cellcolor[rgb]{ .439,  .678,  .278}64.86 $\pm$ 1.81 & 45.01 $\pm$ 1.58 & \cite{rozemberczki2021multi} \\
    {Graphs } &       & {Squirrel} & 5,201 & 217,073 & 2,089 & 5     & 0.2234 & 0.2154 & Accuracy & \cellcolor[rgb]{ .439,  .678,  .278}44.76 $\pm$ 1.39 & 31.28 $\pm$ 0.27 & \cellcolor[rgb]{ .439,  .678,  .278}47.62 $\pm$ 1.27 & 29.17 $\pm$ 1.46 & \cite{rozemberczki2021multi} \\
          & {Benign} & Chameleon-filtered & 890   & 8,854 & 2,325 & 5     & 0.2361 & 0.2441 & \multicolumn{1}{c|}{Accuracy} & \cellcolor[rgb]{ .439,  .678,  .278}41.46 $\pm$ 3.42 & 38.06 $\pm$ 3.98 & \cellcolor[rgb]{ .439,  .678,  .278}44.00 $\pm$ 3.10 & 35.72 $\pm$ 2.23 & \cite{platonov2022critical} \\
          &       &  arXiv-year & 169,343 & 1,166,243 & 128   & 5     & 0.2218 & 0.2778 & ROC AUC & \cellcolor[rgb]{ .439,  .678,  .278}40 $\pm$ 0.26 & 36.36 $\pm$ 0.23 & \cellcolor[rgb]{ .439,  .678,  .278}35.58 $\pm$ 0.22  & 34.11 $\pm$ 0.17 & \cite{lim2021large} \\
          &       &  amazon-ratings & 24,492 & 93,050 & 300   & 5     & 0.3804 & 0.3757 & Accuracy & \cellcolor[rgb]{ .439,  .678,  .278}50.05 $\pm$ 0.67 & 49.55 $\pm$ 0.81 & \cellcolor[rgb]{ .439,  .678,  .278}40.69 $\pm$ 0.42 & 38.60 $\pm$ 0.41 & \cite{platonov2022critical} \\
          &       & Wiki-cooc & 10,000 & 2,243,042 & 100   & 5     & 0.3435 & 0.175 & \multicolumn{1}{c|}{Accuracy} & \cellcolor[rgb]{ .439,  .678,  .278}95.40 $\pm$ 0.41 & 89.38 $\pm$ 0.42 & \cellcolor[rgb]{ .439,  .678,  .278}72.38 $\pm$ 0.78 & 48.86 $\pm$ 0.37 & \cite{zhou2024opengsl} \\
\cmidrule{2-15}          &       & Squirrel-filtered & 2,223 & 46,998 & 2,089 & 5     & 0.2072 & 0.1905 & Accuracy & 37.33 $\pm$ 1.88 & \cellcolor[rgb]{ .439,  .678,  .278}38.30 $\pm$ 1.22 & \cellcolor[rgb]{ .439,  .678,  .278}37.54 $\pm$ 2.13 & 30.14 $\pm$ 1.53 & \cite{platonov2022critical} \\
          &       & Penn94 & 41,554 & 1,362,229 & 5     & 2     & 0.4704 & 0.4828 & Accuracy & \cellcolor[rgb]{ .439,  .678,  .278}82.08 $\pm$ 0.31 & 74.68 $\pm$ 0.28 & 67.06 $\pm$ 0.19 & \cellcolor[rgb]{ .439,  .678,  .278}73.72 $\pm$ 0.5 & \cite{lim2021large} \\
          & {Ambiguous} &  pokec & 1,632,803 & 30,622,564 & 65    & 2     & 0.4449 &  0.3931    & Accuracy & \cellcolor[rgb]{ .439,  .678,  .278}70.3 $\pm$ 0.1 & 62.13 $\pm$ 0.1 & 52.88 $\pm$ 0.64 & \cellcolor[rgb]{ .439,  .678,  .278}59.89 $\pm$ 0.11 & \cite{lim2021large} \\
          &       &  snap-patents & 2,923,922 & 13,975,788 & 269   & 5     & 0.073 &  0.1857     & Accuracy & \cellcolor[rgb]{ .439,  .678,  .278}35.8 $\pm$ 0.05 & 31.43 $\pm$ 0.04 & 29.65 $\pm$ 0.04 & \cellcolor[rgb]{ .439,  .678,  .278}30.59 $\pm$ 0.02 & \cite{lim2021large} \\
          &       &  twitch-gamers & 168,114 & 6,797,557 & 7     & 2     & 0.545 & 0.556 & Accuracy & \cellcolor[rgb]{ .439,  .678,  .278}62.33 $\pm$ 0.23 & 60.9 $\pm$ 0.11 & 57.9 $\pm$ 0.18 & \cellcolor[rgb]{ .439,  .678,  .278}59.45 $\pm$ 0.16 & \cite{lim2021large} \\
         \cmidrule{1-15}  &       & {Cora} & 2,708 & 5,429 & 1,433 & 7     & 0.8100 & 0.8252 & Accuracy & \cellcolor[rgb]{ .439,  .678,  .278}87.78 $\pm$ 0.96 & 76.44 $\pm$ 0.30 & \cellcolor[rgb]{ .439,  .678,  .278}85.12 $\pm$ 1.64 & 74.3 $\pm$ 1.27 & \cite{yang2016revisiting} \\
          &       & {CiteSeer} & 3,327 & 4,732 & 3,703 & 6     & 0.7355 & 0.7062 & Accuracy & \cellcolor[rgb]{ .439,  .678,  .278}81.39 $\pm$ 1.23 & 76.25 $\pm$ 0.28 & \cellcolor[rgb]{ .439,  .678,  .278}79.66 $\pm$ 0.75 & 75.51 $\pm$ 1.35 & \cite{yang2016revisiting} \\
   {Homophilic } &       & {PubMed} & 19,717 & 44,338 & 500   & 3     & 0.8024 & 0.7924 & Accuracy & \cellcolor[rgb]{ .439,  .678,  .278}88.9 $\pm$ 0.32 & 86.43 $\pm$ 0.13 & \cellcolor[rgb]{ .439,  .678,  .278}86.5 $\pm$ 0.76 & 86.23 $\pm$ 0.54 & \cite{yang2016revisiting} \\
    {Graphs} &       &  minesweeper  & 10,000 & 39,402 & 7     & 2     & 0.6828 & 0.6829 & ROC AUC & \cellcolor[rgb]{ .439,  .678,  .278}72.34 $\pm$ 0.93 & 50.92 $\pm$ 1.25 & \cellcolor[rgb]{ .439,  .678,  .278}82.04 $\pm$ 0.77 & 50.59 $\pm$ 0.83 & \cite{platonov2022critical} \\
          &       &  tolokers & 11,758 & 519,000 & 10    & 2     & 0.5945 & 0.6344 & ROC AUC & \cellcolor[rgb]{ .439,  .678,  .278}77.44 $\pm$ 1.32 & 74.58 $\pm$ 0.75 & \cellcolor[rgb]{ .439,  .678,  .278}73.80 $\pm$ 1.35 & 71.89 $\pm$ 0.82 & \cite{platonov2022critical} \\
          &       &  questions & 48,921 & 153,540 & 301   & 2     & 0.8396 & 0.898 & ROC AUC & \cellcolor[rgb]{ .439,  .678,  .278}72.72 $\pm$ 1.93 & 69.97 $\pm$ 1.16 & \cellcolor[rgb]{ .439,  .678,  .278}71.06 $\pm$ 0.92 & 70.33 $\pm$ 0.96 & \cite{platonov2022critical} \\
    \bottomrule
    \bottomrule
    \end{tabular}%

    }
  \label{tab:homogeneous_benchmark}%
\end{table*}%

Recently, numerous heterophilic benchmark datasets have been developed to validate the efficacy of heterophily-specific models and various homophily metrics have been designed to help people recognize the challenging datasets, especially for tasks defined on homogeneous graphs~\cite{rozemberczki2021multi, pei2020geom, lim2021new, lim2021large, platonov2022critical}. In Section~\ref{sec:homogeneous_benchmark}, we will summarize the most commonly used homogeneous graph datasets and introduce a novel taxonomy to categorize the heterophilic graphs into benign, malignant and ambiguous groups; in Section~\ref{sec:comparison_metrics_synthetic_graphs}, we outline how to evaluate homophily metrics on synthetic graphs, summarize the graph generation methods and compare the results; in Section~\ref{sec:benchmark_heterogeneous_graphs}, we collect several heterogeneous benchmark datasets, which can help to identify the relation between homophily values and the performance of heterogeneous GNNs.

\subsection{Homogeneous Heterophily Benchmarks}
\label{sec:homogeneous_benchmark}

\subsubsection{Benign, Malignant and Ambiguous Heterophily for Graph Models}

Although it is widely believed that homophily is beneficial for GNN performance while heterophily has a harmful effect on it, some empirical evidence shows that the graph-aware model does not always perform worse than its coupled graph-agnostic model on heterophilic graphs~\cite{luan2021heterophily, ma2021homophily, luan2023we}. This indicates that heterophily is not always malignant~\cite{ma2021homophily,luan2022revisiting} and it is imperative to identify the real challenging subsets of heterophily datasets. 

As stated in Section~\ref{sec:graph_aware_and_agnostic_models} and ~\cite{luan2024graph}, the existing homophily metrics are not a reliable tool to find out the difficult datasets. And currently, the empirical comparison of graph-aware model and its coupled graph-agnostic model is the only feasible way to recognize them. Although quite time-consuming, we have trained the models with fine-tuned hyperparameters~\footnote{Note that hyperparameter tuning is critical for fair model comparison and some previous results are already found unreliable without hyperparameter fine-tunning~\cite{ma2021homophily,luo2024classic}} accordingly and conducted comprehensive comparisons to classify most of the existing benchmark datasets. Based on the results, we classify the heterophilic datasets into benign, malignant and ambiguous heterophily datasets. The real challenges lie in malignant and ambiguous datasets, which should be paid more attention in model evaluation. A model with good performance only on benign heterophilic datasets cannot be validated as an effective method for heterophily problem. To our best knowledge, this is the first attempt to introduce such taxonomy on heterophily datasets.



\paragraph{Results} We collect 27 most commonly used benchmark datasets for heterophily research in recent years~\cite{pei2020geom, rozemberczki2021multi, lim2021new, lim2021large, sun2022beyond, platonov2022critical, zhou2024opengsl}, and run 2 baseline GNNs (GCN, SGC-1) and their coupled graph-agnostic models (MLP-2,MLP-1) with fine-tuned hyperparameters. The dataset statistics and experimental results are shown in Table \ref{tab:homogeneous_benchmark}. We observe that

\begin{itemize}
    \item \textbf{Malignant Heterophily:} \textit{Cornell, Wisconsin, Texas, Film, Deezer-Europe, genius, roman-empire, BlogCatalog, Flickr} and \textit{BGP} are the most challenging heterophilic datasets, where graph-aware models consistently underperform their coupled graph-agnostic models. The harmful heterophilic aggregation occurs on these datasets and new proposed methods should be verified on these datasets to show their effectiveness on heterophily.
    
    \item \textbf{Benign Heterophily:} \textit{Chameleon, Squirrel, Chameleon-filtered, arXiv-year, amazon-ratings} and \textit{Wiki-cooc} are benign heterophily datasets, where the heterophilic aggregation is actually beneficial for GNNs. The test results of heterophily-specific GNNs on this subset of graphs cannot be used as reliable evidence to show the model capacity on heterophilic graphs.
    
    \item  \textbf{Ambiguous Heterophily:} \textit{Squirrel-filtered, Penn94, pokec, snap-patents} and \textit{twitch-gamers} are ambiguous heterophily datasets, where linear and non-linear models have inconsistent comparison results. For example, GCN (non-linear model) outperforms MLP-2 while SGC-1 (linear model) underperforms MLP-1 on \textit{Penn94, pokec, snap-patents} and \textit{twitch-gamers}; while GCN underperforms MLP-2 and SGC-1 outperforms MLP-1 on \textit{Squirrel-filtered}. This indicates that model non-linearity and heterophilic graph structure must have synergy on the performance of GNNs. However, the ambiguity caused by the synergy has no theoretical explanation for now. This could be an interesting and important area of work in the future. This subset of datasets is also considered tough heterophily.
    
    \item \textbf{A Good GNN for Heterophily:} A heterophily-specific GNN is considered a good model if it performs significantly better than baseline models on heterophily datasets, especially on malignant and ambiguous heterophilic graphs, and perform at least as good as baseline models on homophilic graphs. Such standard is proposed because some GNNs are found to sacrifice their capabilities on homophilic graphs to achieve relatively better performance on heterophilic graphs, \eg{}H$_2$GCN~\cite{zhu2020beyond}, LINKX~\cite{lim2021large}, GPRGNN~\cite{chien2021adaptive} and BernNet~\cite{he2021bernnet}.
\end{itemize}

\subsubsection{Model Reassessment}
Serious flaws have been found for model evaluation on the benchmark datasets. Some people point out the problem and conduct comprehensive experiments to reassess the SOTA models.

Luo \etal~\cite{luo2024classic} find that, with slight hyperparameter tuning, classic GNNs can outperform SOTA graph transformers (GTs)  on 17 out of 18 datasets, across homophilic and heterophilic graphs. This indicates that the widely accepted superiority of GTs over GNNs may have been overclaimed and should be overturned, due to suboptimal hyperparameter configurations in GNN evaluations.

Liao \etal~\cite{liao2024benchmarking} classify the spectral GNNs by the filters and divede them into three categories, fixed filter, variable filter and filterbank. Then, they reassess the performance of different spectral GNNs on heterophilic graphs and found that (1) fixed filters typically fail under heterophily, while variable filters and filter bank models excel under heterophily; (2) the prediction accuracy of high-degree nodes is usually significantly lower under heterophily and the hypothesis that "nodes with higher degrees are naturally favored by GNNs" no longer holds. In fact, the degree-wise bias is more sensitive, but not necessarily beneficial, to high-degree nodes with regard to varying graph conditions.

There are still plenty of works need to be done in the future for fair and reliable model evaluation on heterophily benchmark datasets.
\subsection{Comparison of Homophily Metrics on Synthetic Graphs}
\label{sec:comparison_metrics_synthetic_graphs}
Homophily metrics are proposed to identify the tough heterophily datasets, and people usually verify and compare the effectiveness of the metrics by synthetic graphs. In Section~\ref{sec:synthetic_graph_generation}, we summarize three most widely used generation methods for synthetic graphs; in Section~\ref{sec:homophily_metric_evaluation_comparison}, we present the evaluation methods for homophily metrics and compare the results.

\subsubsection{Generation Methods for Synthetic Graphs}
\label{sec:synthetic_graph_generation}

There are mainly three ways to generate synthetic graphs for the evaluation of homophily metrics.

\paragraph{Regular Graph (RG)}

\begin{figure}[ht]
    \centering
     {
     \subfloat[RG: Model Curves]{
     \captionsetup{justification = centering}
     \includegraphics[width=0.45\textwidth]{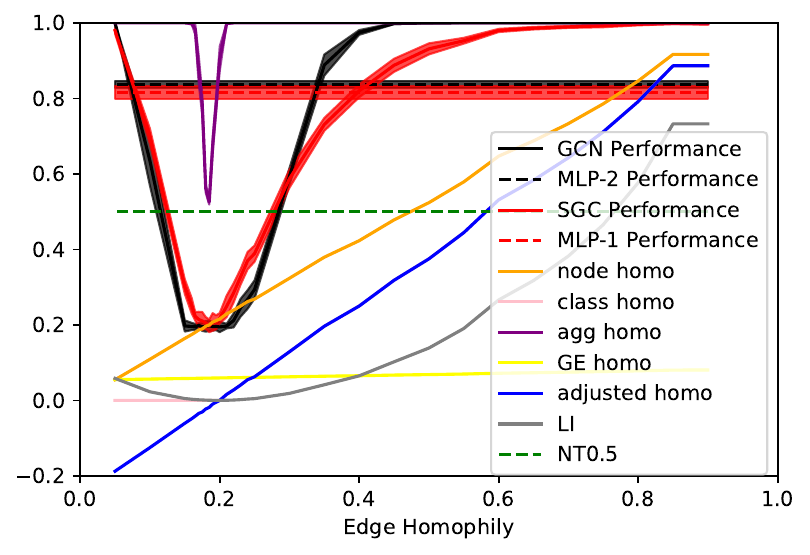}
     }
     \subfloat[RG: Metric Curves]{
     \captionsetup{justification = centering}
     \includegraphics[width=0.45\textwidth]{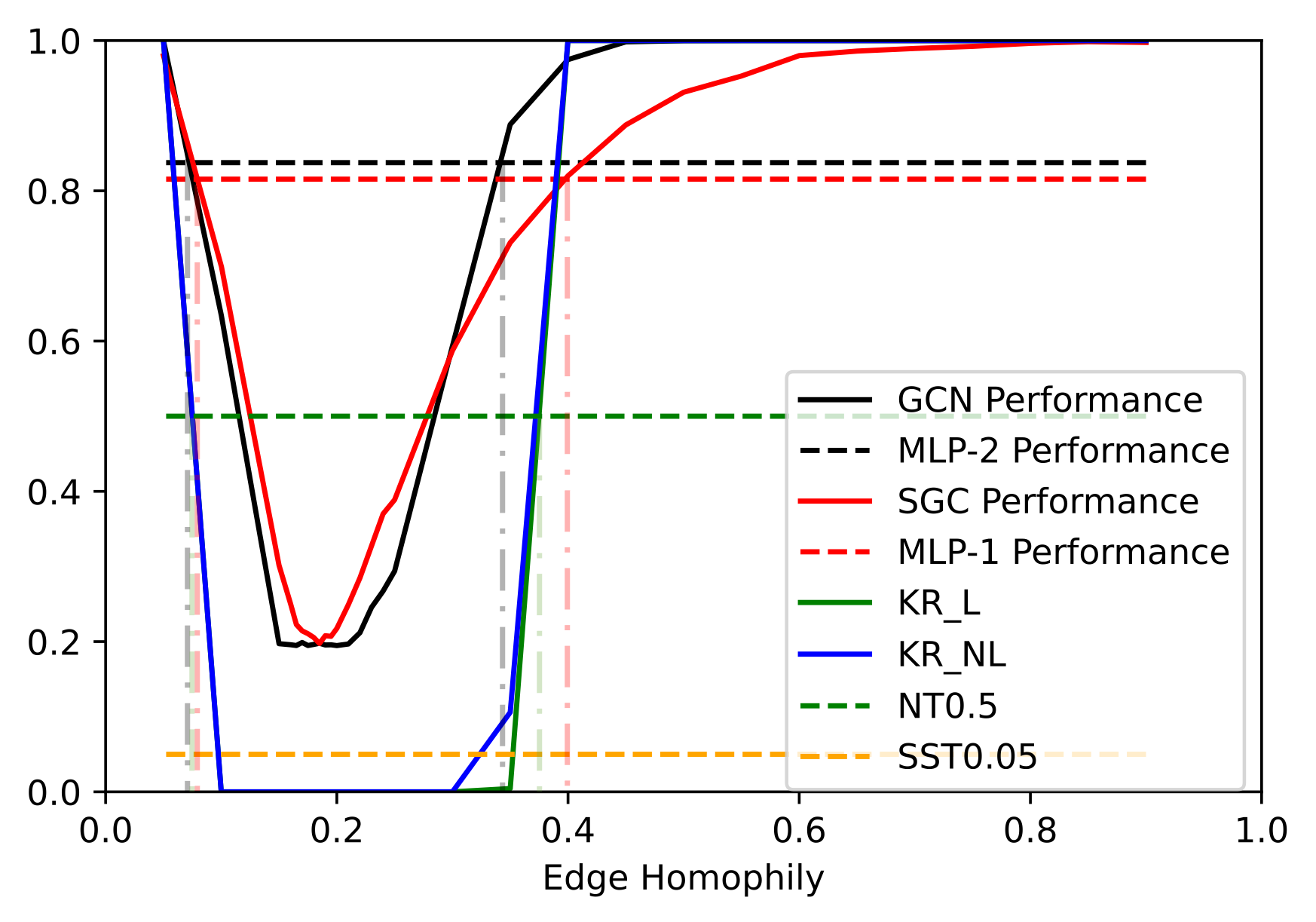}
     }\\
     \subfloat[PA: Model Curves]{
     \captionsetup{justification = centering}
     \includegraphics[width=0.5\textwidth]{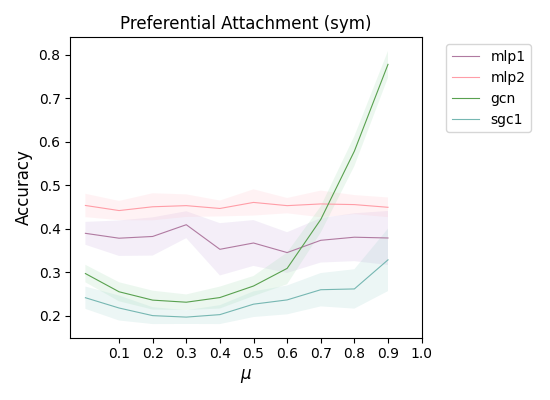}
     }\hspace*{-0.2cm}
     \subfloat[PA: Metric Curves]{
     \captionsetup{justification = centering}
     \includegraphics[width=0.5\textwidth]{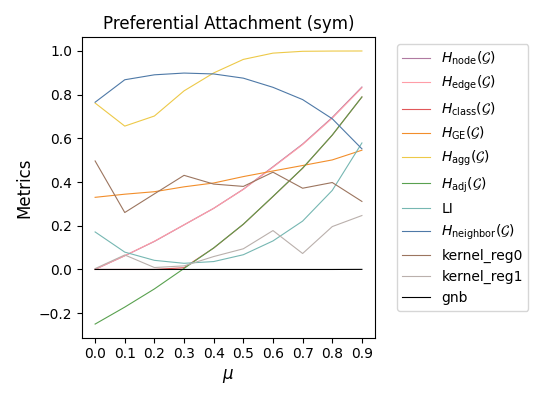}
     }\\
     \subfloat[GenCat: Model Curves]{
     \captionsetup{justification = centering}
     \includegraphics[width=0.5\textwidth]{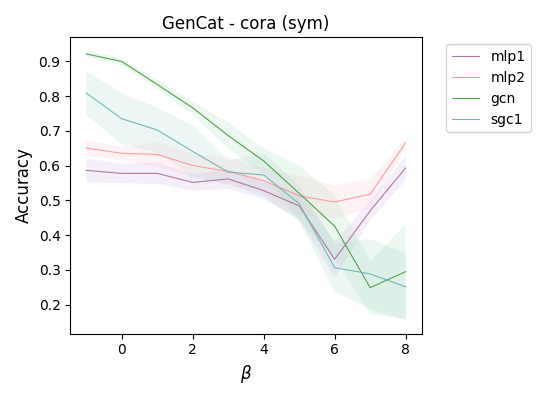}
     }\hspace*{-0.2cm}
     \subfloat[GenCat: Metric Curves]{
     \captionsetup{justification = centering}
     \includegraphics[width=0.5\textwidth]{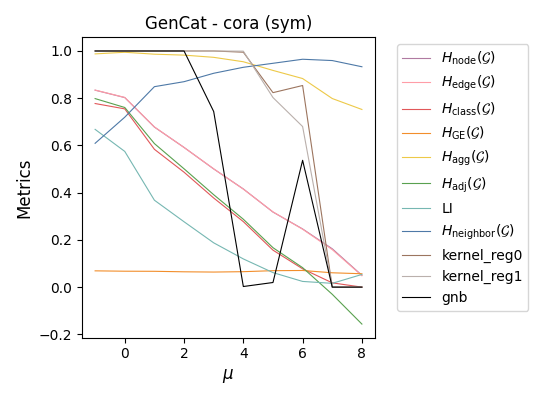}
     }
     }
     \caption{Comparison of metrics on synthetic graphs with different generation methods. Figure (a)(b) are from~\cite{luan2024addressing}.}
     \label{fig:comparison_metrics}
\end{figure}

Luan \etal~\cite{luan2022revisiting} are the first to use synthetic graphs to evaluate and compare homophily metrics, and the graph generation process is: (1) $10$ graphs are generated for each of the $28$ edge homophily levels, from $0.005$ to $0.95$, with a total of $280$ graphs; (2) Every generated graph has five classes, with $400$ nodes in each class. For nodes in each class, $800$ random intra-class edges and [$\frac{800}{\textit{H}_\text{edge}(\mathcal{G})} -800$] inter-class edges are uniformly generated ; (3) The features of nodes in each class are sampled from node features in the corresponding class of the base datasets, \eg{} the results in Figure~\ref{fig:comparison_metrics} (a)(b) are based on the node features from \textit{Cora}. 

\paragraph{Preferential Attachment (PA)}
Karimi \etal~\cite{karimi2018homophily} incorporate homophily as an additional parameter to the preferential attachment (PA)~\cite{barabasi1999emergence} model and Abu-El-Haija \etal~\cite{abu2019mixhop} extend it to multi-class settings, which is widely used in graph machine learning community. We give a summary of the generation process based on~\cite{abu2019mixhop} as follows.

Suppose graph $\mathcal{G}$ has a total number of $N$ nodes, $C$ classes, and a homophily coefficient $\mu$, the generation begins by dividing the $N$ nodes into $C$ equal-sized classes. Then, $\mathcal{G}$ (initially empty) is updated iteratively. At each step, a new node $v_i$ is added, and its class $y_i$ is randomly assigned from the set $\{1, \ldots, C\}$. Whenever a new node $v_i$ is added to $\mathcal{G}$, a connection between $v_i$ and an existing node $v_j$ in $\mathcal{G}$ is established based on the probability $\bar{p}_{ij}$, which is calculated as follows,
\begin{equation}
\begin{aligned}
\label{eq:puv}
&p_{ij} = \begin{cases}
d_j \times \mu, & \mbox{if $y_i = y_j$} \\
d_j \times (1-\mu) \times w_{d(y_i, y_j)}, & \mbox{otherwise}
\end{cases}, \\
&\bar{p}_{ij} = \frac{p_{ij}}{\sum_{k: v_k \in \mathcal{N}(v_i)} p_{ik}}
\end{aligned}
\end{equation}
where $y_i$ and $y_j$ are class labels of node $i$ and $j$ respectively, and $w_{d(y_i, y_j)}$\footnote{The code for calculating $w_{d(y_i, y_j)}$ is not open-sourced and we obtain the code from the authors of~\cite{abu2019mixhop}.} denotes the ``cost'' of connecting nodes from two distinct classes with a class distance of $d(y_i, y_j)$ \footnote{The distance between two classes simply implies the shortest distance between the two classes on a circle where classes are numbered from 1 to $C$.  For instance, if $C = 6$, $y_i = 1$ and $y_j = 5$, then the distance between $y_i$ and $y_j$ is $2$.}. The weight exponentially decreases as the distance increases and is normalized such that $\sum_d w_d = 1$.  For a larger $\mu$, the chance of connecting with a node with the same label increases. Lastly, the features of each node in the output graph are sampled from overlapping 2D Gaussian distributions. Each class has its own distribution defined separately.

\paragraph{GenCat}

GenCat~\cite{maekawa2022beyond,maekawa2023gencat}  generates synthetic graphs based on a real-world graph and a hyperparameter $\beta$ controlling the homophily/heterophily property of the generated graph. According to base graph and $\beta$, class preference mean ${M}^{(\beta)} \in \Rbb^{C\times C}$, class preference deviation ${D}^{(\beta)} \in \Rbb^{C\times C}$, class size distribution and attribute-class correlation $H \in \Rbb^{F\times C}$ are calculated, which are then used to create three latent factors: node-class membership proportions  ${U} \in [0, 1]^{N\times C}$, node-class connection proportions ${U}' \in [0, 1]^{N\times C}$, and attribute-class proportions ${V} \in [0, 1]^{F\times C}$. Here, $C$, $F$ and $N$ is the number of classes, features and nodes of the base graph. Finally, the synthetic graph is generated using these latent factors.

The class preference mean between class $c_1$ and class $c_2$ is initially calculated as:
$$
{M}_{c_1, c_2} = \frac{1}{|\Omega_{c_1}|} \sum_{i \in \Omega_{c_1}} \left ( \sum_{j \in \Omega_{c_2}} {A}_{ij} / \sum_j {A}_{ij} \right ),
$$
where $\Omega_{c_k} = \{v | {Z}_{v, k} = 1 \}$ is the set of nodes with class $c_k$.
Then ${M}_{c_1, c_2}$ is adjusted by $\beta$ as:
$$
{M}_{c_1, c_2}^{(\beta)} = \begin{cases}
\max({M}_{c_1, c_2} - 0.1 * \beta, 0) & (c_1 = c_2)\\
{M}_{c_1, c_2} + 0.1 * \beta/(C-1) & (c_1 \neq c_2)
\end{cases}.
$$
For a larger $\beta$, 
fewer edges would be generated later between nodes with the same class, thus corresponding to a more heterophilic graph. Then the range of $\beta$ is determined as $\{\left \lfloor 10 {M}_{\text{avg}} \right \rfloor -9, \left \lfloor 10 {M}_{\text{avg}} \right \rfloor -8, \ldots, \left \lfloor 10 {M}_{\text{avg}} \right \rfloor \}$. The average of intra-class connections is calculated as ${M}_{\text{avg}} = \frac{1}{C} \sum_{c_i} {M}_{c_i, c_i}$.

\subsubsection{Evaluation and Comparison of Metrics}
\label{sec:homophily_metric_evaluation_comparison}

\paragraph{Evaluation of Metrics}
The evaluation includes the following steps: (1) generate synthetic graphs with different homophily-related hyperparameters, \eg{} edge homophily for regular graphs, $\mu$ for PA model and $\beta$ for GenCat; (2) for each generated graph, nodes are randomly splitted into train/validation/test sets, in proportion of 60\%/20\%/20\%; (3) each baseline model (GCN, SGC-1, MLP-2 and MLP-1) is trained on every synthetic graph with the same hyperparameter searching range as~\cite{luan2022revisiting}, the mean test accuracy and standard deviation of $10$ runs are recorded; (4) calculate the corresponding metric values for each synthetic graph; (5) plot the metric curves and the performance curves of baseline models \wrt{} the homophily-related hyperparameters, compare their correlations.
\paragraph{Comparison and Observations}
The performance curves of baseline models are shown in Figure~\ref{fig:comparison_metrics} (a)(c)(e) and the metric curves are shown in Figure~\ref{fig:comparison_metrics} (a)(b)(d)(f). From the results, we observe that
\begin{itemize}
    \item \textbf{The GNN performance curves have different shapes under different generation methods.} The curves in RG (Figure~\ref{fig:comparison_metrics}(a)) are fully U-shaped, which indicates the performance of GNNs in low-homophily area can rebound up to the same level as the high-homophily area. However, in PA and GenCat, the curves are partially U-shaped, which implies that the performance in heterophily area cannot rebound back to the same level as homophily area.
    \item \textbf{The correlations between metrics and GNN performance are different.} For example in Figure~\ref{fig:comparison_metrics}(d), the curves of $\text{KR}_\text{L}, \text{KR}_\text{NL}$ and GNB highly correlate the curves of baseline GNNs on regular graphs. However, on PA, the curves of label informativeness and aggregated homophily have the highest correlations; and on GenCat, the curves of label informativeness and $\text{KR}_\text{L}, \text{KR}_\text{NL}$ exhibit the strongest correlation. 
\end{itemize}

The above observations indicate that 
\begin{itemize}
    \item Homophily value is not the only factor to influence GNN performance. Other factors, such as node degree distributions, local homophily distributions, structure-feature synergy might also highly impact the behavior of GNN.
    \item Our previous comparison of metrics on synthetic graphs could be questionable, because graphs with different generation methods might give different results. In future studies, a newly proposed metric should be tested on the synthetic graphs with all the three generation methods to get a comprehensive evaluation and comparison.
    \item It is evident that getting consistent and strict comparison results of the metrics only based on observation is hard. Thus, a quantitative evaluation benchmark for homophily metrics is required.
\end{itemize}

\subsection{Benchmarks for Heterogeneous Graphs}
\label{sec:benchmark_heterogeneous_graphs}
\begin{table}[htbp]
  \centering
  \caption{Meta-path subgraph edge homophily and the performance of heterogeneous GNNs (The results are adopted from \cite{guo2023homophily})}
  \resizebox{1\hsize}{!}{
    \begin{tabular}{c|ccccc}
    \toprule
    \toprule
    Datasets & Meta-Path & $\text{H}_\text{edge}$ & GCN   & Overall GCN & HAN \\
    \midrule
          & PcP   & 0.8789 & 91.49 $\pm$ 0.27 & \multicolumn{1}{c}{\multirow{5}[2]{*}{92.12 $\pm$ 0.23 }} & \multirow{5}[2]{*}{90.79 $\pm$ 0.43} \\
          & PAP   & 0.7936 & 89.76 $\pm$ 0.22 &       &  \\
    ACM   & PSP   & 0.6398 & 68.21 $\pm$ 1.13 &       &  \\
          & PcPAP & 0.7584 & 87.54 $\pm$ 0.55 &       &  \\
          & PcPSP & 0.5885 & 68.16 $\pm$ 1.22 &       &  \\
    \midrule
    \multirow{2}[2]{*}{IMDB}  & {MDM} & 0.6141 & {52.78 $\pm$ 1.01} & \multirow{2}[2]{*}{56.84 $\pm$ 2.66 } & \multirow{2}[2]{*}{55.81 $\pm$ 1.69} \\
          &  MAM  & 0.4443 & 51.31 $\pm$ 0.68 &       &  \\
    \midrule
          & {NSpN} & 0.2125 & 24.75 $\pm$ 0.96 & \multirow{3}[2]{*}{23.49 $\pm$ 0.57 } & \multirow{3}[2]{*}{26.75 $\pm$ 0.35} \\
    Liar  & {NSuN} & 0.1814 &  22.79 $\pm$ 0.41  &       &  \\
          & NCN   & 0.1885 & 23.34 $\pm$ 0.67 &       &  \\
    \bottomrule
    \bottomrule
    \end{tabular}%
    }
  \label{tab:hetgnn_homophily}%
\end{table}%

Compared with homogeneous graphs, it is much more complex to identify heterophily problem on heterogeneous graphs (HetG), where graph models are affected by multiple meta-path induced subgraphs with different subgraph homophily values. Several empirical studies have been conducted to explore the relationship between homophily and heterogeneous GNNs (HetGNNs).

\paragraph{Identify Heterophily for Heterogeneous GNNs}
Table~\ref{tab:hetgnn_homophily} shows the experimental results from~\cite{guo2023homophily}. We can observe that (1) GCN performs better on the meta-path induced subgraphs with higher homophily values than those with lower homophily values; (2) the performance of overall GCN~\footnote{Overall GCN is the GCN that is train on the entire HetG, \ie{} a combination of all subgraphs.} and HAN~\cite{wang2019heterogeneous}  are closely related to the average level of all subgraphs; (3) the performance of overall GCN and HAN is also close to GCN on the subgraph with the largest homophily value. The above evidence shows that there is high possibility that the performance of HetGNNs correlate with the homophily levels of the meta-path induced subgraphs. 

However, in the future, more benchmark datasets should still be tested to conclusively identify the relation between HetGNNs and homophily, and new benchmark datasets should be built to evaluate models and metrics. Furthermore, it is necessary to theoretically study how the synergy between subgraphs affects HetGNNs, \eg{} whether the mid-homophily pitfall~\cite{luan2024graph} exists for HetGNNs and what does it look like due to the existence of multiple subgraphs. A comprehensive relation, such as Figure~\ref{fig:comparison_metrics}(a)(c)(e), needs to be depict for HetGNNs. Note that, because of the unique properties, the previous studies on homogeneous graphs might not be valid for HetGs and we need to develop some new tools or theoretical frameworks to investigate the heterophily problem for HetGNNs. 

\begin{table}[htbp]
  \centering
  \caption{Connection Strength and Homophily (The results are adopted from \cite{hu2021ogb})}
   \resizebox{1\hsize}{!}{
    \begin{tabular}{cccc}
    \toprule
    \toprule
    Meta-Path & Connection Strength & $\text{H}_\text{edge}$ & \#Edges \\
    \midrule
    P-P   & 1     & 57.8  & 2,017,844 \\
    \midrule
          & 1     & 46.12 & 88,099,071 \\
          & 2     & 57.02 & 12,557,765 \\
    P-A-P & 4     & 64.03 & 1,970,761 \\
          & 8     & 66.65 & 476,792 \\
          & 16    & 70.46 & 189,493 \\
    \midrule
          & 1     & 3.83  & 159,884,165,669 \\
          & 2     & 4.61  & 81,949,449,717 \\
    P-A-I-A-P & 4     & 5.69  & 33,764,809,381 \\
          & 8     & 6.85  & 12,390,929,118 \\
          & 16    & 7.7   & 4,471,932,097 \\
    \midrule
    All Pairs & 0     & 1.99  & 782,926,523,470 \\
    \bottomrule
    \bottomrule
    \end{tabular}%
    }
  \label{tab:connection_strength_homophily}%
\end{table}%

\paragraph{Connection Strength}
Hu \etal~\cite{hu2021ogb} define the connection strength, which indicates the number of different possible paths along the meta-path, \eg{} meta-path “Paper-Author-Paper (P-A-P)” with connection strength 3 means that at least three authors are shared for the two papers of interest. It is similar to edge weight. The results in Table~\ref{tab:connection_strength_homophily} show that: (1) different meta-paths have different levels of homophily and compared to the direct connection, certain meta-paths have much higher homophily; (2) for a given meta-path, the edges with stronger connection strength tend to have higher homophily levels. Therefore, connection strength potentially has strong correlation with the performance of HetGNNs as well, and we should study such relationship in the future. Connection strength is another unique characteristic for studying heterophily on HetGs compared to homogeneous graphs.

\begin{figure*}[ht]
    \centering
     \includegraphics[width=1\textwidth]{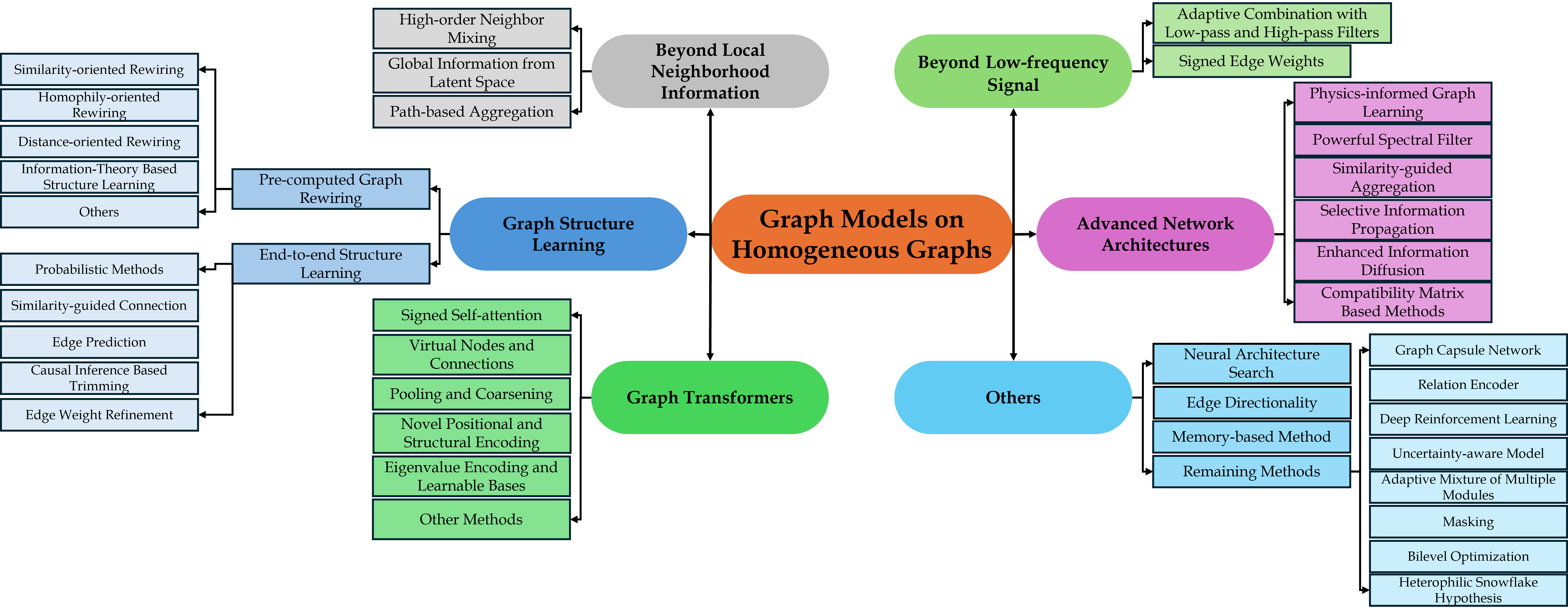}
     \caption{An overview of graph models for homogeneous graphs with heterophily.}
     \label{fig:diagram_homogeneous_graph}
\end{figure*}
\section{Supervised Learning on Homogeneous Graphs with Heterophily}
\label{sec:graph_models_for_heterophily}

Most graph models designed for heterophily problem are built on homogeneous graph and in this section, we will summarize and categorize the most representative methods~\footnote{Note that some methods can share similar features with each other, \eg{} global information from latent space, similarity-guided aggregation and graph rewiring methods. We mainly categorize them by their most prominent features.}. This section is organized as follows,
\begin{itemize}
    \item \textbf{Beyond Low-Frequency Signal} (Section~\ref{sec:beyond_low_frequency_signal}) GNNs are considered low-pass filters~\cite{maehara2019revisiting} to mainly capture low-frequency graph signals. However, high-frequency signal, which can catch the neighborhood dissimilarity, is found to be helpful for heterophily but is missing in most GNNs. High-pass filter and signed (negative) message passing are two major techniques to make use of high-frequency signals for GNNs.
    \item \textbf{Beyond Local Neighborhood Information} (Section~\ref{sec:beyond_local_neighborhood_information}) The dissimilarity depicted by heterophily is a property of the local neighborhood, and to find nodes with similar labels, the target nodes can get help from global information. This branch of methods includes obtaining global information from high-order neighbors, latent space, and path-based aggregation.
    \item \textbf{Advanced Network Architectures} (Section~\ref{sec:advanced_network_architectures}) This group of methods tries to design more sophisticated and powerful network architectures to make better use of heterophilic information. They include physics-informed graph learning, powerful spectral GNNs, similarity-guided aggregation, selective message propagation, enhanced information diffusion, and compatibility matrix based methods.
    \item \textbf{Graph Structure Learning} (Section~\ref{sec:graph_structure_learning}) This collection of approaches attempts to break the harmful or task-irrelevant connections and rebuild the edges to get a better graph structure for learning. They include pre-computed rewiring and end-to-end structure learning methods.
    \item \textbf{Graph Transformer} (Section~\ref{sec:graph_transformer}) Graph Transformer (GT) is a special type of graph model, which adapts self-attention and various encoding techniques to capture long-range dependency between nodes and is potentially helpful for heterophily. In this subsection, we will summarize the heterophily-specific GTs.
    \item \textbf{Others} (Section~\ref{sec:others}) There also exist some less representative approaches that are still worth mentioning. They are good additions to the mainstream methods, \eg{} neural architecture search, edge directionality,  memory-based methods, \etc{}
\end{itemize}
In the following subsections, we will introduce the above methods in detail. In  Figure~\ref{fig:diagram_homogeneous_graph}, we provide an overview of the graph models for supervised learning on homogeneous graphs with heterophily. 
\subsection{Beyond Low-Frequency Signal}
\label{sec:beyond_low_frequency_signal}

The incorporation of high-frequency signals is found to be one of the simplest and most effective methods for learning on heterophilic graphs~\cite{chien2021adaptive, bo2021beyond, luan2022revisiting}. The extracted high-frequency components can be utilized to replace or complement the aggregated features in Equation \eqref{eq:message_passing} for message passing. High-pass filter and signed edge weights are two typical ways to capture high-frequency signals, and we will review them in this subsection.

\subsubsection{Adaptive Mixing with High-pass Filters}
\label{sec:adaptive_combination_lp_hp_filters}

GPRGNN~\cite{chien2021adaptive} uses learnable weights that can be both positive and negative for feature propagation. This allows GPRGNN to adapt to heterophilic graphs and to handle both high- and low-frequency parts of the graph signals. FB-GNNs~\cite{luan2022complete} and ACM-GNNs~\cite{luan2022revisiting} add filterbanks to the uni-channel GCN with identity, LP and HP filtering channels, and propose a novel adaptive channel mixing mechanism to combine the channel information on a node-by-node basis. The multi-channel architecture can exploit beneficial neighbor information from different channels adaptively for each node, and the identity filter could guarantee less information loss of the input signal. Sun \etal~\cite{sun2022improving} argue that the complement of the original graph incorporates a high-pass filter and propose Complement Laplacian Regularization (CLAR) for an efficient enhancement of high-frequency components. Liu \etal~\cite{liu2023imbalanced} introduce GraphSANN, a method designed for imbalanced node classification on both homophilic and heterophilic graphs. It includes an adaptive subgraph extractor and a multi-filter subgraph encoder to construct different filtering channels to capture high-frequency information. Chen \etal~\cite{chen2023dirichlet} develop a framelet augmentation strategy by adjusting the update rules with positive and negative increments for low-pass and high-pass filters, respectively. On the basis of that, they design the Energy Enhanced Convolution (EEConv), which provably enhances the Dirichlet energy of the graph, thereby improving the model performance on heterophilic graphs. Zhang \etal~\cite{zhang2024unleashing} introduce a continuous GNN with high-pass filtering, alongside a feature aggregation technique to combine features from low-pass and high-pass networks, respectively. A dimension-level masking mechanism is employed to select the task-specific feature dimension and enhance the model adaptability to varying degrees of heterophily.

\subsubsection{Signed Edge Weights}
\label{sec:signed_message_passing}

FAGCN ~\cite{bo2021beyond} learns edge-level aggregation weights as GAT ~\cite{velivckovic2018graph}, but allows the weights to be negative, which enables the network to capture high-frequency components in the graph signals. GGCN~\cite{yan2022two} uses the cosine similarity to send signed neighbor features under certain constraints of relative node degrees. In this way, the messages are allowed to be multiplied by a negative sign or a positive sign. Intuitively, signed messages consist of the negated messages sent by neighbors of the opposing classes and the positive messages sent by neighbors of the same class. GloGNN~\cite{li2022finding} estimates a coefficient matrix to model the relative importance of nodes. With respect to a simple attention mechanism, this formulation allows for negative values as coefficients and is more efficient thanks to a change in the order of matrix multiplications. Choi \etal propose~\cite{choi2023signed} two strategies to improve signed propagation: (1) edge weight calibration, \ie{} they block the information propagation of highly similar nodes, which may decrease the discrimination ability in inference; (2) (Parameter update) confidence calibration, which penalizes the maximal and sub-maximal prediction values to be similar in order to reduce the probability of conflict evidence. 

\subsubsection{Others}
TEDGCN~\cite{yan2023trainable} redefines the depth of GCN as a trainable parameter continuously adjustable within $(-\infty,\infty)$ and the depth can be optimized without the prior knowledge regarding the properties of input graph. Negative depth intrinsically enables high-pass filtering via augmented topology for graph heterophily.

\subsection{Beyond Local Neighborhood Information}
\label{sec:beyond_local_neighborhood_information}

The standard GNNs aggregate information from local neighborhoods to learn embeddings for the targets node~\cite{kipf2016classification,velivckovic2018graph}. However, in heterophilic networks, nodes of the same class are more likely to be disconnected, and similar nodes can be far from each other. Therefore, several GNNs try to extend the scope of neighbor set $\mathcal{N}_u$ to reach nodes located further away in the graph. In this subsection, we will summarize the methods for propagating information beyond local neighborhoods.

\subsubsection{{High-order Neighbor Mixing}}
\label{sec:high_order_neighbor_mixing}

The (geodesic or shortest-path) distance $d(i,j)$ between two nodes, $i$ and $j$, in a graph is the number of edges in a shortest path connecting them~\cite{skiena1990combinatorics, bouttier2003geodesic}. There may be more than one shortest path between two nodes and if there is no path connecting $i,j$, $d(i,j)$ is defined as infinite. For node $i$, if $d(i,j) = k$, then $j$ is called the $k$-hop neighbor of $i$. If $k\geq 2$, $j$ is referred to as the high-order neighbor of $i$; otherwise, $j$ is the direct (1-hop) neighbor of $i$.

High-order neighbor mixing allows nodes to receive features from distant neighbors, which is potentially beneficial in heterophilic graphs. For example, MixHop~\cite{abu2019mixhop} propagates 1-hop and 2-hop neighborhood messages, which are encoded by different linear transformations and mixed by concatenation. H$_2$GCN~\cite{zhu2020beyond} designs ego- and neighbor-embedding separation, aggregation of high-order neighborhoods, and combination of intermediate representations to generalize the power of GNNs beyond the homophily setting. UGCN~\cite{jin2021universal} proposes to aggregate information from 1-hop, 2-hop and top-$k$ most similar nodes based on features. The attention mechanism is leveraged to flexibly choose which neighborhood is more relevant to the current graph. TDGNN~\cite{wang2021tree} uses the tree decomposition to separate neighbors at different $k$-hops into multiple subgraphs, propagates information on these subgraphs in parallel, and then merges the representations through a simple sum or a weighted sum with learnable attention coefficients.

\subsubsection{{Global Information from Latent Space}}
\label{sec:global_information_from_latent_space}

Unlike high-order neighbors, which are directly defined by the distance according to graph structure, some methods capture long-range information by reconnecting distant nodes in a latent space, which essentially rebuild the neighbor set.

Geom-GCN~\cite{pei2020geom} pre-computes unsupervised node embeddings in a latent space and  defines the geometric relationships. It uses a geometric aggregation scheme and a bi-level aggregator to capture the information of structural neighborhoods, which can be distant nodes. NL-GNN~\cite{liu2021non} and GPNN~\cite{yang2022graph} adopt the attention mechanism or leverage a pointer network~\cite{vinyals2015pointer} on a sampled subgraph to rank the potential neighbor nodes according to the attention scores or the relevant relationships to the target node. The most similar potential neighbors can be discovered and selected from the latent space. HOG-GCN ~\cite{wang2022powerful} tries to model the relative importance between two nodes based on a homophily degree matrix with the label propagation technique to explore how likely a pair of nodes belong to the same class. Deformable GCN~\cite{park2022deformable} simultaneously learns node representations and node positional embeddings, and efficiently perform Deformable Graph Convolution (Deformable GConv) in multiple latent spaces with latent neighborhood graphs. Deformable GConv adaptively deforms the convolution kernels to handle heterophily and variable-range dependencies between nodes. Huang \etal~\cite{huang2022local} propose PowerEmbed, a simple layer-wise normalization technique, which can provably express the top-$k$ leading eigenvectors of the graph operator, constructing a latent space based on the eigenvectors and producing a list of representations ranging from local features to global signals. Dong \etal~\cite{dong2024differentiable} propose Differentiable Cluster Graph Neural Network (DC-GNN), which embeds the clustering inductive bias into GNNs and each node is learned to be clustered into multiple global clusters, so that we can establish connections between the distant nodes.

\subsubsection{Path-based Aggregation}
\label{sec:path_based_aggregation}

PathNet~\cite{sun2022beyond} proposes a novel path aggregation paradigm with a maximal entropy path sampler to capture the structure context information, and leverages order and distance information of path to encode complex semantic information in graphs. RAW-GNN~\cite{jin2022raw} utilizes breadth-first random walk search to capture homophily information, and depth-first search to collect heterophily information to make it suitable for both homophilic and heterophilic graphs. Besides, it replaces the conventional neighborhoods with path-based neighborhoods and introduces a new path-based aggregator based on Recurrent Neural Networks. Xie \etal~\cite{xie2023pathmlp} design a similarity-based path sampling strategy to capture smooth paths containing high-order homophily and propose PathMLP to encode messages carried by paths through adaptive path aggregation.

\subsection{Advanced Network Architectures}
\label{sec:advanced_network_architectures}

Besides high-order information, people devised lots of sophisticated architectures with advanced learning strategies to address heterophily problem. These methods mainly refine the AGGREGATE and UPDATE functions in the message passing framework and we will provide an organized overview of the representative approaches in this subsection.

\subsubsection{Physics-Informed Graph Learning}
\label{sec:physics_informed_graph_learning}
Physics-Informed Graph Learning (PIGL) incorporates observation data and physical laws, such as the distribution rule of continuous space~\cite{jia2021physics,gao2022physics}, into graph representation learning. The physical constraints, \eg{} repulsive and attractive forces, can help to model the complicated interactions between nodes~\cite{peng2022physics}, including the heterophilic relationship.

Wang \etal~\cite{wang2022acmp} develop a new message passing framework by considering the Allen-Cahn force, which is associated with phase transitions of an interacting particle system influenced by both attractive and repulsive forces. The resulting dynamics form a reaction-diffusion process, leading to the Allen-Cahn message passing (ACMP) method, where the numerical iteration for the particle system solution constitutes the message passing propagation. GREAD~\cite{choi2023gread}  features a reaction-diffusion layer with three standard reaction equations from natural sciences and four additional reaction terms, including a specially developed Blurring-sharpening (BS) term.  Zhao \etal~\cite{zhao2023graph} propose a model based on the convection-diffusion equation (CDE), which includes a diffusion term to aggregate information from homophilic neighbors and a convection term that allows controlled information propagation from heterophilic neighbors. It can explicitly control the propagation velocity at each node, which generalizes the curvature-preserving models such as Beltrami~\cite{chamberlain2021beltrami} and mean-curvature~\cite{song2022robustness}. Zhang \etal~\cite{zhang2023steering} utilize prototypes, \ie{} class centers, of labeled data to guide the learning process, treating graph learning as a discrete dynamic system. By adopting the pinning control concept from automatic control theory, they design feedback controllers for the learning process, aiming to reduce the disparity between features derived from message passing and the optimal class prototypes of nodes. Thus, it can produce class-relevant features and adjust the information aggregated from incompatible neighbors in heterophilic graphs. Park \etal~\cite{park2024mitigating} propose to apply the reverse of the heat diffusion process to learn the node states in the past time steps, which are far away from the equilibrium state of diffusion process and are shown to be helpful for learning distinguishable node representations in heterophilic graphs.

\subsubsection{Powerful Spectral GNNs}
\label{sec:expressive_spectral_filter}
Spectral GNN is a type of network that is built on graph signal filtering techniques in the spectral domain. People try to develop more expressive spectral filters to enhance the AGGREGATE function in the message passing framework for better performance on heterophilic graphs.

\paragraph{Strong Polynomial Filters}
BernNet ~\cite{he2021bernnet} designs a scheme to learn arbitrary graph spectral filters with Bernstein polynomial to address heterophily. Wang \etal~\cite{wang2022howpowerful} propose JacobiConv, which uses Jacobi basis due to its orthogonality and flexibility to adapt to a wide range of weight functions.  Li \etal~\cite{li2023pc} propose a two-fold filtering mechanism to extract homophily in heterophilic graphs and vice versa, utilizing an extension of the graph heat equation for heterophilic aggregation of global information from a long distance, which can be precisely approximated by Poisson-Charlier (PC) polynomials. Based on this, they develop a powerful graph convolution, PC-Conv, and its instantiation, PCNet, to exploit information at multiple orders. Huang \etal~\cite{huang2024universal} develop an adaptive heterophily basis by incorporating graph heterophily levels and integrating it with the homophily basis. A universal polynomial basis UniBasis and a general polynomial filter UniFilter are then created. Huang~\cite{huang2024optimizing} develop an adaptive Krylov subspace approach to enhance polynomial bases with provable controllability over the graph spectrum to adapt graphs with various heterophily levels. Furthermore, they introduced AdaptKry, an optimized polynomial graph filter derived from the bases in the adaptive Krylov subspaces, and extended AdaptKry, which incorporates multiple adaptive Krylov bases without incurring extra training costs.

\paragraph{Adaptive Node-Specific Filtering}
NFGNN~\cite{zheng2023node} uses node-oriented spectral filtering, which applies local spectral filtering by employing filters translated to specific nodes, effectively addressing the challenge of diverse local homophily patterns. Guo \etal introduce DFS~\cite{guo2023graph}, a diverse spectral filtering framework, which automatically learns node-specific filter weights to exploit and balance the local and global structure information adaptively. 

\paragraph{Inductive Spectral GNNs}
Xu \etal~\cite{xu2024slog} propose SLOG to address the limited expressive power of polynomial filters in the spectral domain and transductive limitation for spectral GNNs. Specifically, they develop SLOG(L) for inductive learning on heterophilic graphs, which incorporates the global
uniform filter and the local adaptive filter to capture the local frequency signals.

\subsubsection{Similarity-Guided Aggregation}
\label{sec:similarity_guided_aggregation}
The basic message passing framework, in which the feature aggregation is guided by the existing adjacency relation, is shown to be misleading on graphs with heterophily. Some new models attempt to find a better guidance beyond the original graph for neighborhood information aggregation, and the most common and representative way is the similarity-guided aggregation.

UGCN ~\cite{jin2021universal} and SimP-GCN ~\cite{jin2021node} choose top $k$ similar node pairs in terms of cosine similarity of features for each node and construct the neighbor set through kNN algorithm. Similarly, BM-GCN~\cite{he2022block} leverages an MLP to estimate a block similarity matrix, representing the likelihood of interconnections between nodes of two classes. This block matrix is then integrated in the graph convolution operation to increase the strength of the connection between nodes that are more likely relevant to each other. Zou \etal~\cite{zou2023similarity} propose the Similarity-Navigated Graph Neural Network (SNGNN), which uses Node Similarity matrix coupled with mean aggregator. Two variants, SNGNN+ and SNGNN++, are devised to select similar neighbors and preserve distinguishable features, which can integrate the knowledge from both node features and graph structure to handle heterophily problem. Chen \etal~\cite{chen2023lsgnn} develop LSGNN, a model that employs the local similarity (LocalSim) aware multi-hop fusion to guide the learning of node-level weights and Initial Residual Difference Connection (IRDC) to extract more informative multi-hop information. Zhu \etal~\cite{zhu2024universally} introduce $d$ Neighbor Similarity Preserving Graph Neural Network (NSPGNN), which employs a dual-kNN graph to guide neighbor-similarity propagation to improve the robustness against adversarial attacks on heterophilic graphs.

\subsubsection{{Enhanced Information Diffusion}}
\label{sec:enhance_information_diffusion}
In addition to similarity-guided aggregation, several methods try to boost the expressivity of GNNs by modifying the information diffusion process in AGGREGATE function with more advanced and sophisticated mechanisms.

Bodnar \etal~\cite{bodnar2022neural} utilize cellular sheaf theory to equip graph with a geometrical structure and show that the underlying geometry of the graph is closely related to the performance of GNNs on heterophilic graphs. Then, they introduce a hierarchical sheaf diffusion process based on the sheaf Laplacian and develop Sheaf Convolutional Network (SCN), which uses the discrete sheaf diffusion process for effective learning on heterophilic graphs. In LWGNN~\cite{dai2022label}, the authors observe that label-wise propagation is more distinctive on heterophilic graphs. Due to unknown labels, propagation is driven by pseudo-labels estimated for each node. Dong \etal~\cite{dong2022protognn} develop ProtoGNN, which effectively combines node features with structural information, and learns multiple class prototypes with a slot-attention mechanism. With the prototype representations, it can bypass local neighborhoods to capture global information, making it powerful when node features are prominent and local graph information is scarce. Ordered GNN~\cite{song2023ordered} proposes to order the messages passing into the node representation, with specific blocks of neurons targeted for message passing within specific hops. This is achieved by aligning the hierarchy of the rooted-tree of a central node with the ordered neurons in its node representation. Azabou \etal~\cite{azabou2023half} introduce Half-Hop, an edge upsampling method that adds “slow nodes” at each edge to mediate communication between a source and a target node. Cheng \etal~\cite{cheng2023prioritized} propose PPro, which is a versatile framework to learn prioritized node-wise message propagation in GNNs. It consists of a backbone GNN model, a propagation controller to determine the optimal propagation steps for nodes, and a weight controller to compute the priority scores for nodes. ClenshawGCN~\cite{guo2023clenshaw} equips GCN with two residual modules for heterophily: the adaptive initial residual connection and the negative second order residual connection. Micheli \etal~\cite{micheli2024echo} propose Graph Echo State Networks (GESN), which follows the reservoir computing paradigm with different reservoir designs, \eg{} ring, sparse and grouped reservoirs, for heterophily. Reservoir computing is a node embedding method, which recursively fuses the input and latent neighborhood features with random initialized and untrained input-to-reservoir and reservoir recurrent weights.

\subsubsection{Selective Message Propagation}
\label{sec:selective_message_propagation}

Models with selective message propagation can learn to retain the useful or relevant information and discard the detrimental parts in node embeddings, or selectively update node representations with different rates.

DMP~\cite{yang2021diverse} extends the standard label-based homophily measure to an attribute-based homophily measure by considering the connectivity among nodes that share the same features. Motivated by the observation that different features show different levels of heterophily, the authors propose to learn a weight for each feature in the node embeddings and aggregate messages using these attribute-wise weights. GBK-GNN~\cite{du2022gbk} is a GCN with two kernels, one tailored to deal with homophilic node pairs, the other for heterophilic pairs. A gate (encoded through a MLP) is trained to understand which of the two kernels should be applied to each node pair. FSGNN~\cite{maurya2022simplifying} proposes to use Softmax as a regularizer and "soft-selector" of features aggregated from neighbors at different hop distances. Rusch \etal present~\cite{rusch2022gradient} Gradient Gating (G$^2$), a novel framework that is based on gating the output of GNN layers with a mechanism for multi-rate flow of message passing information across nodes of the underlying graph, and local gradients are harnessed to further modulate message passing updates. ASGC~\cite{chanpuriya2022simplified} proposes adaptive simple graph convolution, which adaptively learns smooth or non-smooth filters, and thus can adapt to both homophilic and heterophilic structures. Guo \etal~\cite{guo2022gnn} propose Edge Splitting GNN (ES-GNN), which conduct edge splitting and subgraphs information propagation alternatively to disentangle the task-relevant and irrelevant features adaptively. Finkelshtein \etal~\cite{finkelshtein2023cooperative} propose cooperative graph neural networks, where each node can learn to choose from the action set \{'listen', 'broadcast', 'listen and broadcast', 'isolate'\} for message propagation. It provides a more flexible paradigm for message passing, where each node can determine its own strategy based on their state, effectively exploring the graph topology while learning. Pei \etal~\cite{pei2024multi} argue and demonstrate that "if messages are separated and independently propagated according to their category semantics, heterophilic mixing can be prevented". To this end, they designed the Multi-Track Graph Convolutional Network (MTGCN), where nodes are learned through the attention mechanism to be categorized to different tracks and node representations in different tracks are updated independently.

\subsubsection{Compatibility Matrix Based Methods}
\label{sec:compatibility_matrix_based_methods}
\paragraph{Compatibility Matrix} Let ${Y} \in \mathbb{R}^{N \times C}$ where $Y_{v j}=1$ if $y_v=j$, and $Y_{v j}=0$ otherwise. Then, the compatibility matrix is defined as:
$$CB = \left({Y}^{\top} {A} {Y}\right) \oslash \left({Y}^{\top} {A E}\right)
$$
where $\oslash$ is Hadamard (element-wise) division and ${E}$ is a $N \times C$ all-ones matrix. 

Compatibility matrix models the (empirical) connection probability of nodes between each pair of classes, \eg{} $CB_{c_1c_2}$ means a node in class $c_1$ has probability $CB_{c_1c_2}$ to connect to the the node in $c_2$. It enables GNNs to learn from heterophilic graphs and is powerful for the scenarios when node features are incomplete or missing. Moreover, the learned compatibility information can provide interpretable understanding of the connectivity patterns within the graph. 

CPGNN ~\cite{zhu2021graph} models label correlations through a compatibility matrix and propagates a prior belief estimation into the GNN by using the compatibility matrix. CLP~\cite{zhong2022simplifying} learns the class compatibility matrix and then aggregates label predictions using label propagation algorithm weighted by the compatibility of the class labels. Zheng \etal~\cite{zheng2024revisiting} reformulate GNNs with heterophily into a unified heterophilic message-passing (HTMP) mechanism and reveal that their success is due to the implicit enhancement of compatibility matrix among classes. To fully exploit the potential of the compatibility matrix, they propose Compatibility Matrix-aware Graph Neural Network (CMGNN), which uses a constraint enhanced compatibility matrix to construct desired neighborhood messages.

\subsection{Graph Structure Learning}
\label{sec:graph_structure_learning}
GNNs are highly sensitive to the quality of the given graph structures~\cite{zhu2021survey} and suboptimal connections, such as heterophilic edges, often exist in real-world graphs~\cite{dai2018adversarial}. However, the data-centric methods often overlook the potential defects of the underlying graph structure, which may lead to inferior performance. In pursuit of a better graph structure for downstream tasks, Graph Structure Learning (GSL) has attracted considerable attention in recent years. GSL is a data-centric approach that jointly optimizes the graph structure and the corresponding GNN models~\cite{fatemi2021slaps, zhu2021survey, zhou2024opengsl}, and it is important for dealing with the heterophily problem~\footnote{Benchmark for GSL and heterophily: Zhou \etal~\cite{zhou2024opengsl} establish OpenGSL, the first comprehensive benchmark for Graph Structure Learning (GSL), which enables a fair comparison among state-of-the-art GSL methods by evaluating them across various popular datasets using uniform data processing and
splitting strategies.}. In this subsection, we summarize GSL methods for heterophily and categorize them into two classes: pre-computed graph rewiring and end-to-end structure learning.

\subsubsection{Pre-computed Graph Rewiring}
\label{sec:precomputed_graph_rewiring}
Pre-computed graph rewiring methods calculate the optimized graph structure before training the models. The graph structure optimization and model training are two separate processes and cannot be integrated into a single differentiable pipeline.

\paragraph{Homophily-oriented Rewiring}
Wang \etal~\cite{wang2023improving} reconstruct the 2-hop graph structure, the $k$NN graph structure built with semantic features, and $k$NN graph structure built with structural features. These three new graph structures with high homophily are integrated into a Multi-View Graph Fusion Network (MVGFN) along with the original graph structure to learn a group of more expressive features. Zheng \etal~\cite{zheng2023finding} introduce Graph cOmplementAry Learning (GOAL), which utilizes graph complementation to find the missing-half structural information for a given graph including homophily- and heterophily-prone topology, and complement graph convolution to handle the complemented graphs. Gong \etal~\cite{gong2023neighborhood} design Neighborhood Homophily (NH) to measure the label complexity in neighborhoods and integrate this metric into GCN architecture, leading to the Neighborhood Homophily-based Graph Convolutional Network (NHGCN). Neighbors are grouped and aggregated based on estimated NH values, and the node predictions are used to continually refine NH estimations; the metric estimation and model inference are optimized iteratively. Xue \etal~\cite{xue2024datacentric} pre-process the heterophilic graphs by training an edge classifier to transform it to its homophilic counterpart, and then apply a homophilic GNN on the transformed graph.

\paragraph{Distance-oriented Rewiring}
Li \etal~\cite{li2023restructuring} propose an adaptive spectral clustering algorithm to reconstruct graphs with higher homophily. It restructures the graph according to node embedding distance derived from a division of Laplacian spectrum to maximize homophily. Based on a new class of Laplacian matrices which provably have more flexibility to control the diffusion distance between nodes than the conventional Laplacian, Lu \etal~\cite{lu2024representation} propose Directional Graph Attention Network (DGAT), which includes topology-guided neighbour pruning and edge adding mechanisms to remove the noisy nodes and capture the helpful long-range neighborhood information with the new Laplacian. Furthermore, DGAT is able to combine  feature-based attention with global directional information extracted from the graph topology to enable topological-aware information propagation. 

\paragraph{Similarity-Oriented Rewiring}
WRGNN~\cite{suresh2021breaking} takes the degree sequence of neighbor nodes as a metric of structural similarity to reconstruct the graph in which different neighbors are indicated by different relations of edges. Jiang \etal~\cite{jiang2021gcn} develop the efficient-spectral-clustering (ESC) and ESC with anchors (ESCANCH) to aggregate feature representations from similar nodes effectively. The authors also create a learnable rewired adjacency matrix using a unique data pre-processing method and similarity learning, which enhances the structure learning for graphs with low homophily. DHGR~\cite{bi2022make} rewires graphs by adding homophilic edges and pruning heterophilic edges, guided by comparing the similarity of label/feature-distribution of node neighbors. Deac~\cite{deac2023evolving} propose Evolving Computation Graphs (ECG), a method that employs weak classifiers to generate node embeddings to define a similarity metric between nodes. With the metric, edges are selected in a kNN fashion to rewiring the computation graph of GNNs, adding connections between nodes that are likely to be in the same class.

\paragraph{Information-Theory Based Structure Learning}
The SE-GSL framework~\cite{zou2023se} uses structural entropy and an encoding tree to abstract the hierarchy of the graph, maximizing the content of embedded information by merging auxiliary neighborhood attributes with the original graph. It also includes a novel sample-based mechanism to reconstruct the graph structure based on node structural entropy distribution, enhancing connectivity and robustness, especially on graphs with noisy and heterophilic structures. Choi \etal~\cite{choi2023node} propose to use node mutual information (MI) to capture dependencies between nodes in heterophilic graphs and design k-MIGNN, which makes use of MI as the connection weights in the message propagation.

\paragraph{Others}

As prior research has proven that low-rank approximation could achieve perfect recovery under certain conditions, Liang \etal~\cite{liang2023predicting} propose the Low-Rank Graph Neural Network (LRGNN), where they recover a global label relationship matrix to replace the aggregation matrix by solving a robust low-rank matrix approximation problem. Xu \etal~\cite{xu2023node} propose duAL sTructure learning (ALT) framework, which decomposes a given graph and extract complementary graph signals adaptively for node classification. Wu \etal~\cite{wu2023leveraging} propose a label-free structure learning method, which trains a link prediction model on the graph data and uses it to construct graph filter as an initialization for the adaptive filter-based GNNs.

\subsubsection{End-to-end Structure Learning}
\label{sec:end_to_end_structure_learning}
The end-to-end structure learning methods insert the structure optimization step into the process of model training and the whole pipeline is differentiable.

\paragraph{Probabilistic Methods}
Wang \etal~\cite{wang2021graph} propose Graph Estimation Networks (GEN), in which graph estimation is implemented based on Bayesian inference to maximize the posterior probability, which attains mutual optimization with GNN parameters in an iterative (EM) framework. Wu \etal~\cite{wu2023learning} propose Learning to Augment (L2A) framework, which simultaneously learns the GNN parameters and the graph structure augmentations in a variational inference framework. It applies two auxiliary self-supervised tasks to exploit both global position and label distribution information in the graph structure to further reduce the reliance on training labels and improve applicability to heterophilic graphs.

\paragraph{Edge Prediction}
Zhao \etal~\cite{zhao2021data} develop the GAUG framework, which can effectively leverage a neural edge predictor to promote intra-class edges and demote inter-class edges, increasing the performance of GNN on node classification via edge prediction. Ye \etal~\cite{ye2021sparse} propose SGAT, which is able to discard useless edges in the network and perform attention-weighted message passing based on the remaining edges. On some heterophilic benchmarks, up to 80\% of the edges are found to be redundant or not impactful for model performance.

\paragraph{Similarity-guided Connection} 
Choi \etal~\cite{choi2022finding} introduce a supplementary module, which uses optimal transport to measure the similarity between two nodes with their subgraphs. With a predefined confidence ratio, they apply label propagation on a subset of high-confidence edges and prevent connections between dissimilar nodes. Yang \etal~\cite{yang2024graph}  propose Graph Neural Networks with Soft Association between Topology and Attribute (GNN-SATA), which consists of a Graph Pruning Module (GPM) to remove irrelevant or false edges, and the Graph Augmentation Module (GAM) to add useful connections to the original graph. A similarity matrix is constructed to establish the connections between features and graph structures.

\paragraph{Causal Inference Based Trimming}

In order to estimate and weaken the Distraction Effect (DE) of neighboring nodes, He \etal~\cite{he2024cat} propose the Causally graph Attention network for Trimming heterophilic graph (CAT). They estimate DE by Total Effect, which can be derived from causal inference by semantic cluster intervention. They follow the intervened attention learning and graph trimming steps to learn attention scores and to identify the neighbors with the highest DE, \ie{} distraction neighbors, and remove them.

\paragraph{Edge Weight Refinement} 
Yan \etal~\cite{yan2022two} show the effectiveness of two end-to-end structure learning strategies in GGCN: structure-based edge correction, which learns corrected edge weights from structural properties, and feature-based edge correction, which learns signed edge weights from node features.

\subsection{Graph Transformers}
\label{sec:graph_transformer}
\subsubsection{A Brief Introduction}
We first provide a brief introduction to graph transformer.
\paragraph{Self-Attention}
Self-attention is the main component of transformer. For layer $k>0$, given node feature matrix ${X}^{(k)} \in \mathbb{R}^{n \times d}$, a single self-attention head is computed as follows,
\begin{align*}
& \operatorname{Attn}\left({X}^{(k)}\right):=\operatorname{Softmax}\left(\frac{{Q K^\top}}{\sqrt{d_k}}\right) {V},\\
& {Q}:={X}^{(k)} {W}_Q, \; {K}:={X}^{(k)} {W}_K, \; \text { and } \; {V}:={X}^{(k)} {W}_V,
\end{align*}
where the Softmax is applied row-wise; ${W}_Q, {W}_K \in \mathbb{R}^{d \times d_k}$, and ${W}_V \in \mathbb{R}^{d \times d}$; $d_k$ denotes the feature dimension of ${Q}$ and ${K}$. Self-attention essentially learns a fully-connected graph structure by projecting ${X}^{(k)}$ linearly. Then, multi-head attention can be applied to concatenate multiple self-attention heads together, and a MLP module, residual connections, and normalization operations are followed to produce the final output.

\paragraph{Structural and Positional Encodings}
Positional and structural encodings have been leveraged to enhance the performance of Graph Transformers. Positional encodings can make nodes aware of their relative position to the other nodes in a graph, and structural encodings can extract graph structure information on local, relative, or global levels. Some commonly used encoding methods are node degrees~\cite{chen2022structure}, eigenvalues of graph Laplacian~\cite{kreuzer2021rethinking}, shortest-path distance between nodes~\cite{li2020distance}, and relative random walk probabilities encoding~\cite{ma2023graph}, \etc{}

Graph Transformer can capture long-range dependency between distant nodes, which is considered to be more effective than MPNN-like models to address the local heterophily problem~\cite{muller2023attending} \footnote{Comments from Liheng Ma: Vanilla Transformers also suffered from over-smoothing and rank-collapse~\cite{dong2021attention,wang2022anti}. However, to the best of my knowledge, it is unknown and under-explored whether the regular GTs (Graphormer, SANs, GraphGPS, GD-Graphormer, GRIT, \etc{}) can handle heterophilic graphs.}. Muller \etal~\cite{muller2023attending} demonstrated that adding global attention to GCN can empirically facilitate the information propagation and improve model performance. In the following subsection, we will summarize graph transformers for learning on heterophilic graphs.

\subsubsection{Graph Transformers for Heterophily}
\paragraph{Signed Self-Attention}
Chen \etal~\cite{chen2023signgt} propose the Signed Attention-based Graph Transformer (SignGT) to adaptively capture various frequency information from the graphs. Specifically, SignGT produces signed attention values according to the semantic relevance of node pairs to preserve diverse frequency information between different node pairs. Lai \etal~\cite{lai2023self} propose Self-Attention Dual Embedding Graph Convolutional Network (SADE-GCN), a dual GNN architecture that encodes node features and graph topology in separate embeddings and allows negative and asymmetric attention weights.

\paragraph{Virtual Nodes and Connections}
Kong \etal~\cite{kong2023goat} propose GOAT, a scalable global graph transformer, where each node conceptually attends to each other by a projection matrix and a codebook~\cite{van2017neural}  computed by the K-Means algorithm, and homophily/heterophily relationships can be learned adaptively from the data. Fu \etal~\cite{fu2024vcrgraphormer} propose Virtual Connection Ranking based Graph Transformer (VCR-Graphormer), which captures the content-based global information and encodes the heterophily information into the token list. The input graph is partitioned into several clusters, and a super node is assigned to each cluster. The super node connects every node within the cluster with structure-aware virtual connections, which can propagate shared information.

\paragraph{Pooling and Coarsening}

Adaptive Node Sampling for Graph Transformer (ANS-GT)~\cite{zhang2022hierarchical} employs a multi-armed bandit algorithm to adaptively sample nodes for attention mechanisms, enhancing the model's ability to focus on more informative nodes. Furthermore, ANS-GT introduces coarse-grained global attention with graph coarsening, which helps the graph transformer capture long-range dependencies to address heterophily. Gapformer~\cite{liu2023gapformer} utilizes both local and global graph pooling methods to coarsen large-scale nodes into a reduced set of pooling nodes. Attention scores are calculated exclusively among these pooling nodes instead of across all nodes, enabling the propagation of long-range information to effectively handle heterophily.

\paragraph{Novel Positional/Structural Encodings}

Deformable Graph Transformer (DGT)~\cite{park2022deformable} constructs multiple node sequences with various sorting criteria to consider both structural and semantic proximity, and combines them with learnable Katz Positional Encodings. MPformer~\cite{2024mpformer} introduces an information aggregation module, Tree2Token, which aggregates central node and its neighbor information at various hops, treating node features as token vectors and serializing these token as sequences. For the newly generated sequence, they develop a position encoding technique, Tree2Token, which employs the shortest path distance to encode relative positional relationships between nodes. It captures feature distinctions between neighboring nodes and ego-nodes, facilitating the incorporation of heterophilic relationships into the transformer architecture. 

\paragraph{Eigenvalue Encoding and Learnable Bases}
Bo \etal~\cite{bo2022specformer} propose Specformer, which encodes the range of eigenvalues via positional embedding to capture both magnitudes of frequency and relative frequency information. Based on the learned eigenvalue representations, they also design a decoder with a bank of learnable bases to construct a permutation-equivariant and non-local graph convolution, which is able to capture high-frequency graph signals. Through ablation studies, eigenvalue encoding is found to be more effective for graph transformers on heterophily problem.

\paragraph{Other Methods}

Heterophily Learning Network (HL-Net)~\cite{lin2022hl} leverages graph transformer to handle both the heterophily and homophily contexts in scene graph generation. It consists of an adaptive reweighting transformer module and a heterophily-aware message passing scheme, which distinguishes the heterophily and homophily patterns between objects and relationships, and modulates the message passing in graphs accordingly. Wang \etal~\cite{wang2023heterophily} propose a Heterophily-aware Graph Attention network (HGAT)  that assigns different weights to edges according to different heterophilic types, which enables nodes to acquire appropriate information from distinct neighbors. HGAT employs local label distribution as the underlying heterophily, which can adapt to different homophily ratios of the networks.

\subsection{Others} 
\label{sec:others}

\subsubsection{Neural Architecture Search}
\label{sec:neural_architecture_search}
Traditional architecture design of neural network is a time-consuming process that relies heavily on human experience and expertise. Neural Architecture Search (NAS) is a technique that aims to automate architecture engineering and outperform manually designed architectures~\cite{elsken2019neural}. There are three key components in NAS: search space, search strategy, and performance estimation strategy. Architecture design is important for learning on heterophilic graphs as well, and in the following paragraph, we will summarize NAS methods for heterophily.

Zeng \etal~\cite{zeng2021relational} propose LADDER-GNN, a ladder-style GNN architecture, which separates messages from different hops, assigns different dimensions for them, and concatenates them to obtain node representations. They develop a conditionally progressive NAS strategy to explore an effective hop-dimension relationship and propose an efficient approximate hop-dimension relation function to facilitate rapid configuration of LADDER-GNN. Wei \etal~\cite{wei2022enhancing} propose Intra-class Information Enhanced Graph Neural Networks (IIE-GNN), in which the intra-class information can be extracted based on seven carefully designed blocks in four levels. A novel set of operations in each block is developed to construct the supernet for NAS and two architecture predictors are designed to select node-wise architectures to design the node-aware GNNs. Wei \etal~\cite{wei2022designing} develop F2GNN, which provides a framework to unify the existing topology designs with feature selection and fusion strategies. Based on the framework, a NAS method was built, which contains a set of selection and fusion operations in the search space and an improved differentiable search algorithm. Zheng \etal~\cite{zheng2023auto} develop Auto-HeG, a NAS method to automatically build heterophilic GNN models that incorporate heterophily in search space design, supernet training, and architecture selection. It constructs a comprehensive search space for various heterophily of graphs, dynamically shrinks the initial search space based on layer-wise heterophily variation to form an efficient supernet, and utilizes a heterophily-aware distance criterion to select specialized heterophilic GNN architectures in a leave-one-out fashion. Wei \etal~\cite{wei4825405searching} propose Heterophily-Agnostic Graph Neural Network (HAGNN), which includes a heterophily-agnostic search space for each node and an architecture controller to guide the operation selection in a self-taught way.

\subsubsection{Edge Directionality}
\label{sec:edge_directionality}
Rossi \etal~\cite{rossi2023edge} show that treating the edges as directed increases the effective homophily of the graph, suggesting a potential gain from the correct use of directionality information. They introduce Directed Graph Neural Network (Dir-GNN), which can be used to extend any MPNN to account for edge directionality information by performing separate aggregations of the incoming and outgoing edges. Sun \etal~\cite{sun2024breaking} develop AMUD, a method to analyze the correlation between node profiles and network topology, providing a framework to adaptively represent natural directed graphs as either undirected or directed to maximize their benefits in graph learning. Following this, they introduce Adaptive Directed Pattern Aggregation (ADPA), a new directed graph learning paradigm for AMUD. Koke \etal~\cite{koke2023holonets} use complex analysis and spectral theory to extend conventional spectral convolution to directed graph, and show that directed spectral convolutional networks can achieve SOTA results on heterophilic node classification.  Chaudhary \etal~\cite{chaudhary2024gnndld} propose Graph Neural Network with Directional Label Distribution (GNNDLD), a model that leverages edge direction and label distribution across different neighborhood hops, combines multi-layer features to maintain both low- and high-frequency components, and decouples node feature aggregation from transformation to prevent oversmoothing.

\subsubsection{Memory-based Methods}
\label{sec:memory_based_methods}
Chen \etal~\cite{chen2022memory} introduce a Memory-based Message Passing (MMP) mechanism that separates messages into memory and self-embedding components, allowing nodes to communicate with memory cells, and adaptively update their self-embeddings and memory cells through a learnable control mechanism. A decoupling regularization loss function is deployed to improve feature propagation and discrimination. Graph Memory Networks for Heterophilic Graphs (HP-GMN)~\cite{xu2022hp} incorporates local statistics that can effectively capture local heterophily information, and a memory module that can learn global patterns of the graph. Regularization methods are deployed to keep the memory units representative and diverse to learn high-quality global patterns. Zhao \etal~\cite{zhao2024disambiguated} introduce DisamGCL, a framework that uses a memory cell to encode the historical variance of predicted label distributions to dynamically identify regions of the graph with ambiguous node representations. To disambiguate these nodes and steer their representations away from the mixed semantic areas, the framework incorporates a contrastive learning objective, promoting stronger distinctions from their dissimilar neighbors.

\subsubsection{Remaining Methods}
\label{sec:remaining_methods}
This part covers some less widely used yet still important techniques.

\paragraph{Graph Capsule Network} NCGNN~\cite{yang2022ncgnn} proposes a vectorized node-level capsule (a group of neurons) for feature aggregation to efficiently preserve the distinctive properties of nodes, and develops a novel dynamic routing procedure that adaptively selects appropriate capsules for aggregation to remove noisy neighborhood information. It can prevent feature over-mixing and generate interpretable high-level capsules for each class.

\paragraph{Relation Encoder} 
Shi \etal~\cite{shi2022vr} introduce relation vectors into the message-passing framework, which are more flexible and expressive in modeling homophily and heterophily relations between nodes. Then, Variational Relation Vector Graph Neural Network (VR-GNN) is proposed, where the encoder generates relation vectors based on the graph structure, feature and label, and the decoder incorporates the relation vectors into the message-passing mechanism, where messages are computed by translating neighbors along connections. Wu \etal~\cite{wu2023beyond} introduce a relation-based frequency adaptive GNN (RFA-GNN), which can address heterophily and heterogeneity in a unified framework. It decomposes an input graph into several latent relation graphs, and develops a mechanism to adaptively capture signals across different frequencies in each corresponding relation space.

\paragraph{Deep Reinforcement Learning} 
Peng \etal~\cite{peng2023graphrare} introduce GraphRARE, a general GNN framework built upon node relative entropy and deep reinforcement learning. The node relative entropy measures node feature and structural similarity by mutual information between node pairs. A deep reinforcement learning-based algorithm is used to select informative nodes while filtering out noisy ones based on the relative entropy, which can refine graph topology and avoid mixing useful and noisy information.

\paragraph{Uncertainty-aware Model}
Liu \etal~\cite{liu2022ud} find that GNNs are severely biased towards homophilic nodes and develop an Uncertainty-aware Debiasing (UD) framework, which retains the knowledge of the biased model on certain nodes and compensates for the nodes with high uncertainty. Specifically, UD has an uncertainty estimation module to approximate the model uncertainty to recognize nodes with local heterophilic neighborhood. It trains a debiased GNN by pruning the biased parameters with certain nodes, and re-trains the pruned parameters from scratch on nodes with high uncertainty to mitigate the bias.

\paragraph{Adaptive Mixing of Multiple Modules}
Different modules encode different aspects of features and structural information, and can be  combined by a mixing method for heterophily. MWGNN~\cite{ma2022meta} proposes to divide graph convolution operations into three learnable components to encode node features, topological structure and positional identity, separately. These three matrices are then adaptively merged with an attention mechanism to learn different dependencies of the graph. Chen \etal~\cite{chen2023exploiting} introduce the Conv-Agnostic GNN framework (CAGNN), which decouples the node features into the discriminative component for downstream tasks and an aggregation feature for graph convolution. Then, a shared mixer module is developed to adaptively evaluate and incorporate neighbor information.

\paragraph{Masking} 
Gao \etal develop GraphTU~\cite{gao2023topology}, which employs an effective gradient-guided mask to block biased statistics from heterophilic neighbors without any modification to GNN architecture.

\paragraph{Bilevel Optimization}
Bilevel optimization is an approach with two interconnected levels, where the optimal solution from a lower-level energy function is utilized as the input to an upper-level objective. This method enhances the interpretability of model architectures, and Zheng \etal~\cite{zheng2024graph} model graph machine learning under the bilevel optimization framework. Then, they propose BloomGML and reveal how to calibrate the neighborhood embedding distance distribution to address heterophily problem.

\paragraph{Heterophilic Snowflake Hypothesis}
Wang \etal~\cite{wang2024snowflake} put forward the heterophily snowflake hypothesis, which is a transformation of the snowflake hypothesis~\cite{wang2023snowflake}, \ie{} “one node, one receptive field”, to heterophilic graphs. It states that each node can have its own unique aggregation hop and pattern, just like each snowflake is unique and possesses its own
characteristics. Based on this hypothesis, they employ a proxy label predictor, generating pseudo-label probability distributions for different nodes, which assists nodes in determining
whether they should aggregate their associated neighbors.

\section{Heterophily on Heterogeneous Graphs}
\label{sec:heterophily_on_heterogeneous_graphs}

As discussed in the earlier sections, the majority of current research on the heterophily problem is focused on homogeneous graphs. However, since the real-world graphs often demonstrate much more complicated relations, characterized by the various types of nodes and edges, and the heterogeneity in their global and local structures, it is non-trivial to explore heterophily in the context of heterogeneous graphs. In this section, we will give an outline of the models designed for heterogeneous graphs with heterophily and put forward future directions.

\subsection{Models}

\paragraph{Heterophily-Specific HetGNNs} Ahn \etal~\cite{ahn2022descent} introduce HALO, a framework where layers are derived from optimization steps descending a relation-aware energy function, with a minimizer that is fully differentiable with the energy function parameters. Specifically, HALO employs a compatibility matrix to model various heterophily relationships effectively. Li \etal~\cite{li2023hetero} develop Hetero$^2$Net, a heterophily-aware GNN that efficiently handles both homophilic and heterophilic heterogeneous graphs by integrating masked meta-path prediction and masked label prediction tasks. The masked meta-path prediction module learns disentangled homophilic and heterophilic representations and captures richer graph signals for downstream tasks. The masked label prediction is able to enhance message propagation among nodes that exhibit strong label heterophily. 

\paragraph{Graph Rewiring Method} Guo \etal~\cite{guo2023homophily} find that current HetGNNs may have performance degradation when handling HetGs with less homophilic properties, and they propose Homophily-oriented Deep Heterogeneous Graph Rewiring (HDHGR) to improve the performance of HetGNNs. A meta-path similarity learner (MSL) is used to learn pairwise similarity under different meta-paths, which can be used to rewire meta-path subgraphs by removing edges with low similarities and adding edges with high similarities.

\paragraph{Label Propagation} Deng \etal~\cite{deng2019label} explore the heterophily issue within heterogeneous graphs, introducing a $\mathcal{K}$-partite label propagation approach to address the complex interplay of heterogeneous nodes/relations and heterophily propagation. They formulate a comprehensive $\mathcal{K}$-partite label propagation model that accommodates both vertex-level and propagation-level heterogeneity.

\paragraph{Link Prediction on HetGs with Heterophily} Xiong \etal~\cite{xiong2023ground} explore the challenge of heterophily within heterogeneous graphs for link prediction tasks. They introduce the Pro-SEAL model, along with an innovative labeling technique designed to enhance the ability of GNNs to distinguish structural information in heterogeneous graphs. This labeling strategy effectively identifies (1) nodes at varying distances from the target nodes; (2) nodes with distinct labels but identical distances to the target nodes; and (3) the target nodes themselves. By incorporating heterophily, this approach can improve HetGNN performance.

\subsection{The Next Step}
\label{sec:heterogeneous_gnn_next}
In the near future, several important works need to be done before we start to put efforts on model development.

\begin{itemize}
    \item \textbf{Identify the heterophily issue for meta-path based HetGNNs} Although Guo \etal~\cite{guo2023homophily} did some preliminary experiments to study the relation between the homophily ratio of meta-path induced subgraphs and the performance of HetGNNs, their empirical observations were far from conclusive. The experiments only cover three datasets which are insufficient. The synergy among subgraphs is underexplored~\footnote{Different combinations of subgraphs with different homophily levels should be tested for questions such as "will 9 homophilic subgraphs and 1 heterophilic subgraphs lead to performance degradation of HetGNNs? How about 1 homophilic and 9 heterophilic subgraphs or 5 homophilic and 5 heterophilic subgraphs? And how low should the homophily values of subgraphs be (\eg{} 0.01, 0.1 or 0.3?) to cause inferior performance? Is there any synergy between the structures of subgraphs and node features?"} and the mid-homophily pitfall~\cite{luan2024graph} phenomenon should be re-defined and re-verified. We should first identify and conclude that heterophily is a valid and non-trivial challenge for HetGNNs before working on it.
    \item \textbf{Heterophilic benchmark datasets for HetGs} There already exist lots of benchmark datasets for HetGs. However, none of them are designed to study heterophily problem. Here are some references which can help to build up the new heterophilic benchmarks for HetGs~\cite{lv2021we, hu2021ogb, liu2023datasets, guo2023homophily, li2023hetero}.
    \item \textbf{Homophily metrics for HetGs} Several homophily metrics for HetGs~\cite{guo2023homophily, li2023hetero} have been proposed, adapting definitions originally established for homogeneous graphs. However, their effectiveness to indicate whether the graph-aware models can outperform graph-agnostic models has not been investigated. Better metrics for HetGs need to be designed in the future.
\end{itemize}

\section{Unsupervised Learning on Heterophilic Graphs}
\label{sec:unsupervised_learning}
\begin{figure*}[ht!]
\centering
\resizebox{1.0\textwidth}{!}{
\begin{forest}
    for tree={%
        my node,
        l sep+=5pt,
        grow'=east,
        edge={gray, thick},
        parent anchor=east,
        child anchor=west,
        if n children=0{tier=last}{},
        edge path={
            \noexpand\path [draw, \forestoption{edge}] (!u.parent anchor) -- +(10pt,0) |- (.child anchor)\forestoption{edge label};
        },
        if={isodd(n_children())}{
            for children={
                if={equal(n,(n_children("!u")+1)/2)}{calign with current}{}
            }
        }{}
    }
    [\textbf{Unsupervised Learning} \ref{sec:unsupervised_learning}, draw=black
        [\textbf{Contrastive Learning} \ref{sec:contrastive.learning}, draw=red, top color=red!10, bottom color=red!10
            [\textbf{Incorporation of High-Pass Filter}, draw=red, top color=red!30, bottom color=red!30
                [HLCL~\cite{yang2024contrastive}, draw=red]
                [S3GCL~\cite{wan2024s3gcl}, draw=red]
            ]
            [\textbf{Multi-View Contrastive Learning}, draw=red, top color=red!30, bottom color=red!30
                [MUSE~\cite{yuan2023muse}, draw=red]
                [GASSER~\cite{yang2023augment}, draw=red]
            ]
            [\textbf{Similarity-Based Methods}, draw=red, top color=red!30, bottom color=red!30
                [HomoGCL~\cite{li2023homogcl}, draw=red]
                [SimGCL~\cite{liu2024simgcl}, draw=red]
            ]
            [\textbf{Optimized Neighborhood Sampling}, draw=red, top color=red!30, bottom color=red!30
                [NeCo~\cite{he2023contrastive}, draw=red]
            ]
            [\textbf{Single-Pass Graph Contrastive Learning}, draw=red, top color=red!30, bottom color=red!30
                [SP-GCL~\cite{wang2023single}, draw=red]
            ]
            [\textbf{Beyond Local Information}, draw=red, top color=red!30, bottom color=red!30
                [GraphACL~\cite{xiao2024simple}, draw=red]
                [LRD-GNN~\cite{yang2024self}, draw=red]
            ]
        ]
        [\textbf{Other Unsupervised Learning Methods} \ref{sec:self.supervised.learning}, draw=blue, top color=blue!10, bottom color=blue!10
            [\textbf{Message Decoupling}, draw=blue, top color=blue!30, bottom color=blue!30
                [DSSL~\cite{xiao2022decoupled}, draw=blue]
                [MVGE~\cite{lin2023multi}, draw=blue]
            ]
            [\textbf{High-Frequency Signals}, draw=blue, top color=blue!30, bottom color=blue!30
                [PairE~\cite{li2022graph}, draw=blue]
            ]
            [\textbf{High-Order Neighborhood Information}, draw=blue, top color=blue!30, bottom color=blue!30
                [HGRL~\cite{chen2022towards}, draw=blue]
            ]
            [\textbf{Edge Discrimination}, draw=blue, top color=blue!30, bottom color=blue!30
                [GREET~\cite{liu2023beyond}, draw=blue]
            ]
            [\textbf{Other Self-Supervised learning Methods}, draw=blue, top color=blue!30, bottom color=blue!30
                [MGS~\cite{ding2023self}, draw=blue]
                [PGSL~\cite{chen2024pareto}, draw=blue]
            ]
        ]
    ]
\end{forest}
}
\caption{An overview of unsupervised learning of graph models on heterophilic graphs. We only include a small portion of unsupervised learning graph models for heterophily in this diagram, more models and details can be found in each section.}
\label{diag:unsupervised}
\end{figure*}
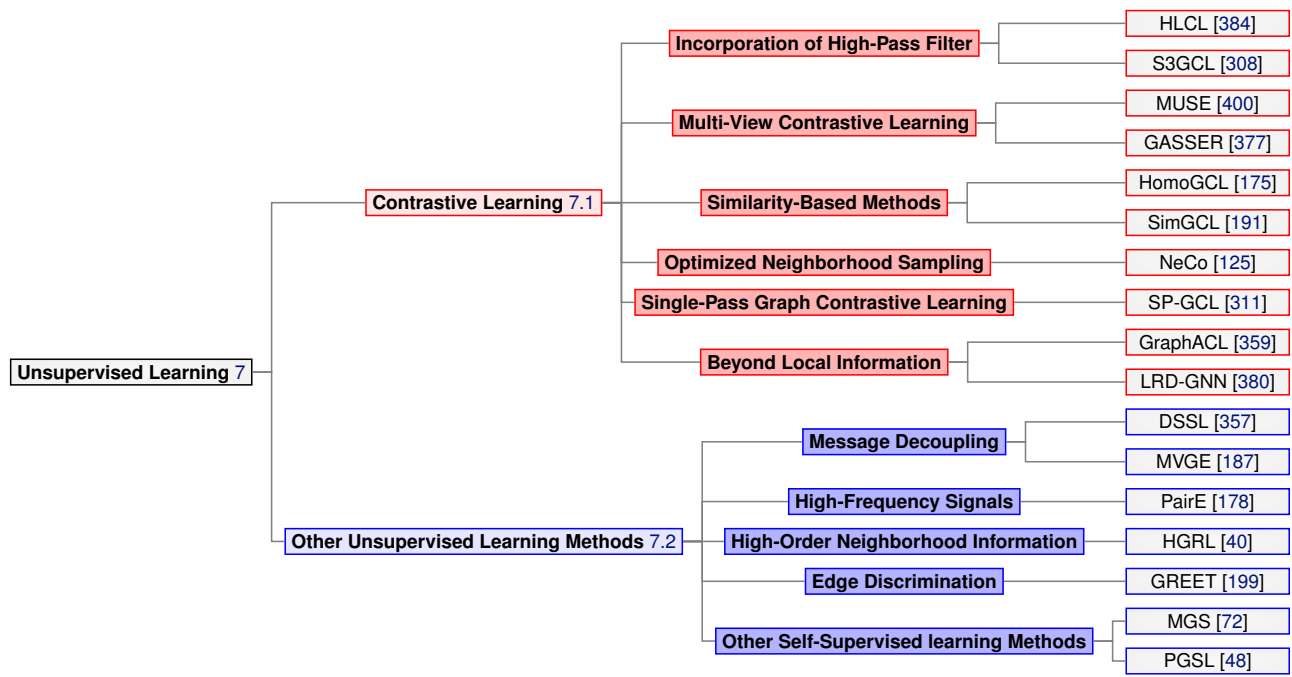

Graph unsupervised learning methods can learn node embeddings and discover underlying structures without labeled nodes~\cite{perozzi2014deepwalk, grover2016node2vec, velivckovic2018deep}. The idea of heterophily can be exploited in graph unsupervised learning, and in this section, we explore contrastive learning for graphs with heterophily in Section~\ref{sec:contrastive.learning} and summarize other unsupervised learning methods in Section~\ref{sec:self.supervised.learning}. In Figure~\ref{diag:unsupervised}, we present an overview of unsupervised learning models on heterophilic graphs. 


\subsection{Contrastive Learning}
\label{sec:contrastive.learning}
Graph Contrastive Learning (GCL) methods learn representations by contrasting positive samples against negative ones from graph data~\cite{you2020graph, hassani2020contrastive, zhu2020deep, zhu2021empirical}. Many GCL methods are developed for learning on heterophilic graphs.

\paragraph{Incorporation of High-Pass Filter}
Yang \etal~\cite{yang2024contrastive} propose HLCL, which uses a high-pass and a low-pass graph filter to generate different views of the same node. HLCL can capture both the dissimilarity and the similarity between neighboring nodes, which are important for modeling heterophilic graphs. Chen \etal~\cite{chen2023polygcl} propose PolyGCL, which uses spectral polynomial filters to generate different views of the graph, and then performs contrastive learning between the low-pass and high-pass views. PolyGCL can be used to capture both the smoothness and the diversity of graph features to deal with the heterophily issue. Wu \etal~\cite{wu2023beyond} propose RFA-GNN, which adopts a frequency-adaptive mechanism to flexibly provide low- and high-frequency signals for graphs with different heterophily ratios. Wo \etal~\cite{wo2024graph} propose a GCL framework to generate homophilic and heterophilic views through interventional view generation, which can disentangle homophilic and heterophilic structure from the original graph. Low-pass and high-pass filters are used to encode the homophilic and heterophilic views separately for GCL. Wan \etal~\cite{wan2024s3gcl} propose Spectral, Swift, Spatial Graph Contrastive Learning (S3GCL), which deploys a cosine-parameterized Chebyshev polynomial to selectively emphasize certain high/low frequency ranges, effectively decomposing graphs into two biased-pass views.

\paragraph{Multi-View Contrastive Learning}
Yuan \etal~\cite{yuan2023muse} propose MUSE, which leverages multi-view contrastive learning and information fusion to capture both the local and global information of nodes. Khan \etal~\cite{khan2023contrastive} introduce a novel multi-view method that integrates diffusion filters on graphs to capture high-order graph structures and structural equivalence in heterophilic graphs.  Yang \etal~\cite{yang2023augment} recently argue that existing GCL methods that use random topology corruption to generate augmentation views are ineffective for heterophilic graphs. Therefore, they propose GASSER, which utilizes selective spectrum perturbation to tailor the impact of edge perturbation on different frequency bands of graph structures. GASSER can generate augmentation views that are adaptive, controllable, and invariant to the key information, while discarding the unimportant part.

\paragraph{Similarity-Based Methods}
Li \etal~\cite{li2023homogcl} devise HomoGCL, which calculates node similarities to obtain node soft clustering assignments. The assignment scores indicate the probabilities that the neighbors are true positive samples. Liu \etal~\cite{liu2024simgcl} propose a novel Similarity-based Graph Contrastive Learning model (SimGCL), which handles heterophilic graphs by generating augmented views with higher homophily ratios at the topological level.

\paragraph{Optimized Neighborhood Sampling}
He \etal~\cite{he2023contrastive} introduce a new parameterized neighbor sampling component that can dynamically adjust the size and composition of the neighbor sets for each node. This component can optimize both the positive sampling of contrastive learning and the message passing of graph neural networks. 

\paragraph{Single-Pass Graph Contrastive Learning} Wang \etal~\cite{wang2023single} argue that the existing graph contrastive learning methods require two forward passes for each node to construct the contrastive loss, which is costly and may not work well on heterophilic graphs. Thus, they propose a new Single-Pass Graph Contrastive Learning method (SP-GCL) that leverages the concentration property of features obtained by neighborhood aggregation on both homophilic and heterophilic graphs, and further introduces a single-pass GCL loss based on this property.

\paragraph{Beyond Local Information} Xiao \etal~\cite{xiao2024simple} propose Asymmetric Contrastive Learning for Graphs (GraphACL), which leverages an asymmetric view of the neighboring nodes, where the center node is encoded by a GNN model and the neighbor node is encoded by a perturbed version of the same model. GraphACL can capture both local information and global monophily similarity, which are important for modeling heterophilic graphs. Additionally, Yang \etal~\cite{yang2024self} propose Low-Rank Tensor Decomposition-based GNN (LRD-GNN-Tensor), which  constructs the node attribute tensor and performs low-rank tensor decomposition to incorporate important long-distance information for heterophilic graphs.

\subsection{Other Unsupervised Learning Methods}
\label{sec:self.supervised.learning}

\subsubsection{Message Decoupling}
Xiao \etal~\cite{xiao2022decoupled} develop a Decoupled Self-Supervised Learning (DSSL) framework for heterophilic graphs, which decouples the semantic structure of the graph into latent variables that capture different aspects of node similarities, such as attribute similarity, label similarity, and link similarity. Then, DSSL uses these latent variables to generate self-supervised tasks for node representation learning, such as reconstructing node attributes, predicting node labels, and inferring node links. Zhong \etal~\cite{zhong2022unsupervised} propose SELf-supErvised Network Embedding (SELENE) framework for learning node embeddings on both homophilic and heterophilic graphs. SELENE addresses the heterophily issue by formulating the unsupervised network embedding task as an $r$-ego network discrimination problem, where $r$-ego networks are subgraphs centered around a node within a radius $r$. SELENE uses a dual-channel feature embedding pipeline to discriminate $r$-ego networks based on node attributes and structural information separately. Lin \etal~\cite{lin2023multi} propose Multi-View Graph Encoder (MVGE) for unsupervised graph representation learning. MVGE adopts a multi-view perspective and uses diverse pretext tasks to capture different signals in the graph, such as node attributes, node degrees, and node clustering coefficients. MVGE also introduces a simple operation to balance the commonality and specificity of the node embeddings in both the attribute and structural levels. 

\subsubsection{High-Frequency Signals}
Li \etal~\cite{li2022graph} propose an unsupervised graph embedding method called PairE, which uses two paired nodes as the basic embedding unit to capture both high-frequency and low-frequency signals between nodes. PairE consists of a multi-self-supervised autoencoder that performs two important pretext tasks: one to retain the high-frequency signal by discriminating ego networks, and another to enhance the representation of commonality by reconstructing node attributes. 

\subsubsection{High-Order Neighborhood Information} Chen \etal~\cite{chen2022towards} propose HGRL, which learns the node embeddings by preserving the original node features and capturing informative distant neighbors. These two important properties are achieved by employing well-crafted pretext tasks that are optimized with estimated high-order mutual information.

\subsubsection{Edge Discrimination}
Liu \etal~\cite{liu2023beyond} introduce Graph Representation learning method with Edge
hEterophily discriminaTing (GREET), which learns node embeddings by discriminating homophilic and heterophilic edges. GREET addresses the heterophily issue by using an edge discriminator to infer edge homophily/heterophily from feature and structure information, and a dual-channel feature embedding to contrast the encodings of homophilic and heterophilic edges.

\subsubsection{Other Graph Self-Supervised Learning (GSSL) Methods} 
Wu \etal~\cite{wu2023homophily} reveal that GSSL and GNNs have consistent optimization goals in terms of improving the graph homophily. They propose Homophily-Enhanced Self-supervision for Graph Self-supervised Learning (HES-GSL), which leverages the homophily property of graphs to provide more self-supervision for learning the graph structure, especially when the labeled data are scarce. Ding \etal~\cite{ding2023self} first propose Measure Graph Structure (MGS), a metric to measure the graph structure, which analyzes the correlation between the structural similarity and the embedding similarity of graph pairs. Furthermore, they propose Pre-training GNNs based on the MGS Metric (PGM) method, which can enhance the graph structural information captured by the GNNs. Chen \etal~\cite{chen2024pareto} explore the problem of multi-tasking GSSL on heterophilic graphs and propose Pareto Graph Self-supervised Learning (PGSL), which makes a trade-off between the main tasks and auxiliary pretext tasks. Specifically, PGSL formulates the multi-tasking GSSL as a multi-objective optimization problem and solves it with preferred vector and personalized optimization. Park \etal~\cite{park2024enhancing} propose Label Propagation on k-Nearest Neighbor Graphs (LPkG), an efficient approach that first constructs a supplementary heterophilic graph, and uses it to perform label propagation to produce self-supervision for GNN models. To overcome the lack of supervision problem in GSSL, Yang \etal~\cite{yang2024gauss} propose GrAph-customized Universal Self-Supervised Learning (GAUSS) that exploits local attribute distributions by replacing the global parameters with locally learnable propagation. Its superior performance and robustness to noise have been extensively verified on both homophilic and heterophilic graphs.

\section{Theoretical Understanding of Homophily and Heterophily} 
\label{sec:theoretical_understanding}
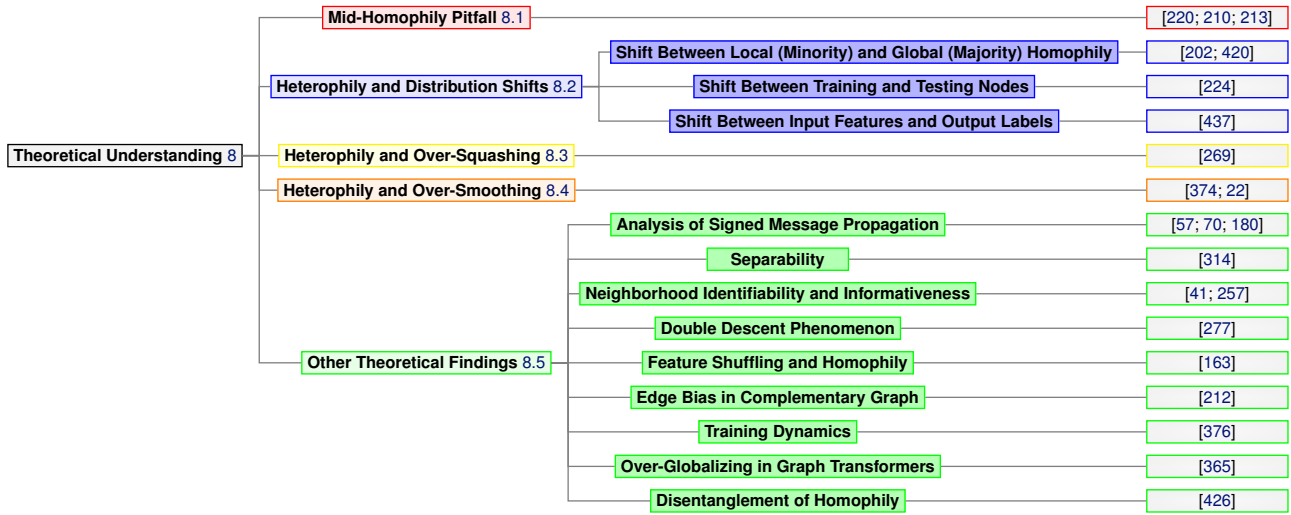
\begin{figure*}[ht!]
\centering
\resizebox{1.0\textwidth}{!}{
\begin{forest}
    for tree={%
        my node,
        l sep+=5pt,
        grow'=east,
        edge={gray, thick},
        parent anchor=east,
        child anchor=west,
        if n children=0{tier=last}{},
        edge path={
            \noexpand\path [draw, \forestoption{edge}] (!u.parent anchor) -- +(10pt,0) |- (.child anchor)\forestoption{edge label};
        },
        if={isodd(n_children())}{
            for children={
                if={equal(n,(n_children("!u")+1)/2)}{calign with current}{}
            }
        }{}
    }
    [\textbf{Theoretical Understanding} \ref{sec:theoretical_understanding}, draw=black
        [\textbf{Mid-Homophily Pitfall} \ref{sec:homo.pitfall}, draw=red, top color=red!10, bottom color=red!10
            [\cite{ma2021homophily, luan2022revisiting, luan2024graph}, draw=red]
        ]
        [\textbf{Heterophily and Distribution Shifts} \ref{sec:distribution.shift}, draw=blue, top color=blue!10, bottom color=blue!10
            [\textbf{Shift Between Local (Minority) and Global (Majority) Homophily}, draw=blue, top color=blue!30, bottom color=blue!30
                [\cite{loveland2024performance, zhao2024disambiguated}, draw=blue]
            ]
            [\textbf{Shift Between Training and Testing Nodes}, draw=blue, top color=blue!30, bottom color=blue!30
                [\cite{mao2023demystifying}, draw=blue]
            ]
            [\textbf{Shift Between Input Features and Output Labels}, draw=blue, top color=blue!30, bottom color=blue!30
                [\cite{zhu2023explaining}, draw=blue]
            ]
        ]
        [\textbf{Heterophily and Over-Squashing} \ref{sec:heterophily_oversquashing}, draw=yellow, top color=yellow!10, bottom color=yellow!10
            [\cite{rubin2023geodesic}, draw=yellow]
        ]
        [\textbf{Heterophily and Over-Smoothing} \ref{sec:heterophily_oversmoothing}, draw=orange, top color=orange!10, bottom color=orange!10
            [\cite{yan2022two, bodnar2022neural}, draw=orange]
        ]
        [\textbf{Other Theoretical Findings} \ref{sec:other.theory.finding}, draw=green, top color=green!10, bottom color=green!10
            [\textbf{Analysis of Signed Message Propagation}, draw=green, top color=green!30, bottom color=green!30
                [\cite{choi2023signed, di2023understanding, liang2024sign}, draw=green]
            ]
            [\textbf{Separability}, draw=green, top color=green!30, bottom color=green!30
                [\cite{wang2024understanding}, draw=green]
            ]
            [\textbf{Neighborhood Identifiability and Informativeness}, draw=green, top color=green!30, bottom color=green!30
                [\cite{chen2023exploiting, platonov2023characterizing}, draw=green]
            ]
            [\textbf{Double Descent Phenomenon}, draw=green, top color=green!30, bottom color=green!30
                [\cite{shi2024statistical}, draw=green]
            ]
            [\textbf{Feature Shuffling and Homophily}, draw=green, top color=green!30, bottom color=green!30
                [\cite{lee2024feature}, draw=green]
            ]
            [\textbf{Edge Bias in Complementary Graph}, draw=green, top color=green!30, bottom color=green!30
                [\cite{luan2023we}, draw=green]
            ]
            [\textbf{Training Dynamics}, draw=green, top color=green!30, bottom color=green!30
                [\cite{yang2024dynamics}, draw=green]
            ]
            [\textbf{Over-Globalizing in Graph Transformers}, draw=green, top color=green!30, bottom color=green!30
                [\cite{xing2024less}, draw=green]
            ]
            [\textbf{Disentanglement of Homophily}, draw=green, top color=green!30, bottom color=green!30
                [\cite{zheng2024missing}, draw=green]
            ]
        ]
    ]
\end{forest}
}
\caption{An overview of theoretical understanding of graph homophily and heterophily.}
\label{diag:theoretical_understanding}
\end{figure*}

It is commonly believed that homophily is beneficial for graph learning, and the heterophilic structure has a negative effect on model performance. However, both empirical (See Section ~\ref{sec:homogeneous_benchmark}) and theoretical results in recent studies indicate that the impact of homophily and heterophily is much more complicated than "homophily wins, heterophily loses"~\cite{ma2021homophily,luan2022revisiting, luan2024graph}. Researchers have different understandings about homophily/heterophily and the controversies are widely discussed in the graph community in recent years. In this section, we will summarize recent theoretical studies in different aspects, including mid-homophily pitfall (Section~\ref{sec:homo.pitfall}), distribution shifts (Section~\ref{sec:distribution.shift}), heterophily and over-squashing~\ref{sec:heterophily_oversquashing}, heterophily and over-smoothing~\ref{sec:heterophily_oversmoothing}, and other interesting theoretical findings (Section~\ref{sec:other.theory.finding}), such as separability, double descent phenomenon, training dynamics, over-globalization and homophily disentanglement, \etc In Figure~\ref{diag:theoretical_understanding}, we present an overview of this section.

\subsection{Mid-Homophily Pitfall}
\label{sec:homo.pitfall}
Contrary to the prevalent belief that a graph with the lowest homophily value has the most negative impact on the performance of GNNs, a series of studies have revealed different relationships between homophily and GNN performance. For example, the authors in~\cite{ma2021homophily} state that, as long as nodes within the same class share similar neighborhood patterns, their embeddings will be similar after aggregation. They provide experimental evidence and theoretical analysis, and conclude that homophily may not be necessary for GNNs to distinguish nodes. Luan \etal~\cite{luan2022revisiting} study homophily/heterophily from post-aggregation node similarity perspective and find that heterophily is not always harmful, which is consistent with ~\cite{ma2021homophily}.  Luan \etal~\cite{luan2024graph} point out the deficiency of the current understanding of homophily, which only considers the intra-class node distinguishability (ND). They investigate the impact of homophily principle by considering both intra- and inter-class ND and propose new classifier-based performance metrics. Luan \etal~\cite{luan2024graph} are the first to formally and quantitatively study the relation between homophily and ND, and prove that the medium level of homophily actually has more detrimental effect on ND than extremely low level of homophily. They formally name this phenomenon as the \textbf{mid-homophily pitfall}.

\subsection{Heterophily and Distribution Shifts}
\label{sec:distribution.shift}
There is a line of research that studies the impact of heterophily with distribution (neighborhood patterns) shifts from various perspectives.

\paragraph{Shifts Between Local (Minority) and Global (Majority) Homophily} Loveland \etal~\cite{loveland2024performance} go beyond global homophily and study GNN performance when the local homophily level of a node deviates from the global homophily level. They demonstrate how the shifts in local homophily of nodes in training and test sets can introduce performance degradation, leading to performance discrepancies across local homophily levels. Specifically, they reveal that: (1) GNNs can fail to generalize to test nodes that deviate from the global homophily of a graph, presenting a challenge for nodes with underrepresented homophily levels to be correctly predicted, and (2) high local homophily does not necessarily guarantee high performance for a node. Zhao \etal~\cite{zhao2024disambiguated} find that for nodes belonging to majority classes, an increase in heterophily leads to a substantial drop in performance. However, for nodes in minority and middle-sized classes, higher heterophily does not necessarily lead to performance drop; sometimes, it may even enhance it.

\paragraph{Shifts Between Training and Test Nodes}
Mao \etal~\cite{mao2023demystifying} provide evidence that GNNs actually perform admirably on nodes with the majority neighborhood patterns, \ie{} on homophilic nodes within homophilic graphs and heterophilic nodes within heterophilic graphs, while struggling on nodes with the minority patterns, exhibiting a performance disparity. The authors propose a rigorous \niid PAC-Bayesian generalization bound, revealing that both aggregated feature distance and homophily ratio differences between training and test nodes are key factors leading to the performance disparity.

\paragraph{Shifts Between Input Features and Output Labels} 
Zhu \etal~\cite{zhu2023explaining} quantify the magnitude of conditional shifts between the input features and the output labels and find that both graph heterophily and model architecture exacerbate conditional shifts, leading to the degradation of generalization capabilities. To address this, they propose GCONDA, a graph conditional shift adaptation method that (1) estimates the conditional shifts as the Wasserstein distance between source label distribution and estimated pseudo-label distribution; (2) minimizes the conditional shifts for unsupervised domain adaptation on graphs. 

\subsection{Heterophily and Over-Squashing}
\label{sec:heterophily_oversquashing}
For the tasks that require $K$-hop neighborhood information, the GNN needs to have at least $K$ layers to allow nodes to receive information from other nodes at a radius of $K$~\footnote{In graph theory, the radius of a graph is the minimum among all the maximum distances between a node to all other nodes~\cite{gross2018graph}.}; otherwise, the GNN will suffer from under-reaching and the distant nodes will not be aware of the messages from each other. However, as the number of layers increases, the size of the receptive field (multi-hop neighborhood set) for each node grows exponentially. This will lead to \textbf{over-squashing}: information from the exponentially-growing receptive field is compressed into fixed-length node vectors, which will cause information loss for messages from distant nodes~\cite{alon2020bottleneck}. For GNNs, the Jacobian of node representations has been defined to measure over-squashing~\cite{topping2021understanding}, with low values of the Jacobian indicating poor information flow, \ie{} bottleneck limit. 

With the Jacobian matrix, Rubin \etal~\cite{rubin2023geodesic} develop a unified theoretical framework to understand the combined effect of heterophily and over-squashing on the performance of GNNs. Such effect is named homophilic bottlenecking, \ie{} bottlenecking between nodes of the same type, which provides a theoretical tool to investigate the poor performance of GNNs on heterophilic graphs. To study homophilic bottlenecking, they decompose the performance of GNNs into two aspects measured by two metrics: (1) expressive power, measured by signal sensitivity, which is related to graph homophily; (2) generalization power, measured by noise sensitivity, which should be minimized. Homophilic bottlenecking is captured when they relate signal sensitivity to the graph structure through $l$-order homophily. With the distribution of geodesic distances and assumptions on the distribution of graph structures, they can decouple the bottleneck into under-reaching and over-squashing in a $l$-layer GNN. A complete analysis of homophilic bottlenecking can be derived by the tight bounds on $l$-order homophily, which are established with an asymptotic distribution of geodesic distances in a general random graph family.

\subsection{Heterophily and Over-Smoothing}
\label{sec:heterophily_oversmoothing}
As the depth of GNN gets deeper, the node representations are gradually smoothed out and information loss will occur. This phenomenon is called over-smoothing \cite{li2018deeper, oono2020graph, luan2019break, luan2024addressing}. Yan \etal~\cite{yan2022two} take a unified view to explain the over-smoothing and heterophily problems simultaneously by profiling nodes with two metrics: the relative degree of a node (compared to its neighbors) and the node-level heterophily. Based on the metrics, they develop theoretical framework to identify three distinct cases of node behavior, offering explanations for over-smoothing and heterophily challenges and predicting GCN performance. Bodnar \etal~\cite{bodnar2022neural} demonstrate that by applying the right sheaf structure for the task on graph, over-smoothing and heterophily problems can both be avoided effectively.

\paragraph{Remark} Note that over-smoothing only happens in deep GNNs, but not to shallow GNNs. Heterophily will cause performance degradation to all GNN models, not matter they are deep or shallow.

\subsection{Other Theoretical Findings}
\label{sec:other.theory.finding}
There are some other interesting theoretical findings, unveiling different aspects of homophily and its related topics.

\subsubsection{Analysis of Signed Message Propagation}

Signed message propagation is one of the most commonly used techniques for heterophily. Choi \etal~\cite{choi2023signed} provide a new understanding of signed propagation in multi-class scenarios and point out two drawbacks of it: (1) if two nodes belong to different classes but have a high similarity, signed propagation can decrease the separability; (2) signed propagation contributes to the ego-neighbor separation, but it also increases the uncertainty (\eg{} conflict evidence) of the predictions, which can impede the stability of the algorithm. Di \etal~\cite{di2023understanding} uncover that the negative edge-weight mechanisms can be simply achieved by residual graph convolutional with a channel mixing matrix whose eigenvalues are negative. Liang \etal~\cite{liang2024sign} identify two limitations of signed message passing: undesirable representation update for multi-hop neighbors and vulnerability against oversmoothing issues.

\subsubsection{Separability}
Wang \etal~\cite{wang2024understanding} study the effect of heterophily with a newly proposed Heterophilic Stochastic Block Models (HSBM), and demonstrate that the effectiveness of graph convolution (GC) operations in enhancing separability is determined by the Euclidean distance of the neighborhood distributions and the square root of the average node degree. Furthermore, they find that topological noise negatively affects separability by effectively lowering the average node degree. The separability gains rely on the normalized distance of $l$-hop neighborhood distributions after applying multiple GC operations, and the nodes still possess separability as $l$ goes to infinity in a wide range of regimes. 

\subsubsection{Neighborhood Identifiability and Informativeness}
Chen \etal~\cite{chen2023exploiting} investigate heterophily from a neighbor identifiable perspective, and state that heterophily can be helpful for node classification when the neighbor distributions of intra-class nodes are identifiable. This is similar to~\cite{ma2021homophily}. Platonov \etal~\cite{platonov2023characterizing} define and measure how informative the neighborhood label is for determining the label of the target node, and reveal that the label informativeness strongly correlates to GNN performance.

\subsubsection{Double Descent Phenomenon}
Double descent refers to a phenomenon observed in the learning curve of test risk versus model complexity, where increasing complexity first reduces test risk, and further increase lead to higher risk due to overfitting. Yet, continuing to scale up complexity beyond the point of interpolation into over-parameterized regime can result in a decreasing risk again~\cite{belkin2019reconciling}, which deviates from the traditional bias-variance trade-off~\cite{hastie2009elements}. Double descent was discovered around 1990~\cite{vallet1989linear, opper1990ability, seung1992statistical, watkin1993statistical}, and its significance has been further recognized in modern machine learning~\cite{belkin2019reconciling}, especially in deep learning~\cite{nakkiran2021deep, viering2022shape}.

Shi \etal~\cite{shi2024statistical} are the first to study double descent in graph learning and explore its relationship with homophily/heterophily. They use analytical tools from statistical physics and random matrix theory to precisely characterize generalization in simple graph convolution networks on the contextual stochastic block model. They illuminate the nuances of learning on homophilic vs. heterophilic data and predict the existence of double descent in GNNs. They show how risk is shaped by the interplay between the graph noise, feature noise, and the number of training labels, and expose that a negative self-loop results in much better performance on heterophilic graphs.

\subsubsection{Feature Shuffling and Homophily} Lee \etal~\cite{lee2024feature} empirically find that the performance of GNNs can be significantly enhanced via feature shuffling, and they explore "how randomly shuffling feature vectors among nodes from the same class affect GNNs" from homophily perspective. From their analysis, they argue that feature shuffling perturbs the dependency between graph topology and features (A-X dependence), and they define a metric to measure such dependency. Based on the metric and experimental results, they disclose that reducing A-X dependence with the feature shuffling consistently and substantially boosts GNN performance in high homophilic graphs, while the improvement is smaller in graphs with lower class homophily values.

\subsubsection{Edge Bias in Complementary Graph}
Luan \etal~\cite{luan2023we} derive two metrics to measure the effect of edge bias based on signal smoothness, and then conduct hypothesis testing for the edge bias in original graph versus complementary graph. They find that, when the labels in original graph are not significantly smoother than those in complementary graph, baseline GNNs will underperform MLP, no matter how smooth the node features are in the original graph.

\subsubsection{Training Dynamics}
Yang \etal~\cite{yang2024dynamics} study the training dynamics of GNNs in function space and find that the gradient descent optimization of GNNs implicitly leverages the graph structure to
update the learned function, which can be quantified by a phenomenon called kernel-graph alignment. Theoretically, they establish a strong correlation between generalization and homophily by a data-dependent generalization bound that highly depends on the graph homophily, showing that GNNs are the Bayesian optimal prior model architecture that minimizes the population risk on homophilic graphs. Empirically, they find that, on homophilic graphs, alignment with the graph promotes alignment with the optimal kernel matrix and thus improves generalization, whereas on heterophilic graphs, it adversely affects generalization. These results provide interpretable insights into when and why the learned GNN functions generalize, and highlight the limitations of GNNs on heterophilic graphs.

\subsubsection{Over-Globalizing in Graph Transformers}
Xing \etal~\cite{xing2024less} find that, in practice, the majority of the attention weights in graph transformer are allocated to distant high-order neighbors, regardless of homophilic or heterophilic graphs. They identify such limitation in the global attention mechanism as the over-globalizing problem and conduct theoretical analysis to study its impact on graph transformers.

\subsubsection{Disentanglement of Homophily}
To find out the missing parts in homophily to interpret GNN performance, Zheng \etal~\cite{zheng2024missing} disentangle the effect of homophily into three different aspects: label, structural, and feature homophily, which correspond to the three basic elements of graph data. The synergy between these three components provides a more complete view of the impact of homophily. Based on theoretical analysis, they derive a new composite metric, named Tri-Hom, which considers all three aspects and is verified to have significantly higher correlations with GNN performance than the existing metrics.

\begin{figure*}[htbp!]
    \centering
     \includegraphics[width=1\textwidth]{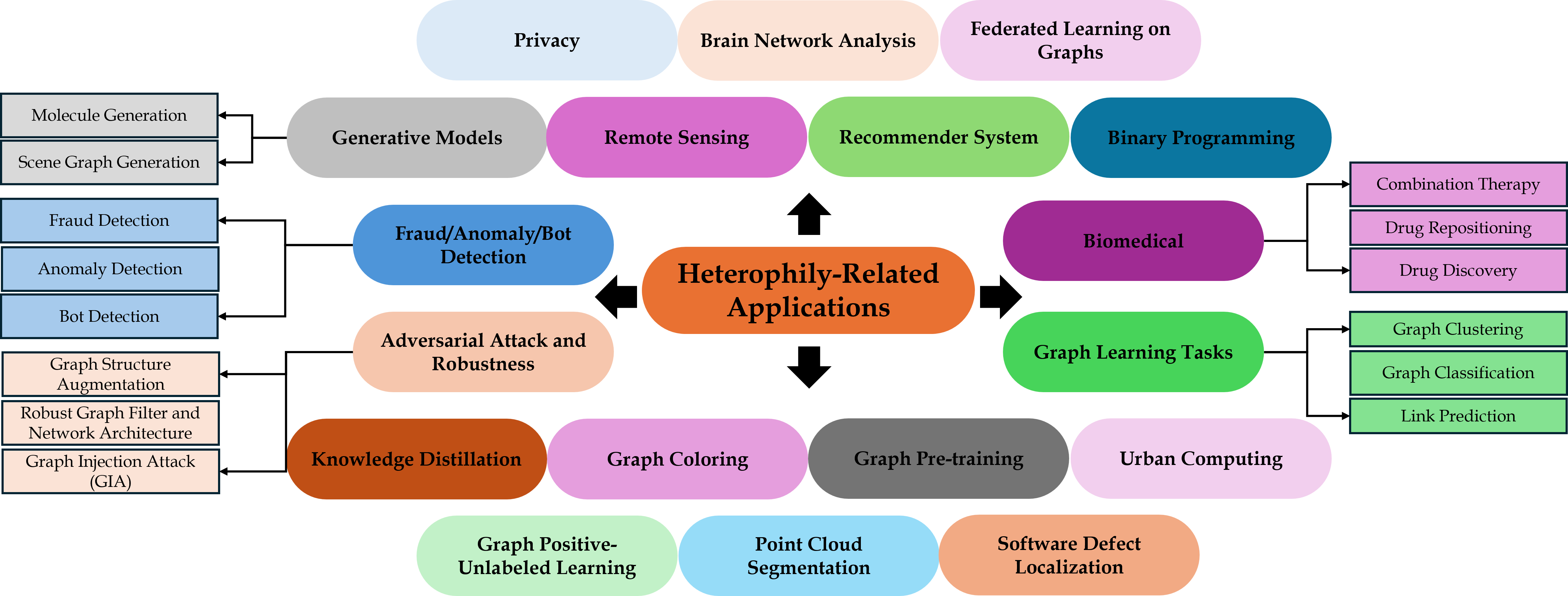}
     \vspace{-0.3cm}
     \caption{An overview of heterophily-related applications in many domains.}
     \label{fig:related_applications}
\end{figure*}

\section{Heterophily-Related Applications}
\label{sec:related_applications}
Numerous real-world applications can be modeled as graphs~\cite{ying2018graph, shlomi2020graph, sanchez2020learning, derrow2021eta, zhao2021consciousness, bongini2021molecular, huamulti2022, lu2024gcepnet}, and the heterophily challenge is potentially relevant to most of them. In this section, we will explore the heterophily-related applications as broad as possible, including:
\begin{itemize}
    \item Section~\ref{sec:detection_tasks}: fraud/anomaly/bot detection 
    \item Section~\ref{sec:generative_models}: generative models 
    \item Section~\ref{sec:graph_learning_tasks}: graph learning tasks
    \item Section~\ref{sec:recommender_system}: recommender system 
    \item Section~\ref{sec:biomedical}: biomedical 
    \item Section~\ref{sec:knowledge_distillation}: knowledge distillation 
    \item Section~\ref{sec:adversarial_attack_robustness}: adversarial attack and robustness 
    \item Section~\ref{sec:privacy}: privacy 
    \item Section~\ref{sec:urban_computing}: urban computing 
    \item Section~\ref{sec:graph_coloring}: graph coloring 
    \item Section~\ref{sec:binary_programming}: binary programming 
    \item Section~\ref{sec:remote_sensing}: remote sensing 
    \item Section~\ref{sec:positive_unlabeled_learning}: graph positive-unlabeled learning 
    \item Section~\ref{sec:point_cloud_segmentation}: point cloud segmentation 
    \item Section~\ref{sec:federated_learning}: federated learning 
    \item Section~\ref{sec:software_detect_localization}: software defect localization 
    \item Section~\ref{sec:brain_network_analysis}: brain networks analysis 
    \item Section~\ref{sec:prompt_pretraining}: graph pre-training
\end{itemize}
In Figure~\ref{fig:related_applications}, we present an overview of this section. 

\subsection{Fraud/Anomaly/Bot Detection}
\label{sec:detection_tasks}
On graph-structured data, heterophily is a common and challenging property for detection tasks, such as fraud/anomaly/bot detection, \etc{} These tasks aim to identify and isolate the abnormal and malicious nodes or subgraphs in the network, which can have significant impacts on the security, privacy and robustness for real-world applications on network data, \eg{} social networks, web graphs and citation networks.

\subsubsection{Fraud Detection}
\label{sec:fraud_detection}
In graph-based fraud detection (GFD) tasks, such as internet scams, data breaches, and business email compromises, nodes can be categorized as either fraudulent or benign. Fraud nodes in the networks are usually surrounded by normal nodes that have been cheated. This inherently heterophilic structure plays an important role for fraud detection~\cite{liu2024arc} and some works have taken neighborhood heterophily into account when designing GNNs for this task.

\paragraph{Neighborhood Partitioning and Grouping}
Shi \etal~\cite{shi2022h2} propose a GNN-based Fraud Detector with homophilic and heterophilic interactions (H$^2$-FDetector), which identifies the homophilic and heterophilic connections in the graph using the supervision of labeled nodes, and designs different aggregation strategies for them. Wang \etal~\cite{wang2023label} propose Group Aggregation enhanced Transformer (GAGA), which leverages the label information of nodes to enhance the node representations and the message passing process. GAGA incorporates a group aggregation module to aggregate the messages from different groups of neighbors with distinct functions. Kim \etal~\cite{kim2023dynamic} introduce a Dynamic Relation-Attentive Graph neural network (DRAG), which learns a node representation for each relation type and aggregates them using a dynamic relation-attentive mechanism that assigns different weights to each relation. Zhuo \etal~\cite{zhuo2024partitioning} argue that the key to applying GNNs for GFD is not to exclude but to distinguish neighbors with different labels during message passing, as fraud graphs often exhibit a mixture of heterophily and homophily. Thus, they introduce Partitioning Message Passing (PMP), which aggregates neighbors in different classes with distinct node-specific aggregation functions. Duan \etal~\cite{duan2024dga} present the Dynamic Grouping Aggregation GNN (DGA-GNN) for fraud detection, which tackles the non-additivity challenges of specific node attributes and the need for discernibility in messages grouped from neighboring nodes. They propose a decision tree binning encoding method to transform non-additive node attributes into bin vectors, which can avoid nonsensical feature generation and align well with the GNN’s aggregation operation. Furthermore, they develop a feedback dynamic grouping strategy to classify graph nodes into two distinct groups based on their labels, and then employ a hierarchical aggregation method to extract more discriminative features for GFD tasks.

\paragraph{Spectral-Based Methods}
Xu \etal~\cite{xu2023revisiting} propose Spectrum-Enhanced and Environment-Constrainted Graph-based Fraud Detection (SEC-GFD) that consists of a hybrid filtering module and a local environmental constraint module. The hybrid filtering module divides the spectrum of the graph into multiple mixed frequency bands, which can capture the heterophily of fraud graphs better than existing GNNs that assume homophily. Wu \etal~\cite{wu2023splitgnn} analyze the spectral distribution of graphs with different heterophily degrees and observe that the heterophily of fraud nodes leads to the spectral energy moving from low-frequency to high-frequency. Then, they introduce SplitGNN, a novel approach that employs an edge classifier to divide the original graph into two subgraphs with heterophilic and homophilic edges. SplitGNN adopts flexible band-pass graph filters to learn node representations from the subgraphs separately, and then aggregates them to obtain the final representations.

\subsubsection{Anomaly Detection}
\label{sec:anomaly_detection}
Graph-based Anomaly Detection (GAD) refers to identifying the rare observations that deviate significantly from the majority of objects in relational and structured data~\cite{ma2021comprehensive,wang2021bipartite}. In GAD, abnormal nodes are sparse and connected to the majority of normal nodes, which leads to the heterophily problem~\footnote{Note that GFD and GAD do have some overlapping. For example, fraudulent are also sparse and often connected to the majority of benign users.}. The existing aggregation-based GNNs smooth the features of neighboring nodes blindly, and thus undermine the discriminative information of the anomalies~\cite{gao2023addressing} and cause difficulty in identifying them. Recently, some methods have been proposed to leverage heterophily property to enhance the performance of GNNs for GAD.

\paragraph{Graph Structure Augmentation}
Gong \etal~\cite{gong2023beyond} propose Sparse Graph Anomaly Detection (SparseGAD), which sparsifies the graph structures to effectively reduce noises from task-irrelevant edges and discover the closely related nodes. It can discover the underlying dependencies between nodes in terms of homophily and heterophily. Zhang \etal~\cite{zhang2024generation} find that in GAD, the variance of class-wise homophily distributions in heterophilic graphs is significantly greater than that in homophilic graphs, and they introduce Class Homophily Variance to describe this phenomenon quantitatively. To mitigate its impact, they propose Homophily Edge Generation Graph Neural Network (HedGe), which generates new connections with low class homophily variance. To alleviate the deceptive homophily connections among anomalies, Wen \etal~\cite{wen2024ta} propose TA-Detector, which introduces a trust classier to distinguish trust and distrust connections.

\paragraph{Metric-Guided Detection}
Gao \etal~\cite{gao2023addressing} argue that heterophily is related to the frequency of the graph, and propose to use graph Laplacian to measure and prune the inter-class edges. Furthermore, they propose Graph Heterophily Resistant Network (GHRN), which is equipped with a robust label-aware high-frequency indicator to measure the 1-hop label change of the center node. Qiao \etal~\cite{qiao2024truncated} propose local node affinity, an unsupervised anomaly scoring measure defined as similarity on node representations, which assigns a larger anomaly score to nodes that are less affiliated with their neighbors. They introduce Truncated Affinity Maximization (TAM) that learns tailored node representations by maximizing the local affinity of nodes to their neighbors on truncated graphs, where heterophilic edges are removed iteratively to mitigate the bias caused by them.

\paragraph{GAD-Specific Information Propagation}
Gao \etal~\cite{gao2023alleviating} notice that the structural distribution shifts vary between anomalies and normal nodes, and resisting high heterophily for anomalies and benefiting the normals from homophily can help address this issue. They develop Graph Decomposition Network (GDN), which teases out the anomaly features on which they constrain heterophilic neighbors and make them invariant, constrain the remaining features for normal nodes to preserve the connectivity of nodes, and reinforce the influence of the homophilic neighborhood.
Xiao \etal~\cite{xiao2024counterfactual} introduce an unsupervised counterfactual data augmentation method for GAD, which learns to identify potential anomalies and translates parts of their neighbors, which are probably normal, into anomalous ones. The translated neighbors are utilized in the aggregation step to produce counterfactual representations of anomalies, which are shown to be distinguishable and can improve detection performance.


\subsubsection{Bot Detection}
\label{sec:bot_detection}
Bots in network data are automated programs that mimic human behaviors and are often used for malicious purposes, such as spreading misinformation~\cite{cresci2020decade}, promoting
extremism~\cite{hamdi2022mining}, and electoral interference~\cite{deb2019perils}. Recent studies~\cite{des2022detecting} have shown that bots intentionally interact more with normal humans, which leads to increased influence of particular information and intentional evasion of detection. Such heterophilic relationship can be easily built with human beings by following normal accounts or replying to the tweets of normal users. Graph-based detection methods with feature aggregation mechanism tend to make neighboring node representations similar, and thus are not suitable for bot detection tasks. Some recent papers take heterophilic relation into their model design.

Li \etal~\cite{li2023multi} propose a multi-modal social bot detection method with learning homophilic and heterophilic connections adaptively (BothH). It consists of a classifier to determine homophilic or heterophilic edges, and an adaptive message propagation strategy for the homophilic and heterophilic connections. Ashmore \etal~\cite{ashmore2023hover} propose Homophilic Oversampling Via Edge Removal (HOVER) method to deal with bot detection, which removes the inter-class edges to reduce heterophily and enhance the node embeddings. Furthermore, they use the node embeddings to oversample the minority bots and generate a balanced class distribution. Wu \etal~\cite{wu2023heterophily} propose Bot Supervised Contrastive Learning (BotSCL), which is a heterophily-aware contrastive learning framework that can adaptively differentiate neighbor representations of heterophilic relations while assimilating the representations of homophilic neighbors. Ye \etal~\cite{ye2023hofa} put forward Homophily-Oriented augmentation and Frequency adaptive Attention (HOFA), which consists of a homophily-oriented graph augmentation module to improve the homophily of Twitter graph by adding edges between similar users, and a frequency adaptive attention module to adjust the attention weights of the edges based on their frequencies, so that homophilic edges are enhanced and heterophilic edges are suppressed. To cope with multi-modal graph-based rumour detection, Nguyen \etal~\cite{nguyenaportable} propose Portable Heterophilic Graph Aggregation for Rumour detection On Social media (PHAROS), which incorporates portable modality-aware aggregation (PMA) to discriminate node features across labels and modalities. 

\subsection{Generative Models}
\label{sec:generative_models}
Graph generative tasks aim to synthesize novel and realistic graphs from latent representations or conditional inputs. Graph heterophily can be exploited for various graph generative tasks, such as molecule generation and scene graph generation. These tasks have significant applications in drug discovery, computer vision, \etc{}.

\subsubsection{Molecule Generation}
\label{sec:molecule_generation}
The concept of heterophily can be applied to enhance generative models on graph-structured data. Wang \etal~\cite{wang2024molecule} propose a novel flow-based method for molecule generation with desirable properties, which is important for domains like material design and drug discovery. They argue that existing methods based on GNNs assume strong homophily, which overlooks the repulsions between dissimilar atoms and leads to information loss and over-smoothing. Thus, they introduce HTFlows, which uses multiple interactive flows to capture heterophily patterns in the molecular space and harnesses these (dis-)similarities in generation. HTFlows consists of three flows: a central flow, a heterophilic flow, and a heterophilic interaction flow. HTFlows is evaluated on several chemoinformatics benchmarks and outperforms existing methods in terms of validity, uniqueness, diversity, and property optimization.

\subsubsection{Scene Graph Generation}
\label{sec:scene_graph_generation}
Scene Graph Generation (SGG) aims to generate a structured representation of a scene by detecting objects and expressing their relationships through predicates in images~\cite{xu2017scene, li2024scene}. SGG plays a significant role in subsequent scene understanding tasks, \eg visual question answering~\cite{antol2015vqa, wu2017visual, lei2023symbolic} and image captioning~\cite{hossain2019comprehensive, yang2022reformer}.

GNNs are often used for SGG~\cite{deng2020generative, lin2022hl}, however, spatial GNNs do not work well in heterophilic SGG, where fine-grained predicates are always connected to a large number of coarse-grained predicates. Smoothing the graph signals hurts the discriminative information of the fine-grained predicates, resulting in decreased performance. 

The heterophily-specific techniques can be used to improve the quality of SGG. Lin \etal~\cite{lin2022hl} propose a novel Heterophily Learning Network (HL-Net) to comprehensively and efficiently explore the heterophily in objects and relationships in scene graphs. HL-Net consists of three modules: (1) an adaptive re-weighting transformer module, which adaptively integrates the information from different layers to exploit both the heterophily and homophily in objects; (2) a relationship feature propagation module that efficiently explores the connections between relationships by considering heterophily to refine the relationship representation; and (3) a heterophily-aware message-passing scheme to further distinguish the heterophily and homophily between objects and relationships, thereby facilitating improved message passing in graphs. Chen \etal~\cite{chen2024kumaraswamy} prove that the spectral energy gradually concentrates towards the high-frequency part when the heterophily on the scene graph increases. Based on this, they propose the Kumaraswamy Wavelet Graph Neural Network (KWGNN), which leverages complementary multi-group Kumaraswamy wavelets to cover all frequency bands, generating band-pass flters adaptively to better accommodate different levels of smoothness on the graph.

\subsection{Graph Learning Tasks}
\label{sec:graph_learning_tasks}
Besides node-level tasks, heterophily also poses significant challenges to other graph learning tasks, such as link prediction, graph classification, and graph clustering.

\subsubsection{Link Prediction}
\label{sec:link_prediction}
Link prediction is an important task that has wide applications in various domains~\cite{lei2013novel, nickel2015review, ceylan2021learning}. The majority of existing link prediction approaches assume the given graph follows the homophily assumption, and design similarity-based heuristics or representation learning approaches to predict links~\cite{newman2001clustering, jeh2002simrank, adamic2003friends}. However, many real-world graphs are heterophilic, which challenges the existing homophily-based link prediction methods.

Zhou \etal~\cite{zhou2022link} find that in heterophilic graphs, two connected nodes might be similar in some latent factors but have low overall feature similarity. Thus, they propose DisenLink, which can learn disentangled representations for each node, where each vector captures the latent representation of a node on one influential factor. DisenLink can model the link formation by comparing the disentangled representations of two nodes on different factors, and perform factor-aware message-passing to facilitate link prediction. Francesco \etal~\cite{Francesco2024LinkPU} introduce GRAFF-LP, an extension of the physics-inspired GNN, GRAFF~\cite{di2023understanding}, to link prediction, which can predict some latent connections on heterophilic graphs. Cho \etal~\cite{cho2024decoupled} find that the inner product based decoder in link prediction exhibits an intrinsic limitation, which hinders the gradient flow during training. Additionally, the message passing scheme is found unexpectedly dominated by the nodes with large norm values. They propose a stochastic Variational Graph Autoencoder (VGAE)-based method that can effectively decouple the norm and angle in the embeddings, and relates the cosine similarity and norm to homophily and node popularity principles, respectively.

Dot product graph embeddings construct vectors for nodes such that the products between two vectors indicate the strength of the connections. It is widely used in link prediction but is found to fail to model the non-transitive relationships~\cite{seshadhri2020impossibility, chanpuriya2020node}. Peysakhovich \etal~\cite{peysakhovich2021pseudo} remove the transitivity assumption by embedding nodes into a pseudo-Euclidean space, giving each node an attract (homophily) and a repel (heterophily) vector. Chanpuriya \etal~\cite{chanpuriya2021interpretable} make use of this attract-repel (AR) decomposition to build an interpretable model that is able to capture heterophily and produces non-negative embeddings to allow link predictions to be interpreted in terms of communities.

\subsubsection{Graph Classification}
\label{sec:graph_classification}
Beyond node-level tasks, the idea of heterophily can be employed to address graph-level tasks.
Ye \etal~\cite{ye2022incorporating} propose a novel architecture called IHGNN (Incorporating Heterophily into Graph Neural Networks) that can handle both homophily and heterophily in graphs. They identify two useful designs for IHGNN: (1) integration and separation of the ego- and neighbor-embeddings of nodes, and (2) concatenation of all the node embeddings as the final graph-level readout function. IHGNN is as expressive as the 1-WL algorithm, and achieves state-of-the-art performance on various graph classification datasets. Ding \etal~\cite{ding2023self} calculate the homophily ratio of several biological graphs, and empirically find out that the high-frequency signal is helpful for GNNs on graph classification tasks. Similar empirical results can be found in the Appendix of~\cite{luan2022complete}.

\subsubsection{Graph Clustering}
\label{sec:graph_clustering}
Graph clustering aims to partition a network into homogeneous groups of nodes and heterophily can be used to improve the performance.

Pan \etal~\cite{pmlr-v202-pan23b} propose a novel graph clustering method with three key components: (1) two unsupervised graph construction strategies to extract homophilic and heterophilic information from any type of graph; (2) a mixed filter based on the new graphs extracts both low- and high-frequency information; and (3) Dual Graph Clustering Network (DGCN) to reduce the adverse coupling between node attribute and topological structure by mapping them into two different subspaces. Gu \etal~\cite{gu2023homophily} introduce Homophily-enhanced structure learning (HoLe), a clustering method that can refine the input graph structure by adding missing links and removing spurious connections, based on the degree of homophily. Wen \etal~\cite{wen2024homophily} propose Adaptive Hybrid Graph Filter for multi-view graph Clustering (AHGFC) that can handle both homophilic and heterophilic graphs. It uses a graph joint process and an aggregation matrix to make the low- and high-frequency signals more distinguishable. Then, an homophily-related adaptive hybrid graph filter is applied to learn the node embedding. Xie \etal~\cite{xie2024provable} introduce a method to separately build graphs with strong homophily and heterophily. They utilize low- and high-pass filters to extract holistic information from the constructed graphs, and incorporate a squeeze-and-excitation block to enhance essential attributes. Achten \etal~\cite{achten2024hencler} challenge the assumption of clustering in heterophilic graphs, \ie{} good clustering satisfies high intra-cluster and low inter-cluster connectivity. They propose Heterophilous Node Clustering (HeNCler) to address the above issue, which includes a newly defined weighted kernel singular value decomposition to create an asymmetric similarity graph.

\subsection{Recommender System}
\label{sec:recommender_system}
The heterophily problem also exists in graph-structured social recommendation. Jiang \etal~\cite{jiang2024challenging} propose a data-centric framework called Social Heterophily-alleviating Rewiring (SHaRe), which adds highly homophilic social relations and cuts heterophilic relations to enhance graph-based social recommendation models. SHaRe also integrates a contrastive loss function to minimize the distance between users and their homophilic friends, and maximize the distance between users and their heterophilic friends.

\subsection{Biomedical}
\label{sec:biomedical}

\subsubsection{Combination Therapy} 
\label{sec:combination_therapy}
Combination therapy, which can improve therapeutic efficacy and reduce side effects, plays an important role in the treatment of complex diseases. However, a large number of possible combinations among candidate compounds limits our ability to identify effective combinations. Chen \etal~\cite{chen2022drug} propose a new computational pipeline, called DCMGCN,
which integrates diverse drug-related information, to predict novel drug combinations. In particular, by quantifying the feature similarities between effective drug pairs, they found that the drug-drug network is heterophilic, which may limit the effectiveness of the graph convolutional network (GCN). Therefore, they leverage the combination of intermediate representations and high-similarity neighborhoods to boost GCN performance on the heterophily and sparse drug-drug network.

\subsubsection{Drug Repositioning}
\label{sec:drug_repositioning}
Drug repositioning is a rapidly growing strategy for drug discovery, as the time and cost needed are significantly less compared to developing new drugs with wet-lab experiments. Most current GNN-based computational methods ignore the heterophily of the constructed drug–disease network, resulting in inefficient predictions. Liu \etal~\cite{liu2024slgcn} develop a structure-enhanced line graph convolutional network (SLGCN) to learn comprehensive representations of drug–disease pairs to address heterophily. SLGCN incorporates structural importance information to identify the irrelevant neighboring nodes and reduce their negative impacts, together with a gated update function to adaptively control the integration of ego biological representation and aggregated structural features.

\subsubsection{Drug Discovery}
\label{sec:drug_discovery}
Another possible application is to discover and design new drugs based on molecular graphs. Molecular graphs are heterophilic, as atoms with different types or properties may form chemical bonds. For example, carbon and oxygen atoms have different electronegativities and valences, but they can form covalent bonds in organic molecules. By using graph neural networks that can incorporate heterophily, we can learn more informative and discriminative embeddings of molecular graphs and improve the performance of drug discovery tasks. For example, Ye \etal~\cite{ye2022incorporating} propose a graph neural network model that integrates and separates the ego- and neighbor-embeddings of nodes, and apply it to graph classification of molecules.

\subsection{Knowledge Distillation}
\label{sec:knowledge_distillation}
To accelerate the inference of GNNs, a promising direction is to distill the GNNs into message-passing-free student multi-layer perceptrons (MLPs)~\cite{zhang2021graph, wu2023extracting, wu2023quantifying, tian2023knowledge}. However, the MLP student cannot fully learn the structure knowledge due to the lack of structure inputs and inexpressive graph representation space, which cause inferior performance in the heterophily scenarios. To address this issue, Chen \etal~\cite{chen2022sa} design a Structure-Aware MLP (SA-MLP) student that encodes both features and structures without message-passing and introduce a structure-mixing knowledge distillation strategy to enhance the learning ability of MLPs for structure information. Yang \etal~\cite{yang2024vqgraph} propose to learn a new powerful graph representation space by directly labeling the local structures of node and introduce a variant of VQ-VAE~\cite{van2017neural} to learn a structure-aware tokenizer on graph data that can encode the local substructure of each node as a discrete code. The discrete codes make up a codebook as a new graph representation space that is able to identify different local graph structures of nodes with the corresponding code indices. Based on the learned codebook, they propose a new distillation target, soft code assignments, to transfer the structural knowledge of each node from GNN to MLP and the model shows performance improvement on heterophilic graphs over GNN.

\subsection{Adversarial Attack and Robustness}
\label{sec:adversarial_attack_robustness}
Researchers have found that small, unnoticeable perturbations of graph structure can catastrophically reduce the performance of SOTA GNNs~\cite{zhang2020gnnguard}, especially on heterophilic graphs~\cite{zhu2022does}. In Section~\ref{sec:graph_structure_augmentation} and ~\ref{sec:robust_filter_network_architecture}, we summarize different methods to improve the robustness of GNNs under heterophily. In Section~\ref{sec:graph_injection_attack}, we introduce graph injection attack and its relationship with homophily.

\subsubsection{Graph Structure Augmentation}
\label{sec:graph_structure_augmentation}
Zhang \etal~\cite{zhang2020gnnguard} develop GNNGUARD, which includes the neighbor importance estimation and the layer-wise graph memory modules, to learn how to best assign higher weights to edges connecting similar nodes while pruning edges between unrelated nodes. The revised edges allow robust neural message propagation in the underlying GNN and can defend against attacks on heterophilic graphs. To achieve good performance with label noise, Cheng \etal~\cite{cheng2024label} propose graph Reconstruction for homophily and noisy label Rectification by Label Propagation (R$^2$LP), which is an iterative algorithm including: (1) graph reconstruction for high homophily; (2) label propagation to rectify the noisy labels; (3) high-confidence label selection to clean label set for the next iteration. Qiu \etal~\cite{qiu2023refining} conduct a quantitative analysis on the robustness of GCNs on heterophilic graphs, identifying the structural out-of-distribution (OOD) problem as the key vulnerability. To address this problem, they develop Latent Homophilic Structures (LHS) that enhances GCN resilience by learning and refining latent structures through a self-expressive technique based on multi-node interactions and a dual-view contrastive learning method. It enables GCNs to aggregate information in a homophilic way on heterophilic graphs and effectively mitigate structural OOD threats. Xie \etal~\cite{xie2024robust} study the problem of robust graph structure learning on heterophilic graphs. They first learn a robust graph with a high-pass filter and then propose a novel self-supervised regularizer to further refine the graph structure.

\subsubsection{Robust Graph Filter and Network Architecture}
\label{sec:robust_filter_network_architecture}
From the theoretical and empirical analyses in~\cite{zhu2022does}, the authors show that for homophilic graph data, impactful structural attacks always lead to reduced homophily, while for heterophilic graph data, the change in the homophily level depends on the node degrees. To address this problem, they utilize separated ego- and neighbor-embeddings, a design principle which has been identified to be useful for heterophilic graph, which can also offer increased robustness to GNNs. Lei \etal~\cite{lei2022evennet} claim that ignoring odd-hop neighbors improves the robustness of GNNs, given the observation that odd-hop neighbors tend to have opposite labels in heterophilic graphs and introduce noise in homophilic graphs. Based on this, they propose EvenNet, which is a new, simple and robust spectral GNN that corresponds to an even-polynomial graph filter. It only aggregates information from even-hop neighbors of each node and can improve the robustness of GNNs against graph structural attacks across homophilic and heterophilic graphs. Zhu \etal~\cite{zhu2024universally} elucidate why attackers often link dissimilar node pairs based on neighbor features across both homophilic and heterophilic graphs. To counter this vulnerability, they introduce $d$ Neighbor Similarity Preserving Graph Neural Network (NSPGNN), a robust model that employs a dual-kNN graph approach to guide neighbor-similarity propagation, leveraging low-pass filters for smoothing features across positive kNN graphs and high-pass filters for distinguishing features across negative kNN graphs. In \etal~\cite{in2024self} inject edge homophily property in graph augmentation strategy to increase the model robustness under adversarial attack. 

\subsubsection{Graph Injection Attack (GIA)}
\label{sec:graph_injection_attack}
Recently, Graph Injection Attack (GIA) emerges as a practical attack scenario on GNNs, where the adversary can only inject several malicious nodes instead of modifying existing nodes or edges, \ie{} Graph Modification Attack (GMA). Chen \etal~\cite{chen2021understanding} find that GIA can be provably more harmful than GMA due to its high flexibility, which leads to great damage to the homophily distribution of the original graph. Therefore, the threats of GIA can be easily alleviated or prevented by homophily-based defenses designed to recover the original homophily distribution. Then, they introduce homophily unnoticeability constraint that enforces GIA to preserve the original homophily, and propose Harmonious Adversarial Objective (HAO) to instantiate it. GIA with HAO can empirically break homophily-based defenses and outperform existing GIA attacks by a significant margin, which can serve for a reliable evaluation of the robustness of GNNs.

\subsection{Privacy}
\label{sec:privacy}
Real-world networks often exhibit high homophily~\cite{mcpherson2001birds} where connected nodes tend to share similar features. This characteristic can be exploited to infer the node information from its neighbors, raising significant concerns for information leakage and privacy preservation challenges~\cite{wu2024provable}.

Hu \etal~\cite{hu2022learning} highlight an instance of privacy risk in social networks, where some users (\ie{} private users) may prefer not to reveal sensitive information that others (\ie{} non-private users) would not mind disclosing. For example, male users are typically less sensitive to their age information than female users, therefore willing to give out their age information on social media. The disclosure potentially exposes the age information of the connected female users in the network, and the homophily property and message-passing mechanism of GNNs can exacerbate individual privacy leakage. This privacy issue can extend beyond highly homophilic graphs, as GNNs are still potentially vulnerable to privacy attacks in graphs with lower levels of homophily~\cite{zhang2023survey}. To address this issue, Hu \etal~\cite{hu2022learning} propose DP-GCN, a novel privacy-preserving GNN model which consists of two modules: (1) Disentangled Representation Learning Module, which disentangles the original non-sensitive attributes into sensitive and non-sensitive latent representations that are orthogonal to each other; (2) Node Classification Module, which trains the GCN to classify unlabeled nodes in the graph with non-sensitive latent representations. 

M{\"u}ller \etal~\cite{muller2023privacy} examine how the graph structure affects the performance of GNNs designed with differential privacy, highlighting that homophily plays a critical role in model utility: for the graph structure with low homophily, differential privacy has a stronger negative impact on model performance compared to graph with high homophily. Yuan \etal~\cite{yuan2024unveiling} investigate Graph Privacy Leakage via Structure (GPS) and introduce the Generalized Homophily Ratio (GHRatio) to quantify the privacy breach risks in GPS. Based on this, they develop a graph private attribute inference attack with various homophily to evaluate the potential for privacy leakage through network structures and propose a graph data publishing method with a learnable graph sampling method for privacy protection. Wang \etal~\cite{wang2024gcl} study the privacy vulnerability of Graph Contrastive Learning (GCL) and  find that under their proposed attack GCL-Leak, higher homophily (modularity) generally results in higher attack accuracy and different GCL models are affected by homophily to different degrees. They explain that, on the graphs with higher homophily, the connected nodes are more likely to exhibit similarities, which leads to less sensitive to edge augmentation compared to less homophilic graphs.

\subsection{Urban Computing}
\label{sec:urban_computing}
Urban dynamics, such as traffic flow, air quality, and human mobility, can be modeled by networks. Urban networks are often heterophilic, as nodes with different attributes or functions may have strong interactions. For instance, roads with different traffic conditions, land uses, or geographic locations can influence each other. Urban computing involves designing and optimizing the structures and functions of urban networks and can be better learned by GNNs with heterophily.

Xiao \etal~\cite{xiao2023spatial} address the problem of applying GNNs to urban graphs, where the nodes are urban objects (such as regions or points of interest) and the edges represent spatial proximity or similarity. They argue that urban graphs often exhibit a unique spatial heterophily property, which means that the dissimilarity of neighbors at different spatial distances can vary greatly. Firstly, Xiao \etal~\cite{xiao2023spatial} propose a metric, called Spatial Diversity Score (SDS), to quantify the spatial heterophily of a graph, and show how it affects the performance of GNNs. They claim that existing GNNs, especially those designed for heterophilic graphs, are not effective in handling urban graphs with high spatial diversity score. Subsequently, they introduce a novel GNN architecture, called Spatial Heterophily-aware Graph Neural Network (SHGNN), which can capture the spatial diversity of heterophily in urban graphs. SHGNN has two key modules: a rotation-scaling spatial aggregation module, which groups the spatially close neighbors and processes each group separately; and a heterophily-sensitive spatial interaction module, which adapts to the commonality and diverse dissimilarity in different spatial groups. They also provide some insights about the learned spatial patterns and heterophily types by the SHGNN.

\subsection{Graph Coloring}
\label{sec:graph_coloring}
The principle of heterophily can be leveraged in GNNs to tackle graph coloring challenge, which addresses the problem of assigning different colors to the nodes of a graph such that no adjacent nodes have the same color. Wang \etal~\cite{wang2024graph} recognize the graph coloring task as a heterophilic problem and introduce negative message passing into a physics-inspired graph neural network, which allows nodes to exchange information with their neighbors in a negative way, such that the node features can become more dissimilar after each message passing layer. They also propose a new loss function that incorporates information entropy of nodes, which measures the uncertainty of the node labels. The loss function increases the uniformity of the color assignment of each node, encouraging the nodes to have high self-information and low mutual information with their neighbors, which leads to better graph coloring results.

\subsection{Binary Programming}
\label{sec:binary_programming}
Integer programming is a mathematical optimization problem, where the decision variables are restricted to be integers~\cite{papadimitriou1998combinatorial}. In particular, Binary Programming (BP) is a special case in which the unknown variables are restricted to take binary values (either 0 or 1) and is known to be NP-complete~\cite{karp2010reducibility}. BP has numerous real-world application scenarios, \eg{} portfolio optimization~\cite{black1992global}, manufacturing and supply chain optimization~\cite{geunes2005supply}, telecommunications network optimization~\cite{resende2008handbook} and resource allocation~\cite{brown1984concept}, \etc{} 

Solving BP problems efficiently is crucial and challenging, and recently Eliasof \etal~\cite{eliasof2024graph} formulate BP as a graph learning task where they leverage GNNs to approximate the solutions. The sensitivity analysis of BP \wrt the input variables sheds light on patterns and behavior of the solutions, providing a way for them to treat the solution of BP as heterophilic node classification. Therefore, they propose Binary Programming GNN (BPGNN), an architecture that integrates graph representation learning techniques with BP-aware features to approximate BP solutions efficiently. Moreover, to address the scarcity of labeled data problem, they introduce a self-supervised data generation mechanism, which enables efficient and tractable training data acquisition for the large-scale NP-complete problem.

\subsection{Remote Sensing}
\label{sec:remote_sensing}

Nowadays, our planet is monitored by a wide variety of sensors that provide different information about the surface of Earth~\cite{campbell2011introduction}. A multi-modal set of information can be obtained and combined from different sensors to analyze complex areas on Earth with high precision~\cite{debes2014hyperspectral}. An important data processing step is to provide labels for these multi-modal datasets and label propagation (LP) method is widely used when we convert the data to a graph \cite{zoidi2015graph}. However, standard LP algorithms cannot be applied to multi-modal remote sensing data directly, because (1) classic LP algorithms only applies to homogeneous graphs while multi-modal data is heterogeneous; (2) LP requires homophily assumption and it will fail on real-world data with heterophily.

To address the above issue, Taelman \etal~\cite{taelman2022exploitation} design a new label propagation method for multi-modal remote sensing data that works on a heterogeneous graph with homophily and/or heterophily interactions through the propagation matrices. The propagation matrices can automatically be estimated from any homogeneous or heterogeneous sparsely-labeled
graph without expert knowledge. In addition, the proposed method can propagate information between uni-modal and multi-modal data points.

\subsection{Graph Positive-Unlabeled Learning}
\label{sec:positive_unlabeled_learning} 
Positive-Unlabeled (PU) learning has been extensively studied on image~\cite{kiryo2017positive} and text~\cite{li2016classifying} data, where a binary classifier is trained on positive-labeled and unlabeled samples. The scenarios for PU learning also widely exist in graph-structured data, \eg{} in online transaction networks, fraudsters are only marked as positive nodes when they are detected and the labels of other users are unknown~\cite{yoo2021accurate}; in pandemic prediction networks, only the infected nodes are identified as positive, while the remaining nodes are label-agnostic~\cite{panagopoulos2021transfer}. 

Existing PU learning methods cannot be directly extended to graph data, because the nodes and edges are not assumed to be independent, which breaks the hypothesis in traditional PU learning that each sample is independently generated. Especially, Wu \etal ~\cite{wu2024unraveling} reveals that the heterophily structure impedes two aspects of PU learning when adapting to graph data: (1) the latent feature entanglement of positive and negative nodes challenges Class-Prior Estimation (CPE), \ie{} the estimation of the fraction of positive nodes among the unlabeled nodes, and violates the irreducibility assumption, causing an overestimated class prior; (2) the heterophilic structures hinder the accurate latent label inference on unlabeled nodes during training.

To address the above issues, Wu \etal~\cite{wu2024unraveling} propose Graph PU Learning with Label
Propagation Loss (GPL), which learns from PU nodes along with a proposed Label Propagation Loss (LPL). The optimization of LPL results in increased weights on homophilic edges and reduced weights on heterophilic edges, which promotes the accuracy of CPE and PU classification. LPL can interact with the vanilla graph PU learning iteratively, which can be formulated as an efficient bilevel optimization process, where the heterophily level is reduced in the inner loop and the classifier is learned in the outer loop.

\subsection{Point Cloud Segmentation}
\label{sec:point_cloud_segmentation}
Point cloud is a widely used 3D data format for 3D computer vision~\cite{guo2020deep, liao2022kitti} and point cloud segmentation is an important task for 3D understanding~\cite{qi2017pointnet, qi2017pointnet++}, which aims to divide and classify the target point cloud into separate regions with different attributes or functions. The three most popular methods, \ie{} point-based method~\cite{qi2017pointnet}, graph-based method~\cite{wang2019graph}, and transformer-based method~\cite{yu2022point},
implicitly assume that connected nodes are likely to have similar attributes and ignore the heterophily nature of some edges, especially at the boundary regions. Therefore, the blending of information from heterophilic neighbors undermines the distinguishability of the node representation~\cite{du2024graph}.

To address the above problem, Du \etal~\cite{du2024graph} propose graph regulation network (GRN), which models the point cloud as a homophilic-heterophilic relation graph to produce finer segmentation boundaries. GRN predicts homophily and heterophily probabilities of the edges, adaptively adjusts the propagation mechanism according to different neighborhood homophily, and has an additional prototype feature extraction module to capture the homophily features of nodes from the global prototype space.

\subsection{Federated Learning on Graphs}
\label{sec:federated_learning}

Federated Learning (FL) is a distributed learning paradigm in which powerful global models are trained by multiple local clients collaboratively without sharing their local data, in consideration of data privacy, commercial competition, resource limitation, and other regulations~\cite{hard2018federated, silva2019federated}. One of the biggest challenges in FL over graph data is the non-IID data across clients, \ie{} heterogeneity regarding link types. More specifically, the links captured in real-world networks which appear to be uniformly distributed can carry different levels of homophily, and the sets of latent link types and subgraph topology can differ across local clients~\cite{xie2023federated}.

To tackle the above problem, Xie \etal~\cite{xie2023federated} design FedLit, a new graph FL framework that can discover latent link-types and model message-passing \wrt the discovered link-types simultaneously. Specifically, FedLit dynamically detect the latent link types during FL via an EM-based clustering algorithm and differentiate the message-passing through different types of links via multiple convolution channels. Zhu \etal~\cite{zhu2024fedtad} reveal that variations in label distribution and structure homophily in local subgraphs lead to significant differences in the class-wise knowledge reliability of multiple local GNNs, misguiding model aggregation. To handle it, they propose topology-aware data-free knowledge distillation technology (FedTAD), improving the reliability of knowledge transfer from the local model to the global model. Li \etal~\cite{li2024adafgl} introduce the structure \niid split and present Adaptive Federated Graph Learning (AdaFGL) paradigm, a decoupled two-step personalized approach. It first uses standard multi-client federated collaborative training to acquire a federated knowledge extractor through aggregation of uploaded models in the final round at the server. Then, each client conducts personalized training based on their local subgraph and the federated knowledge extractor.

The concept of homophily is also found to be closely related to gossip learning~\cite{ghosh2023modeling}

\subsection{Software Defect Localization}
\label{sec:software_detect_localization}

Traditional defect localization identifies defective files, methods, or code lines based on symptoms like defect reports, but often after the defect has already damaged the software. In contrast, Just-In-Time (JIT) defect localization predicts the problematic code lines at the time a defective change is first submitted. JIT enables the detection of issues before they really cause harm, providing timely support for code review.

Zhang \etal~\cite{zhang2024just} construct line-level code graphs for code changes and test whether the homophily assumption holds on the constructed code graphs with nodes labeled as "defective" or "non-defective". Results reveal a non-uniform distribution with most code graphs showing strong homophily but some showing significant heterophily. They adopt FAGCN~\cite{bo2021beyond} to address the uneven homophily challenge in learning from code graph tasks.

\subsection{Brain Network Analysis}
\label{sec:brain_network_analysis}
Analysis of neurodegenerative diseases through brain connectomes is crucial for early diagnosis and onset prediction~\cite{wang2017multi, zhu2018dynamic}. In a neurodegenerative brain connectome study, each sample is represented as a graph, \ie{} brain regions of interests (ROIs) correspond to nodes and connectomic features represent edges. A subset of brain networks progressively deteriorates over time due to a disease~\cite{wang2023hypergraph, qu2023graph}. Also, just like many real-world graphs, brain networks exhibit complex structures with both homophily and heterophily, where dissimilar ROIs can physically attach. Such interplay between homophily and heterophily makes the analysis more difficult~\cite{choneurodegenerative}.

To tackle these challenges, Cho \etal~\cite{choneurodegenerative} propose Adaptive
Graph diffusion network with Temporal regularization (AGT). AGT utilizes node-wise convolution to adaptively capture low  and high-frequency signals within an optimally tailored range for each node. Moreover, AGT captures sequential variations within progressive diagnostic groups with a novel temporal regularization, considering the relative feature distance between the groups in the latent space, and thus yields interpretable results at both node-level and group-level.

\subsection{Graph Pre-training}
\label{sec:prompt_pretraining}
The idea of graph heterophily can be leveraged for graph pre-training and fine-tuning.
Gong \etal~\cite{gong2023prompt} propose prompting graph contrastive learning (PGCL) to leverage graph heterophily for graph pre-training and fine-tuning. PGCL uses a multi-view graph contrastive learning approach as the pre-training task, which creates different views of the same graph and encourages the model to learn from the similarities and differences between them. Ge \etal~\cite{ge2023enhancing} propose SAP to adapt pre-trained GNNs to various tasks with less supervised data. It exploits the structure information of graphs in both pre-training and prompt tuning stages. SAP aligns the semantic spaces of node attributes and graph structure using contrastive learning, and incorporates structure information in prompted graph to elicit more pre-trained knowledge. SAP can improve the performance of GNNs on node classification and graph classification tasks, especially in few-shot scenarios and heterophilic graphs. 

\section{Challenges and Future Directions}
\label{sec:challeges_future_directions}

Graph heterophily is an active research topic, as there are still many open questions and challenges to be addressed numerous real-world scenarios. In this section, we explore how graph heterophily can be leveraged or addressed in various domains, such as
\begin{itemize}
    \item Section~\ref{sec:sptial.temporal.graph}: temporal graphs
    \item Section~\ref{sec:hypergraphs}: hypergraphs
    \item Section~\ref{sec:fairness}: AI fairness
    \item Section~\ref{sec:drug.discovery}: drug discovery
    \item Section~\ref{sec:climate.change}: climate change
    \item Section~\ref{sec:single.cell}: single cell genomics
    \item Section~\ref{sec:computer.vision}: computer vision
    \item Section~\ref{sec:llm}: large language models
    \item Section~\ref{sec:gfm.scaling.law}: graph foundation models and neural scaling law
    \item Section~\ref{sec:graph.structure.learning}: graph structure learning and augmentation
    \item Section~\ref{sec:finance.net}: financial networks
    \item Section~\ref{sec:scalability}: scalability
\end{itemize}


\subsection{Heterophily on Temporal Graphs}
\label{sec:sptial.temporal.graph}

Heterophily GNNs are mostly investigated on classification tasks over static graphs while under-explored on node-level regressions over dynamic graphs. Zhou \etal~\cite{zhou2022greto} revisit node-wise relationships and explore novel homophily measurements on dynamic graphs with both signs and distances, capturing multiple node-level spatial relations and temporal evolutions. They discover that advancing homophily aggregations to signed target-oriented message passing can effectively resolve the discordance and promote aggregation capacity. Therefore, GReTo is proposed, which performs signed message passing in immediate neighborhood, and exploits both local environments and target awareness to realize high-order message propagation. 

In the future, more experimental evidence is needed to identify the relation between the performance of temporal GNNs and the time-varying graph homophily values; heterophilic temporal graph benchmark datasets are required to evaluate the model performance; new performance metrics and theoretical analysis should be established to further understand the heterophily issue for temporal GNNs.

\subsection{Heterophily on Hypergraphs}
\label{sec:hypergraphs}
Compared with simple graph, on which each edge only connects two nodes, a hypergraph can encode data with high-order relations (beyond pairwise relations) using its degree-free hyperedges, which can connect multiple nodes~\cite{feng2019hypergraph}. Heterophily has been proved as a more common phenomenon in hypergraphs than in simple graphs since nodes in a giant hyperedge are more likely to have different labels and irregular patterns~\cite{veldt2023combinatorial}.

Wang \etal~\cite{wang2022equivariant} propose a new Hypergraph neural networks (HNNs) architecture, named ED-HNN, which provably approximates any continuous equivariant hypergraph diffusion operators that can model a wide range of high-order relations. ED-HNN can be implemented efficiently by combining star expansions of hypergraphs with standard message passing neural networks and shows great superiority in processing heterophilic hypergraphs. Zou \etal~\cite{zou2024unig} propose UniG-Encoder, a universal feature encoder for the representation of both graph and hypergraph, which transforms the topological relationships of connected nodes into edge or hyperedge features via a normalized projection matrix. UniG-Encoder can comprehensively exploit the node features and graph/hypergraph topologies in an efficient and unified manner to address heterophily problem. Nguyen \etal~\cite{nguyen2024sheaf} propose Sheaf HyperNetworks (SHNs), which combines cellular sheaf theory with hypergraph networks to mitigate  heterophily problem.

For future research, there will be a need for new homophily metrics, benchmark datasets and models for heterophilic hypergraph representation learning.

\subsection{Homophily and Fairness}
\label{sec:fairness}
A key challenge faced by homophily is the fairness issue. Loveland \etal~\cite{loveland2022graph} reveals a link between group fairness and local assortativity, often induced by homophily, particularly evident in socially-induced clusters sharing a sensitive attribute. It is discovered that not all node neighborhoods are equal, and those dominated by a single category of a sensitive attribute often face challenges in achieving fair treatment, particularly when local class and sensitive attribute homophily diverge. Loveland \etal~\cite{loveland2024performance} theoretically show that the network structure could lead to unfairness in human-centric settings. Cao~\cite{cao2024recommendation} study the evolution of recommendation fairness over time and its relation to dynamic network properties, and show that extreme values for homophily or heterophily can be detrimental to recommendation fairness in the long run, even when group sizes are balanced in the data. They also demonstrate  that promoting homophily in heterophilic networks and heterophily in homophilic networks can improve the fairness of recommendations. Gholinejad \etal~\cite{gholinejad2024heterophily} propose HetroFair, a fair GNN-based recommender system, which includes fairness-aware attention and heterophily feature weighting to generate fairness-aware embeddings.

Besides, Luo \etal~\cite{luo2024your} introduce a Node Injection-based Fairness Attack (NIFA) designed with uncertainty-maximization principle and homophily-increase principle. NIFA is verified to be an effective fairness attack, which can significantly undermine the fairness of mainstream GNNs, including the fairness-aware GNNs, by only injecting 1\% of nodes.

Given that the principle of homophily serves as a default assumption in almost all GNNs, there is a high likelihood that the model we trained may encounter fairness issues. Consequently, future exploration is necessary to tackle this ethical concern.

\subsection{Drug Discovery}
\label{sec:drug.discovery}
In drug discovery, molecules can be modeled as graphs with various type, by setting atoms as nodes and bonds as edges. 
However, the homophily levels of different molecules can vary across different domains, datasets, and tasks, requiring adaptive and robust graph learning methods to handle different levels and types of heterophily. This poses new challenges for representation learning of molecular graphs. Therefore, we propose the following directions for future work on graph heterophily for drug discovery: (1) develop new metrics to quantify and visualize heterophily in molecular graphs, and to compare and evaluate the performance of graph models on graphs with different homophily values; (2) explore new ways to generate and augment graph views that enhance heterophily signals, such as adding or removing edges, perturbing node labels or features, or applying graph transformations or operations; (3) design new architectures and mechanisms to aggregate and integrate heterophilic node embeddings, such as attention, gating, or fusion modules, \etc; and (4) design new graph-level readout functions that preserve and exploit heterophily information, such as concatenation, pooling, graph convolution based on graph similarity or contrastive learning objectives.

\subsection{Climate Change}
\label{sec:climate.change}
Climate change is a global challenge that involves various factors, such as greenhouse gas emissions, land use, soil quality and human activities. These factors can be represented by graphs, with nodes as entities (\eg countries, regions, sectors, species) and edges as relationships (\eg emissions, trade, migration, predation)~\cite{bayraktar2023graph}. Graph heterophily can help capture the diversity of these entities, and reveal new patterns that may not be apparent from homophilic graphs. It can also enable new ways to analyze and visualize the complex interactions between the ecosystem and human activities \cite{bayraktar2023graph}.

For example, graph heterophily can help identify entities that have different features but similar impacts on climate change, or vice versa. This can lead to new strategies for mitigation and adaptation, such as finding alternative sources of energy, reducing consumption, or enhancing resilience. Graph heterophily can also help visualize the spatial and temporal dynamics of climate change, such as how different regions or seasons experience different levels or types of heterophily. This can enhance the communication and understanding of climate change, and foster collaboration and action among different stakeholders.

Graph learning methods have several challenges when dealing with heterophily in climate data, such as defining and measuring heterophily, generating and augmenting heterophilic graph views, aggregating and integrating heterophilic node embeddings, and designing graph-level readout functions to preserve heterophily information.


\subsection{Single-Cell Genomics}
\label{sec:single.cell}
Single cell data refers to the gene expression profiles of individual cells, which can reveal the heterogeneity and diversity of cell types and states~\cite{stuart2019comprehensive}. One potential use of graph heterophily in single cell data is to identify novel or rare cell types that are not well represented by the existing cell type annotations. By constructing a graph based on the similarity or distance of single cell gene expression and applying graph models, one can discover clusters or subgraphs of cells that exhibit high heterophily, indicating that they are distinct from their neighbors. This can help uncover new biological insights or functions of these cells~\cite{wang2021single}.

One challenge of graph heterophily in single cell data is to deal with the noise and uncertainty in data, which can affect the quality and reliability of the graph construction and analysis. Single cell data often suffers from low coverage, dropout events, batch effects, and other sources of variability, which can introduce errors or biases in the gene expression measurements~\cite{stuart2019comprehensive, lahnemann2020eleven}. These can in turn affect the graph structure and properties, such as the degree distribution, the clustering coefficient, or the heterophily ratio. Therefore, robust and scalable methods are needed to handle the noise and uncertainty of single cell data and graph heterophily~\cite{wang2021single, xiao2023simple}.

Another challenge of graph heterophily in single cell data is interpretability. Graph-based methods can provide powerful and flexible ways to model and learn from single cell data, but they can also be complex and opaque, making it hard to understand how they work and why they produce certain outcomes. For example, how does a GCN capture the heterophily of a graph? What are the key features or genes that contribute to the heterophily? How can we validate or verify the results of the graph-based methods? These are some of the questions that need to be addressed to make graph heterophily in single cell data more explainable.

\subsection{Computer Vision}
\label{sec:computer.vision}
In computer vision, graphs are often used to model complex relations between image regions or objects. Heterophily can occur when image regions have distinct visual characteristics, such as colors, textures, or shapes. GNNs with heterophily can better capture these diverse features, thereby improving model performance.

A possible application of graph heterophily is to learn and analyze 3D geometric shapes which can be represented as graphs, where nodes are points or vertices, and edges are connections or faces. 3D geometric graphs can be heterophilic, as nodes with different geometric features or semantic labels may be adjacent. For example, a chair may have nodes with different shapes, colors, or functions, such as legs, seat, or backrest. By using GNNs for heterophily, we can learn more informative and discriminative embeddings for 3D geometric graphs and improve the model performance. Bronstein \etal~\cite{bronstein2021geometric} provide a comprehensive survey on geometric deep learning methods for 3D shapes.

In the context of semantic segmentation, where each pixel in an image is assigned a semantic label (\eg sky, tree, car), graphs can represent pixel neighborhoods~\cite{qi20173d, lu2019graph, xie2021scale}. Heterophily may occur due to varying textures, lighting conditions, or object types and heterophily-aware GNNs can enhance segmentation performance by considering context from diverse neighboring pixels.

For object recognition, graphs can model relationships between objects (\eg bounding boxes) in an image~\cite{qasim2019rethinking, shi2020point}. Heterophily may arise when objects have different shapes, sizes, or appearances. GNNs that adapt to heterophily can improve object recognition and localization by capturing context across diverse object instances.

In scene understanding, scene graphs represent objects and their relationships within a scene. Heterophily can arise due to variations in object categories, attributes, or spatial arrangements. GNNs designed to handle heterophily can enhance scene understanding by considering both local and global context. For example, Lin \etal~\cite{lin2022hl} propose a heterophily learning graph network to generate scene graphs with rich heterophilic information.

In graph-based image retrieval, graphs can encode visual similarities between images~\cite{zhang2020understanding, misraa2020multi, kan2022local, yu2022text}. Heterophily arises when images depict different scenes, objects, or visual styles. GNNs that account for heterophily can improve image retrieval by capturing diverse visual cues.

In cross-modal graphs, where the edge connects images, text, and other modalities, heterophily can occur~\cite{yu2018modeling, misraa2020multi, yu2022text}. For tasks like image captioning or visual question answering, considering diverse modalities is crucial. Heterophily-aware GNNs can enhance cross-modal reasoning and fusion.

In summary, understanding and addressing graph heterophily in GNNs can lead to more robust and accurate computer vision models, especially when dealing with diverse visual objects.

\subsection{Large Language Models}
\label{sec:llm}
Large Language Models (LLMs), such as GPT~\cite{achiam2023gpt} and Llama~\cite{touvron2023llama}, are making significant progress in natural language processing, due to their strong text encoding/decoding ability and emergent capability, \eg{} reasoning~\cite{achiam2023gpt, guo2023gpt4graph, jin2023large}. However, LLMs are mainly designed to process pure texts, while in many real-world applications, the data contains information in the form of graphs. For example, academic networks and e-commerce networks involve text data that is associated with rich graph structure information, and molecules with descriptions include graph data that is paired with rich textual information. LLMs may struggle to capture such complex structural and semantic information in graphs, especially when the graph exhibits high heterophily~\cite{jin2023large}. And on the other hand, graph heterophily can potentially be used in LLMs to handle some complex tasks and perform effective learning and reasoning on graphs. We categorize their relationship into the following taxonomies.

\paragraph{LLM for Graph} LLMs can help to enhance the textual information on graphs, such as generating node or edge attributes, or summarizing graph contents. This can help enrich the graph data with more details or descriptions, or provide a concise overview of the graph data. For example, LLMs can be used to generate product reviews or ratings on e-commerce networks, or summarize the main topics or trends in academic networks~\cite{guo2023gpt4graph, jin2023large}. In particular, Qiao \etal~\cite{qiao2024login} propose a new paradigm called LLMs-as-Consultants to allow the interaction between LLMs and GNNs during training process, where LLMs assist in refining GNN operations either by enhancing node features based on correct LLM predictions, or by adjusting the graph structure (like pruning edges) when LLM predictions are incorrect. This interactive consultation helps GNNs handle heterophilic structures better by adapting to the specific needs identified during the training process.

LLMs can also help to generate and complete graph structures to discover new patterns from graph data or generate new graph structures for various purposes. For example, LLMs can be used to generate new molecules with desired properties, or complete missing citations in academic networks~\cite{guo2023gpt4graph, jin2023large}.

\paragraph{Graph for LLM} Graph information can also be used to help LLMs solve complex reasoning tasks, through prompting via a specific graph structure~\cite{besta2024graph, sun2023think, yu2023thought}. These works treat the reasoning procedure as navigating, searching and retrieving from a reasoning graph for complex tasks, where nodes are sub-problems, analogical problems or intermediate states for a given task. Such graphs often exhibit the characteristics that connected problems or states have significant differences and specific relationships.

All the above directions face challenges when dealing with heterophilic data. For example, LLMs may lose the discriminative power of node representations when aggregating information from heterophilic neighbors, leading to poor performance on downstream tasks~\cite{zhu2023heterophily}. In addition, LLMs can amplify existing biases or imbalances in graph data, such as favoring homophilic or high-degree nodes, resulting in unfair or inaccurate algorithmic outcomes~\cite{zhu2023heterophily}.

To address these challenges, various techniques have been proposed to improve the design and training of LLMs on graphs, such as incorporating graph attention mechanisms, graph regularization methods and graph adversarial training~\cite{zhu2023heterophily, guo2023gpt4graph, jin2023large}. However, there is still much room for further research and innovation in the fast-growing field of graph heterophily and LLMs.

\subsection{Graph Foundation Models and Neural Scaling Law}
\label{sec:gfm.scaling.law}
Graph pretraining is a new research topic in graph learning domain, aiming to develop a graph foundation model (GFM) which is capable of generalizing across different kinds of tasks on graphs with various properties~\cite{liu2023towards, zhao2024all, liu2024one}. GFMs are inspired by the existing foundation models in computer vision and natural language processing domains, which are pre-trained on massive data and can be adapted to tackle a wide range of downstream tasks. GFMs try to leverage the prior knowledge obtained from the pre-training stage and the small amount of data from downstream tasks to achieve better performance and even deliver promising efficacy for zero-shot task.

However, a versatile GFM has not yet been achieved~\cite{mao2024graph}. The key challenge in building GFM is how to enable positive transfer between graphs with diverse structural patterns~\cite{mao2024graph}. One possible solution is to construct a 'graph vocabulary', in which the basic transferable units underlying the graphs encode the invariance of the graphs. Think of it as a set of fundamental building blocks that allow GFMs to generalize effectively~\cite{liu2023towards, ye2023natural, mao2024graph}. \textbf{Finding the invariant and transferable blocks between homophilic and heterophilic graphs is one of the most important steps towards the success of GFMs on node-level tasks.} To construct this graph vocabulary, we point out insights in several essential aspects: (1) understanding the underlying network properties and patterns; (2) designing or learning versatile and universal node tokenization for different graph learning tasks; and (3) ensuring robustness and stability across diverse graphs. In addition, such a vocabulary perspective can potentially advance the future GFM design following neural scaling laws~\cite{liu2024neural}, allowing these models to scale efficiently and adapt to various types of downstream tasks~\cite{liu2023towards, ye2023natural, mao2024graph,zhao2024graphany}. 

Heterophily is also found to be related to data pruning technique, which has been shown to beat neural law scaling by only reserving data with an appropriate fraction of the hardest examples~\cite{sorscher2022beyond}. It can help remove the less valuable training nodes, so that the computational overhead is reduced and the model training process can be accelerated. It is observed that the successful pruning methods prefer hard training examples, \ie{} heterophilic nodes, and the average homophily level of the training nodes will significantly decrease after effective pruning~\cite{wang2024exploring}. This indicates that the heterophilic nodes are potentially more informative than homophilic nodes for GFM.

Although graph heterophily poses unique challenge to the design of GFMs, it can be potentially used to handle heterogeneity among graph-structured data for GFMs. Existing graph datasets cannot be uniformly utilized for pre-training due to the heterogeneity induced by missing features and different semantic spaces~\cite{mao2024graph}. In addition, one may use graph heterophily to guide the learning of node and edge embeddings that capture the semantic similarity and dissimilarity among different types of entities and relations. One can also use graph heterophily to enhance the cross-modal alignment and fusion of multi-modal graphs, such as text, image, video, and audio.

\subsection{Graph Structure Learning and Augmentation}
\label{sec:graph.structure.learning}
Graph (structure) augmentation aims to perturb the graph structure through heuristic or probabilistic rules~\cite{ding2022data,zhaograph,hua2022graph}, enabling the nodes to capture richer contextual information, and thus improving generalization performance. While there have been a few graph structure augmentation methods proposed recently, none of them are aware of a potential negative augmentation problem, which may be caused by overly severe distribution shifts between the original and augmented graphs. 

Wu~\cite{wu2022knowledge} \etal analyze the distribution shifts between the two graphs with graph homophily and measure the severity of an augmentation algorithm suffering from negative augmentation. To tackle this problem, they propose  Knowledge Distillation for Graph Augmentation (KDGA) framework, which helps to reduce the potential negative effects of distribution shifts, \ie{} negative augmentation problem. Specifically, KDGA extracts the knowledge of any GNN teacher model trained on the augmented graphs and injects it into a partially parameter-shared student model that is tested on the original graph.

Recently, Zhou \etal~\cite{zhou2024opengsl} empirically find that there is no significant correlation between the homophily of the learned graph structures and model performance on the tasks, challenging the common belief that higher homophily can lead to better performance. More studies are need to explore the relationship between graph structure learning methods and homophily.

\subsection{Financial Networks}
\label{sec:finance.net}
Graph heterophily can be useful for modeling financial networks, where entities such as banks, firms, or customers may have diverse attributes or behaviors. Graph heterophily can capture multiple aspects of financial networks. 

In the context of credit risk assessment, where the task is to evaluate the probability of default or loss by a borrower or lender~\cite{song2024enhancing}, graphs can depict the credit relationships among entities. These relationships may exhibit heterophily due to variations in credit ratings, loan amounts, or repayment histories. GNNs that account for heterophily can improve credit risk assessment by considering the influence of diverse neighbors on a node’s creditworthiness.

For portfolio management, where task involves selecting the optimal combination of assets to maximize returns and minimize risks, graphs can encode the similarities or correlations between assets, where heterophily arises when assets have different returns, volatilities, or market sectors~\cite{pacreau2021graph, soleymani2021deep}. GNNs that handle heterophily can improve portfolio optimization by capturing the trade-offs between diverse assets.

In fraud detection, where the task aims to identify anomalous or malicious activities in financial transactions, graphs can represent the transaction patterns between entities~\cite{shi2022h2, xu2023node, zhuo2024partitioning}. And heterophily may arise due to different transaction types, amounts, or frequencies. GNNs that adapt to heterophily can improve fraud detection by capturing the context from diverse transaction partners~\cite{shi2022h2, kim2023dynamic, zhuo2024partitioning}.

\subsection{Scalability}
\label{sec:scalability}
Most existing heterophilic GNNs incorporate iterative non-local computations to capture node relationships. However, these approaches have limited application to large-scale graphs due to their high computational costs and challenges in adopting mini-batch schemes. Liao \etal~\cite{liao2023ld2} studied the scalability issues of heterophilic GNN and propose a scalable model, LD2, which simplifies the learning process by decoupling graph propagation and generating expressive embeddings prior to training. Theoretical analysis demonstrates that LD2 achieves optimal time complexity in training, as well as a memory footprint that remains independent of the graph scale. Das \etal~\cite{das2024agsgnn} propose AGS-GNN, a scalable and inductive GNN with attribute-guided sampling. It employs samplers based on feature similarity and diversity, which is modeled by a sub-modular function, to select subsets of neighbors and adaptively captures homophilic and heterophilic neighborhood information with dual channels.

\section{Conclusion}
\label{sec:conclusion}
In this survey, we provide a thoroughly overview of the benchmarks, graph models, theoretical analysis and related applications for heterophilic graph representation learning, and put forward the unique challenges that arise from the special nature of heterophilic data. Notably, we categorize heterophilic datasets into benign, malignant and ambiguous groups, and highlight the latter two as the real challenging ones, which sets up a foundation for model evaluations in future studies. As the field grows, it is crucial to refine and explore better models to exploit the complex patterns of heterophilic graphs.

\section*{Acknowledgements}
Although not in the author list, we still thank Derek Lim and Petar Veli\v{c}kovi\'{c} for their supports on this paper.

\clearpage
\bibliography{example_paper}
\bibliographystyle{icml2024}



\end{document}